\documentclass[10pt,twocolumn,letterpaper]{article}

\usepackage{wacv}              %

\usepackage[accsupp]{axessibility} %
\usepackage[T1]{fontenc}
\usepackage{amsmath}
\usepackage{amssymb}
\usepackage{booktabs}
\usepackage{xcolor}
\usepackage{xspace}
\usepackage{tabularx}
\usepackage{multirow}
\usepackage{enumitem}
\usepackage{marvosym}
\usepackage{adjustbox}
\usepackage[toc,page]{appendix}
\usepackage{graphicx} 
\graphicspath{{images/}}  

\usepackage[pagebackref,breaklinks,colorlinks]{hyperref}

\usepackage[capitalize]{cleveref}
\crefname{section}{Sec.}{Secs.}
\Crefname{section}{Section}{Sections}
\Crefname{table}{Table}{Tables}
\crefname{table}{Tab.}{Tabs.}
\Crefname{figure}{Figure}{Figures}
\crefname{figure}{Fig.}{Figs.}

\def\confName{WACV}
\def\confYear{2024}

\definecolor{c1}{HTML}{cd7ca0}
\definecolor{c2}{HTML}{7ca0cd} 
\definecolor{c3}{HTML}{7ccda9}
\definecolor{c4}{HTML}{cda97c}

\newcommand{\condensedparagraph}[1]{\vspace{0.5em}\noindent\textbf{#1}}

\newcommand{\method}[0]{DREAM\xspace} %
\newcommand{\colorname}{\textcolor{c1}{D}\textcolor{c2}{RE}\textcolor{c3}{A}\textcolor{c4}{M}\xspace}
\newcommand{\colortitle}{Visual \textcolor{c1}{D}ecoding from \textcolor{c2}{RE}versing Hum\textcolor{c3}{A}n Visual Syste\textcolor{c4}{M}}

\begin{document}

\title{\colorname: \colortitle}

\author{Weihao Xia\textsuperscript{1~\Letter} \ \ \ \ Raoul de Charette$^{2}$ \ \ \ \ Cengiz Oztireli$^{3}$ \ \ \ \ Jing-Hao Xue$^{1}$ \\
$^{1}$University College London  \ \ \ \ $^{2}$Inria \ \ \ \ $^{3}$University of Cambridge \\
{\tt\footnotesize \{weihao.xia.21,jinghao.xue\}@ucl.ac.uk, raoul.de-charette@inria.fr, aco41@cam.ac.uk}
}

\maketitle

\newcommand\blfootnote[1]{%
\begingroup
\renewcommand\thefootnote{}\footnote{#1}%
\addtocounter{footnote}{-1}%
\endgroup
}

\begin{abstract}
In this work we present~\method, an fMRI-to-image method for reconstructing viewed images from brain activities,
grounded on fundamental knowledge of the human visual system. We craft reverse pathways that emulate the hierarchical and parallel nature of how humans perceive the visual world. These tailored pathways are specialized to decipher semantics, color, and depth cues from fMRI data, mirroring the forward pathways from visual stimuli to fMRI recordings.
To do so, two components mimic the inverse processes within the human visual system: the Reverse Visual Association Cortex (R-VAC) which reverses pathways of this brain region, extracting semantics from fMRI data; the Reverse Parallel PKM (R-PKM) component simultaneously predicting color and depth from fMRI signals.
The experiments indicate that our method outperforms the current state-of-the-art models in terms of the consistency of appearance, structure, and semantics.
Code will be made publicly available to facilitate further research in this field.

\end{abstract}

\section{Introduction}
\label{sec:intro}

Exploring neural encoding unravel the intricacies of brain function. In last years, we have witnessed tremendous progress in visual decoding~\cite{huang2021fmri} which aims at decoding a Functional Magnetic Resonance Imaging (fMRI) to reconstruct the test image seen by a human subject during the fMRI recording. Visual decoding could significantly affect our society from how we interact with machines to helping paralyzed patients~\cite{tan2010brain}.
However, existing methods still suffer from missing concepts and limited quality in the image results. Recent studies turned to deep generative models for visual decoding due to their remarkable generation capabilities, particularly the text-to-image diffusion models~\cite{rombach2022high,xu2022versatile}. These methods heavily rely on aligning brain signals with the vision-language model~\cite{radford2021learning}. This strategic utilization of CLIP helps mitigate the scarcity of annotated data and the complexities of underlying brain information.

Still, the inherent nature of CLIP, which fails to preserve the scene structural and positional information, limits visual decoding. Hence, current methods have endeavored to incorporate structural and positional details, either through depth maps~~\cite{takagi2023improving,ferrante2023brain} or by utilizing the decoded representation of an initial guessed image~\cite{ozcelik2022reconstruction,scotti2023reconstructing}.
However, these methods primarily focus on merging inputs that fit well within the pretrained generative model for visual decoding, lacking the insights from the human visual system.

\begin{figure}[t]
\centering
\includegraphics[width=0.8\linewidth]{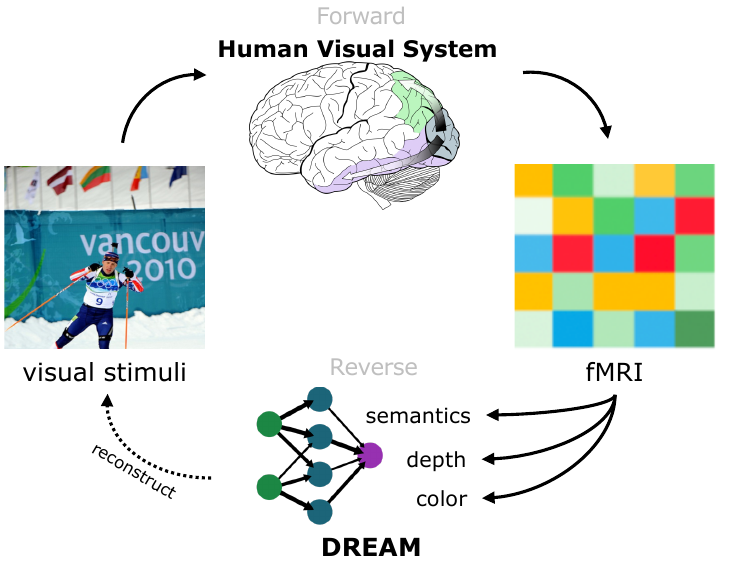} \\
\vspace{-2mm}
\caption{\textbf{Forward and Reverse Cycle}. Forward (HVS): visual stimuli $\mapsto$ color, depth, semantics $\mapsto$ fMRI; Reverse (\colorname): fMRI $\mapsto$ color, depth, semantics $\mapsto$ reconstructed images.
}
\label{fig:teaser}
\end{figure}

\begin{figure}[t]
\centering
\renewcommand{\arraystretch}{0.8}
\setkeys{Gin}{width=0.22\linewidth}
\setlength{\tabcolsep}{1.2pt}
\footnotesize
{
\begin{tabular}[t]{ccccc}
\parbox[t]{2mm}{\multirow{1}{*}{\rotatebox[origin=c]{90}{Test image\hspace{-6em}}}}&\includegraphics{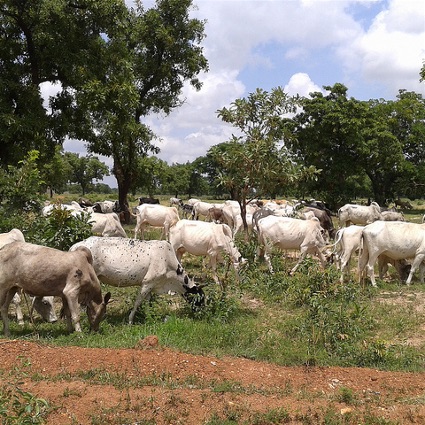} & \includegraphics{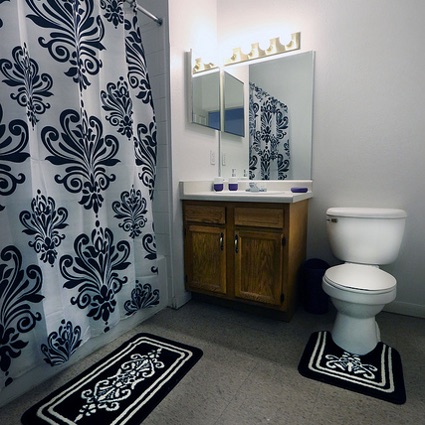} & \includegraphics{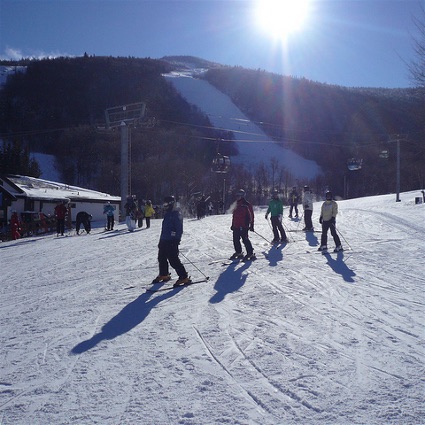} & \includegraphics{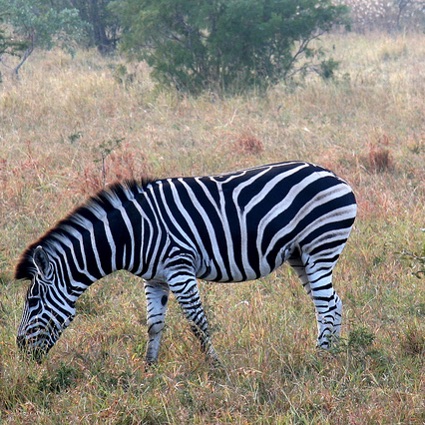} \\
\parbox[t]{2mm}{\multirow{1}{*}{\rotatebox[origin=c]{90}{Reconstruction\hspace{-6em}}}}&\includegraphics{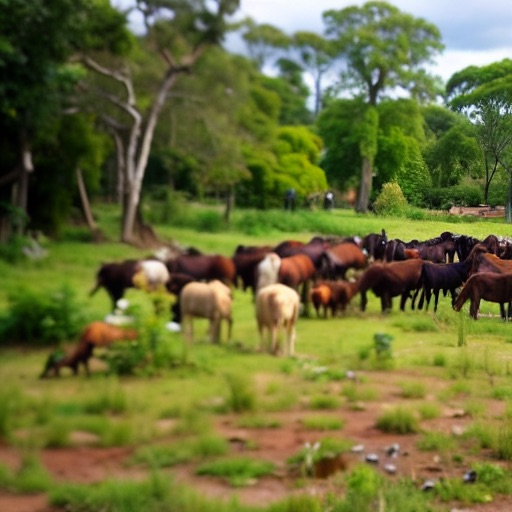} & \includegraphics{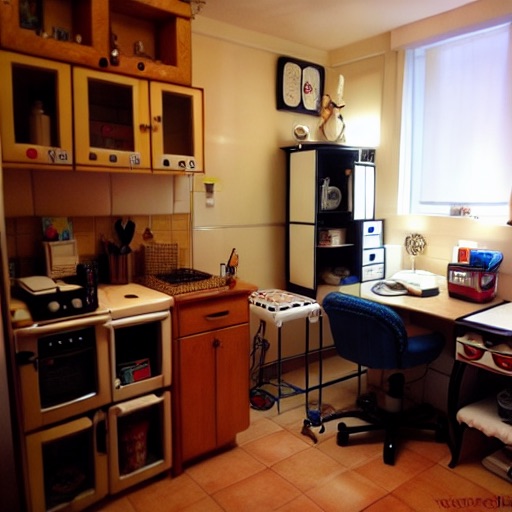} & \includegraphics{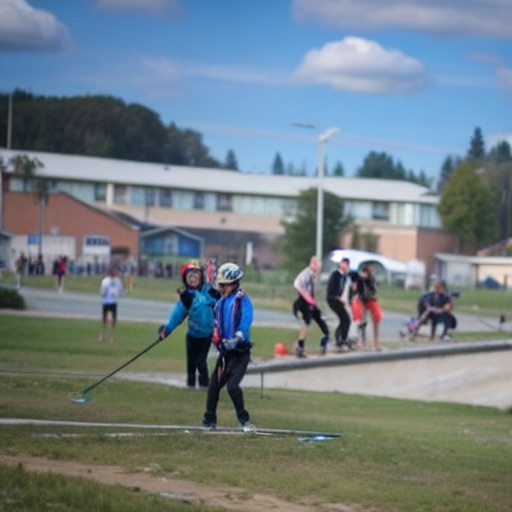} & \includegraphics{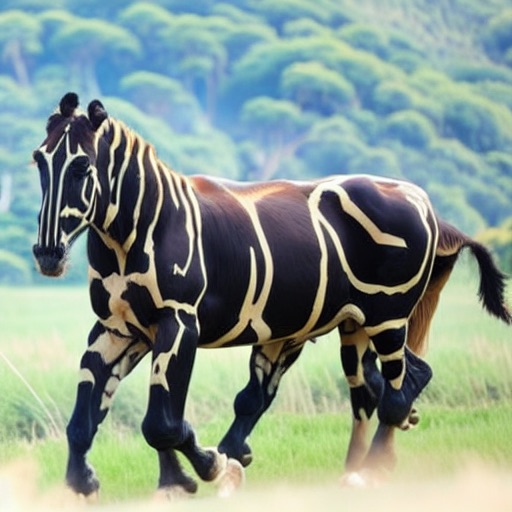} \\
\end{tabular}
\vspace{-2.5mm}
\caption{\textbf{Appearance Inconsistency.}
When decoding the fMRI data of a subject viewing a test image (top), recent visual decoding methods, here~\cite{ozcelik2023brain}, reconstruct images (bottom) which are semantically close but still suffer from strong color inconsistencies.}
\label{fig:color_inconsistency}
}
\end{figure}

We commence our study with the foundational principles~\cite{brodal2004central} governing the Human Visual System (HVS) and dissect essential cues crucial for effective visual decoding. 
Our method draw insights from HVS ---~how humans perceive visual stimuli (forward route in~\cref{fig:teaser})~--- to address the potential information loss during the transition from the fMRI to the visual domain (reverse route in~\cref{fig:teaser}). 
We do that by deciphering crucial cues from fMRI recordings, thereby contributing to enhanced consistency in terms of appearance, structure, and semantics. 
As cues, we investigate: \textit{color} for accurate scene appearance~\cite{mou2023t2i}, \textit{depth} for scene structure~\cite{ranftl2020towards}, and the popular \textit{semantics} for high-level comprehension~\cite{radford2021learning}. 
Our study shows that current visual decoding methods often underscored and unnoticed color  
which in fact plays an indispensable role.
\cref{fig:color_inconsistency} highlights color inconsistencies in a recent work~\cite{ozcelik2022reconstruction}. The generated images, while accurate in semantics, deviate in structure and color from the original visual stimuli. This phenomenon arises due to the absence of proper color guidance.

Following above analysis, we propose \colorname, a visual \textcolor{c1}{D}ecoding method from \textcolor{c2}{RE}versing hum\textcolor{c3}{A}n visual syste\textcolor{c4}{M}.
It aims to mirror the forward process from visual stimuli to fMRI recordings (\cref{sec:preliminary}).
Specifically, we design two reverse pathways specialized in deciphering semantics, color, and depth from fMRI.  
Reverse-VAC (\cref{subsec:method_vac}) replicates the reverse operations of the visual associated cortex, extracting semantics from fMRI. Reverse-PKM (\cref{subsec:method_pkm}) predicts color and depth simultaneously from fMRI signals. 
Deciphered cues are then fed into Stable Diffusion~\cite{rombach2022high} with T2I-Adapter~\cite{mou2023t2i} to guide the image reconstruction (\cref{subsec:meth_generation}).
Our contributions are summarized as follows:
\begin{itemize}%
    \item We scrutinize the limitations of recent diffusion-based visual decoding methods, shedding light on the potential loss of information, and introduce a novel formulation based on the principles of human perception.
    \item We mirror the forward process from visual stimuli to fMRI recordings within the visual system and devise two reverse pathways specialized in extracting semantics, color, and depth information from fMRI data.
    \item We show through experiments that our biologically interpretable method, \method, outperforms state-of-the-art methods while maintaining better consistency of appearance, structure, and semantics.
\end{itemize}

\begin{figure*}[t!]
\centering
\vspace{-2.5mm}
\includegraphics[width=0.85\linewidth]{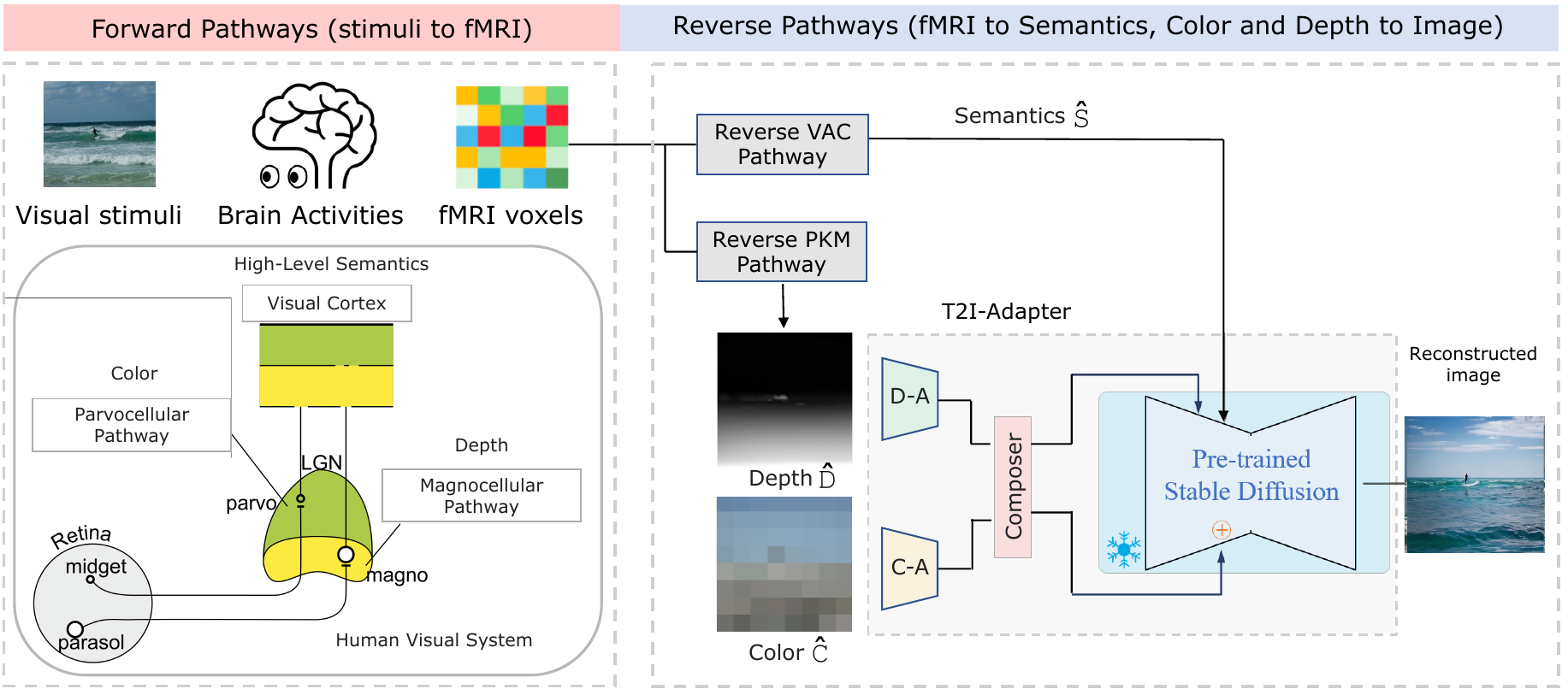}
\vspace{-2.5mm}
\caption{\textbf{Relation of the HVS and Our proposed~\method.} Grounding on the Human Visual System (HVS), we devise reverse pathways aimed at deciphering semantics, depth, and color cues from fMRI to guide image reconstruction.
\textbf{(Left)} Schematic view of HVS, detailed in~\cref{sec:preliminary}. 
When perceiving visual stimuli, connections from the retina to the brain can be separated into two parallel pathways.
The Parvocellular Pathway originates from midget cells in the retina and is responsible for transmitting color information, while the Magnocellular Pathway starts with parasol cells and is specialized in detecting depth and motion. 
The conveyed information is channeled into the visual cortex for undertaking intricate processing of high-level semantics from the visual image. 
\textbf{(Right)} \method mimics the corresponding inverse processes within the HVS: the Reverse VAC (\cref{subsec:method_vac}) replicates the opposite operations of this brain region, analogously extracting semantics $\hat{\texttt{S}}$ as a form of CLIP embedding from fMRI; and the Reverse PKM (\cref{subsec:method_pkm}) maps fMRI to color $\hat{\texttt{C}}$ and depth $\hat{\texttt{D}}$ in the form of spatial palettes and depth maps to facilitate subsequent processing by the Color Adapter (C-A) and the Depth Adapter (D-A) in T2I-Adapter~\cite{mou2023t2i} in conjunction with SD~\cite{rombach2022high} for image reconstruction from deciphered semantics, color, and depth cues $\{\hat{\texttt{S}}, \hat{\texttt{C}}, \hat{\texttt{D}}\}$.
}
\label{fig:overview}
\end{figure*}

\section{Related Work}
\label{sec:related_work}

\subsection{Diffusion Probabilistic Models}

Recently, diffusion models have risen to prominence as cutting-edge generative models. Denoising Diffusion Probabilistic Model is a parameterized  bi-directional Markov chain that utilizes variational inference to produce matching samples. The forward diffusion process is designed to transform any data distribution into a basic prior distribution (e.g., isotropic Gaussian), and the reverse denoising process learns to denoise by learning transition kernels parameterized by deep neural networks such as U-Net. 
In Latent Diffusion Models (LDMs)~\cite{rombach2022high}, diffusion process is applied within the latent space rather than in the pixel space, enabling faster inference and reducing training costs. 
The text-conditioned LDM, known as Stable Diffusion (SD), has gained widespread usage due to its versatile applications and capabilities. 
ControlNet~\cite{zhang2023adding} and T2I-Adapter~\cite{mou2023t2i} aim to enhance the control capabilities even more by training versatile modality-specific encoders. 
These encoders align external control (\eg, sketch, depth, and spatial palette) with internal knowledge in SD, thereby enabling more precise control over the generated output.
Unlike SD, which solely employs the CLIP text encoder, Versatile Diffusion (VD)~\cite{xu2022versatile} incorporates both CLIP text and image encoders, thereby enabling the utilization of multimodal capabilities.

\subsection{Image Decoding from fMRI}

The advancements in visual decoding are closely intertwined with the evolution of various modeling frameworks. For instance, in~\cite{horikawa2017generic}, sparse linear regression was applied to preprocessed fMRI data to predict features extracted from early convolutional layers in a pretrained CNN.
In the past few years, researchers have advanced visual decoding techniques by mapping the brain signals to the latent space of generative adversarial networks (GANs)~\cite{goodfellow2020generative} to reconstruct human faces~\cite{dado2022hyperrealistic} and natural scenes~\cite{shen2019deep,ozcelik2022reconstruction}. 
More recently, visual decoding has reached an unprecedented level of quality~\cite{ozcelik2023brain,takagi2023high,gu2022decoding} with the release of vision-language models~\cite{radford2021learning}, multimodal diffusion models~\cite{rombach2022high,xu2022versatile}, and large-scale fMRI datasets~\cite{allen2022massive}.
Lin~\etal~\cite{lin2022mind} learned to project voxels to the CLIP space and then processed outcomes through a fine-tuned conditional StyleGAN2~\cite{karras2020analyzing} to reconstruct natural images. 
Takagi et al.~\cite{takagi2023high} employed the ridge regression to associate fMRI signals with the CLIP text embedding and the latent space of Stable Diffusion, opting for varied voxels based on different components.
Recent research ~\cite{ozcelik2023brain,lu2023minddiffuser} explored the process of mapping fMRI signals to both CLIP text and image embeddings, subsequently utilizimg the pre-trained Versatile Diffusion model~\cite{xu2022versatile} that accommodates multiple inputs for image reconstruction.

\subsection{Multi-Level Modeling in Visual Decoding}

Hierarchical visual feature representations are frequently utilized in visual decoding.
Early studies~\cite{agrawal2014pixels,horikawa2017generic} have indicated that hierarchical features extracted through pretrained CNN models demonstrated strong correlation with neural activities of visual cortices.
Recent research delved into combining both low-level visual cues and high-level semantics inferred from brain activity, using frozen diffusion models for reconstruction~\cite{ozcelik2023brain,lu2023minddiffuser,zeng2023controllable,scotti2023reconstructing}.
The low-level visual cues are commonly incorporated in an implicit manner, such as utilizing intermediate features predicted by a large-scale vision model~\cite{ozcelik2023brain} or encoded from an initial estimated image~\cite{scotti2023reconstructing}.
The high-level semantics is frequently represented as CLIP embedding.
Recent research~\cite{takagi2023improving, ferrante2023brain} also suggested the explicit provision of both low-level and high-level information, predicting captions and depth maps from brain signals.
Our method differs in terms of how to predict auxiliary information and the incorporation of color.

\section{Preliminary on the Human Visual System}
\label{sec:preliminary}

Human Visual System (HVS) endows us with the ability of visual perception.
The visual information is \textit{concurrently} relayed from various cell types in the retina, each capturing distinct facets of data, through the optic nerve to the brain.
Connections from the retina to the brain, as shown in~\cref{fig:overview}, can be separated into a {parvocellular} pathway and a {magnocellular} pathway\footnote{An additional set of neurons, known as the koniocellular layers, are found ventral to each of the magnocellular and parvocellular layers~\cite{brodal2004central}.}. 
The parvocellular pathway originates from midget cells in the retina and is responsible for transmitting color information, while the magnocellular pathway starts with parasol cells and is specialized in detecting depth and motion. 
The visual information is first directed to a sensory relay station known as the lateral geniculate nucleus (LGN) of the thalamus, before being channeled to the visual cortex (V1) for the initial processing of visual stimuli.
The visual association cortex (VAC) receives processed information from V1 and undertakes intricate processing of high-level semantic contents from the visual image.

The hierarchical and parallel manner where visual stimuli are broken down and passed forward as color, depth, and semantics guided our choices to reverse the HVS for decoding.
For a detailed illustration of human perception and an analysis on the feasibility of extracting desired cues from fMRI recordings, please consult supplementary material.

\section{\method}
\label{sec:method}

The task of visual decoding aims to recover the viewed image \makebox{$I \in \mathbb{R}^{H\times{}W\times{}3}$} from brain activity signals elicited by visual stimuli. %
Functional MRI (fMRI) is usually employed as a proxy of the brain activities, typically encoded as a set of voxels $\texttt{fMRI} \in \mathbb{R}^{1\times{}N}$. 
Formally, the task optimizes $f(\cdot)$ so that $f(\texttt{fMRI}) = \hat{I}$, where $\hat{I}$ best approximates $I$.

To address this task, we propose~\colorname, a method grounded on fundamental principles of human perception.
Following~\cref{sec:preliminary}, our method relies on explicit design of reverse pathways to decipher {$\texttt{S}$emantics}, {$\texttt{C}$olor}, and {$\texttt{D}$epth} intertwined in the fMRI data. %
These reverse pathways mirror the forward process from visual stimuli to brain activity.
Considering that an fMRI captures changes in the brain regions during the forward process, it is feasible to derive the desired cues of the visual stimuli from such recording~\cite{allen2022massive}.

\paragraph{Overview.} 
\cref{fig:overview} illustrates an overview of \method.
It is constructed on two consecutive phases, namely, Pathways Reversing and Guided Image Reconstruction. These phases break down the reverse mapping from fMRI to image
into two subprocesses: \makebox{$\texttt{fMRI} \mapsto \{\hat{\texttt{S}}, \hat{\texttt{C}}, \hat{\texttt{D}}\}$} and \makebox{$\{\hat{\texttt{S}}, \hat{\texttt{C}}, \hat{\texttt{D}}\} \mapsto {\hat{\texttt{I}}}$}.
In the first phase, two~{Reverse Pathways} decipher the cues of semantics, color and depth from fMRI with parallel components:
Reverse Visual Association Cortex (R-VAC, \cref{subsec:method_vac}) inverts operations of the VAC region to extract semantic details from the fMRI, encoded as CLIP embedding~\cite{radford2021learning}, and Reverse Parallel Parvo-, Konio- and Magno-Cellular (R-PKM, ~\cref{subsec:method_pkm}) is designed to predict color and depth simultaneously from fMRI signals.
Given the lossy nature of fMRI data and the non-bijective transformation of image $\mapsto$ fMRI, %
we then cast the decoding process as a generative task while using the extracted ${\hat{\texttt{S}}, \hat{\texttt{C}}, \hat{\texttt{D}}}$ cues as conditions for image reconstruction.
Therefore, in the second phase {Guided Image Reconstruction} (GIR,~\cref{subsec:meth_generation}), we follow recent visual decoding practices~\cite{takagi2023high,takagi2023improving} and employ a frozen SD with T2I-Adapter~\cite{mou2023t2i} to generate images through benefiting here the additional ${\hat{\texttt{S}}, \hat{\texttt{C}}, \hat{\texttt{D}}}$ guidance.

\begin{figure}
    \centering
    \includegraphics[width=0.98\linewidth]{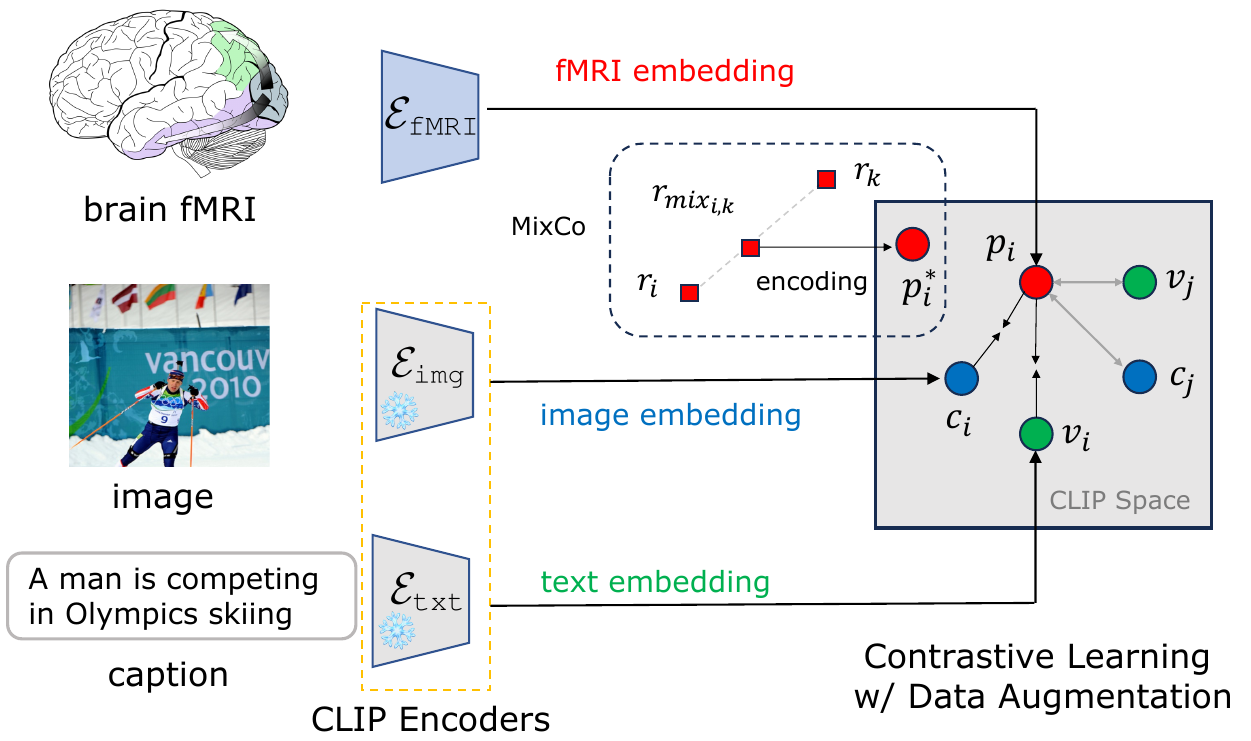}
    \caption{\textbf{R-VAC Training.} To decipher semantics from fMRI, we train an encoder $\mathcal{E}_{\texttt{fmri}}$ in a contrastive fashion which aligns fMRI data with the frozen CLIP space~\cite{radford2021learning}. 
    Data augmentation (represented by the dashed rectangle)~\cite{kim2020mixco} combats the data scarcity of the fMRI modality. See~\cref{subsec:method_vac} for more details.
    }
    \label{fig:method_vac}
\end{figure}

\subsection{R-VAC (Semantics Decipher)}
\label{subsec:method_vac}

The Visual Association Cortex (VAC), as detailed in ~\cref{sec:preliminary}, is responsible for interpreting the high-level semantics of visual stimuli.
We design R-VAC to reverse such process through analogous learning of the mapping from fMRI to semantics \makebox{$\texttt{fMRI}~{\mapsto}~\hat{\texttt{S}}$}.
This is achieved by training an encoder $\mathcal{E}_\texttt{fmri}$ whose goal is to align the fMRI embedding with the shared CLIP space~\cite{radford2021learning}. Though CLIP was initially trained with image-text pairs, prior works~\cite{scotti2023reconstructing,liu2023brainclip} demonstrated ability to align new modalities. 
To fight the scarcity of fMRI data, we also carefully select ad-hoc data augmentation strategy~\cite{kim2020mixco}.

\paragraph{Contrastive Learning.} In practice, we train the fMRI encoder $\mathcal{E}_\texttt{fMRI}$ with triplets of \{fMRI, image, caption\} to pull the fMRI embeddings closer to the rich shared semantic space of CLIP.  %
Given that both text encoder~($\mathcal{E}_\texttt{txt}$) and image encoder~($\mathcal{E}_\texttt{img}$) of CLIP are frozen, we minimize the embedding distances of fMRI-image and fMRI-text which in turns forces alignment of the fMRI embedding with CLIP. The training is illustrated in \cref{fig:method_vac}.
Formally, with the embeddings of fMRI, text, and image denoted by $p$, $c$, $v$, respectively, the initial contrastive loss writes
\begin{equation}
  \mathcal{L}_{p}= - \log \frac{\exp \left(p_i{\cdot}c_i / {\tau}\right)}{\sum_{j=0}^K \exp \left(p_i{\cdot}c_j / \tau\right)} - \log \frac{\exp \left(p_i{\cdot}v_i/{\tau}\right)}{\sum_{j=0}^K \exp \left(p_i {\cdot} v_j / {\tau}\right)},
  \label{eqn:contrastive}
\end{equation}
where $\tau$ is a temperature hyperparameter. The sum for each term is over one positive and $K$ negative samples.
Each term represents the log loss of a $({K}{+}1)$-way softmax-based classifier~\cite{he2020momentum}, which aims to classify $p_i$ as $c_i$ (or $v_i$). 
The sum over samples of the batch size $n$ is omitted for brevity. The joint image-text-fMRI representation is intended for potential retrieval purposes~\cite{scotti2023reconstructing}. The fMRI-image or fMRI-text components are utilized for specific tasks.

\paragraph{Data Augmentation.} An important issue to consider is that there are significant fewer fMRI samples (${\approx{}}10^4$) compared to the number of samples used to train CLIP~($10^8$), which may damage contrastive learning~\cite{zhu2022balanced,jiang2021self}.
To address this, we utilize a data augmentation loss based on MixCo~\cite{kim2020mixco}, which generates mixed fMRI data $r_{\text{mix}_{i, k}}$ from convex combination of two fMRI data $r_i$ and $r_k$: 
\begin{equation}
r_{\text{mix}_{i, k}}=\lambda_i \cdot r_i+\left(1-\lambda_i\right) \cdot r_k,
\end{equation}
where $k$ represents the arbitrary index of any data in the same batch, and its encoding writes $p_i^* = \mathcal{E}_\texttt{fmri}(r_{\text{mix}_{i, k}})$.
The data augmentation loss, which excludes the image components for brevity, is formulated as

\begin{equation}
\begin{aligned}
\mathcal{L}_\text{MixCo}= & -\sum_{i=1}^{n} \biggl[\lambda_i \cdot \log \frac{\exp \left(p_i^*{\cdot}c_i/{\tau}\right)}{\sum_{j=0}^K \exp \left(p_i^*{\cdot}c_j/{\tau}\right)}  \\
& + \left(1-\lambda_i\right) \cdot \log \frac{\exp \left(p_i^*{\cdot}c_{k}/{\tau}\right)}{\sum_{j=0}^K \exp \left(p_i^*{\cdot}c_j/{\tau}\right)} \biggr].
\end{aligned}
\end{equation}

Finally, the total loss is a combination of $\mathcal{L}_p$ and $\mathcal{L}_\text{MixCo}$ weighted with hyperparameter $\alpha$:
\begin{equation}
\mathcal{L}_{total}=\mathcal{L}_p +\alpha \mathcal{L}_\text{MixCo}\,.
\end{equation}

\subsection{R-PKM (Depth \& Color Decipher)}
\label{subsec:method_pkm}

While R-VAC provides semantics knowledge, the latter is inherently bounded by the CLIP space capacity, unable to encode spatial colors and geometry. 
To address this issue, inspired by the human visual system, we craft the R-PKM component to reverse pathways of the Parvo-, Konio- and Magno-Cellular (PKM), subsequently predicting color and depth from the fMRI data denoted as $\texttt{fMRI} {\mapsto} \{\hat{\texttt{C}}, \hat{\texttt{D}}\}$.
While color and depth can be represented in various ways (\eg, histograms, graphs), we represent them as~\textit{spatial color palettes} and \textit{depth maps} to facilitate the reconstruction guidance, as discussed in~\cref{subsec:meth_generation}. Visuals are in~\cref{fig:overview}.

In practice, we formulate the problem as RGBD estimation. 
The color palette is then derived from the RGB prediction by first $\times$64 downscaling it and then upscaling back to its original size.
There are readily available methods for {{fMRI} {$\mapsto$} {RGBD} %
 mapping~\cite{takagi2023improving,ferrante2023brain} but they offer limited performance due to the scarcity fMRI data. 
Instead, we introduce a multi-stage encoder-decoder training~\cite{gaziv2021more}, which benefits from both the scarcely available (fMRI, RGBD) pairs and the abundant RGBD data without fMRI. 
\cref{fig:method_pkm} shows the training procedure for R-PKM.

\condensedparagraph{Stage 1.} Given limited pairs $\{(r, d)\}$=\{fMRI, RGBD\}, we first train an encoder to map RGBD to their corresponding fMRI data. 
To compensate for the absence of depth in fMRI datasets, we use MiDaS-estimated depth maps~\cite{ranftl2020towards} as surrogate ground-truth depth.
The encoder is trained with a convex combination of mean square error and cosine proximity between the input  $r$ and its predicted counterpart $\hat{r}$:
\begin{equation}
    \mathcal{L}_r(r, \hat{r})=\beta \cdot \operatorname{MSE}(r, \hat{r})-(1-\beta) \cos (\angle(r, \hat{r})), 
\end{equation}
where $\beta$ is determined empirically as a hyperparameter.

\condensedparagraph{Stage 2.} Similar to stage 1, we now train the decoder with pairs $\{(r, d)\}$ in a supervised manner:
\begin{equation}
    \mathcal{L}_s(d, \hat{d})=\|d-\hat{d}\|_1+\mathcal{J}(\hat{d}),
    \label{eqn:decoder_loss}
\end{equation}
where $\hat{d}=\mathcal{D}(r)$ and the total variation regularization $\mathcal{J}(\hat{d})$ encourages spatial smoothness in the reconstructed $\hat{d}$.

\condensedparagraph{Stage 3.} To address the scarcity of fMRI data and improve the model generalization to unseen categories, we employ a self-supervised strategy to finetune the decoder while keeping the encoder frozen. This facilitate the usage of any natural images (\eg, from ImageNet~\cite{deng2009imagenet} or LAION~\cite{schuhmann2022laion}) along with their estimated depth maps, without need of paired fMRI and image data.
Hence, we train solely with the RGBD data by ensuring a cycle consistency through the Encoder-Decoder transformation, \ie $\hat{d} = \mathcal{D}(\mathcal{E}(d))$, with the loss in~\cref{eqn:decoder_loss}. 
Given that this stage involves images for which fMRI data was never collected, the model greatly improves its generalization capability.

\begin{figure}[t!]
    \centering
    \includegraphics[width=0.98\linewidth]{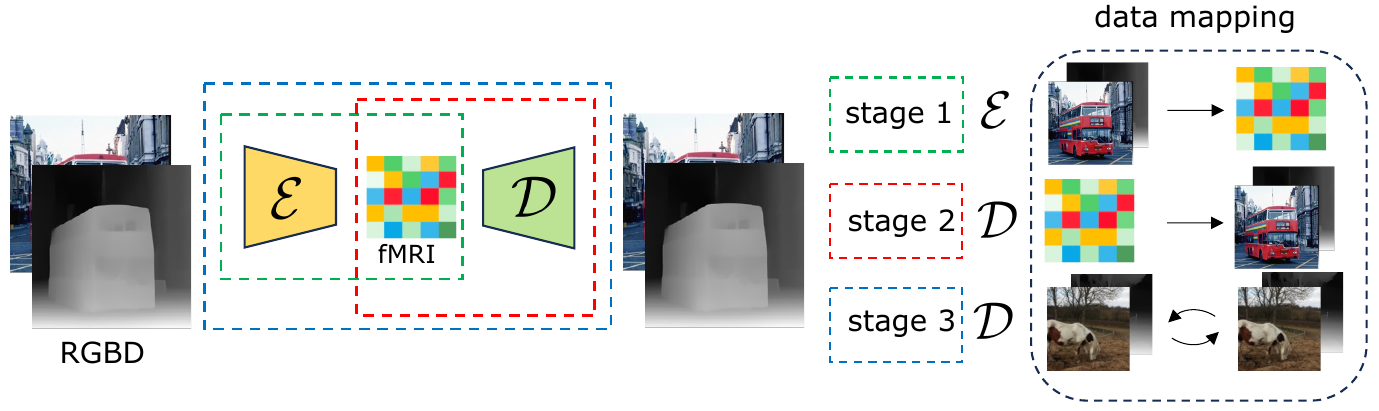}
    \caption{\textbf{R-PKM Training.} 
        Our multi-stage training reverses the PKM pathway and decodes color and depth cues in fMRI data.
        Stages 1 and 2 employ (RGBD, fMRI) pairs to train an encoder $\mathcal{E}$ that maps RGBD to fMRI and a decoder $\mathcal{D}$ to decode RGBD from fMRI. 
        Stage 3 benefits from additional RGBD images without fMRI to train $\mathcal{D}$ in a cycle-consistent manner $\hat{d} = \mathcal{D}(\mathcal{E}(d))$, while keeping $\mathcal{E}$ frozen. 
        See~\cref{subsec:method_pkm} for details.
    }
    \label{fig:method_pkm}
\end{figure}

\subsection{Guided Image Reconstruction (GIR)}
\label{subsec:meth_generation}

Equipped with R-VAC~(\cref{subsec:method_vac}) and R-PKM~(\cref{subsec:method_pkm}), our method can decipher semantics $\hat{\texttt{S}}$, color $\hat{\texttt{C}}$, and depth $\hat{\texttt{D}}$ in the form of CLIP embedding, spatial color palette, and depth map. 
Finally, guided image reconstruction {$\{\hat{\texttt{S}}, \hat{\texttt{C}}, \hat{\texttt{D}}\} \mapsto \hat{I}$} completes the reverse mapping of the forward process during the visual perception $I \mapsto \texttt{fMRI}$.

We utilize Stable Diffusion (SD)~\cite{rombach2022high} to reconstruct the final image from the predicted CLIP embedding $\hat{\texttt{S}}$ and the additional guidance from predicted color palette $\hat{\texttt{C}}$ and depth map $\hat{\texttt{D}}$.
Such guidance is produced using the color adapter $\mathcal{R}_c$ and the depth adapter $\mathcal{R}_d$ within T2I-adapter~\cite{mou2023t2i}.
This process is formulated as follows:
\begin{equation}
\label{eq:im_guided}
\begin{aligned}
\text{F}_{\mathcal{R}} & = \omega_c \mathcal{R}_c\left(\hat{\texttt{C}}\right) + \omega_d \mathcal{R}_d\left(\hat{\texttt{D}}\right), \\
\hat{I}  & = \text{SD}\left(z, \text{F}_{\mathcal{R}}, \hat{\texttt{S}}\right),
\end{aligned}
\end{equation}
where $z$ is a random noise, $\omega_c$ and $\omega_d$ are adjustable weights to control the relative significance of the adapters.

\begin{figure*}
    \centering
    \includegraphics[width=0.85\linewidth]{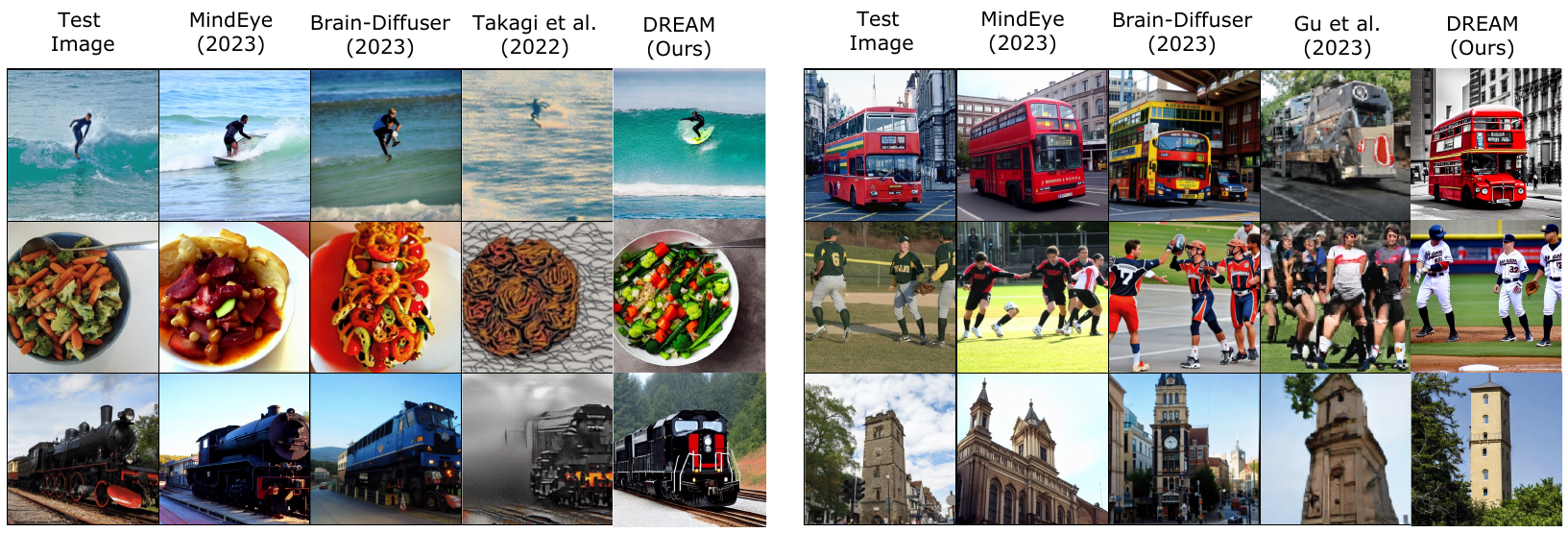}
    \vspace{-2mm}
    \caption{\textbf{Sample Visual Decoding Results from the SOTA Methods on NSD.} 
    }
    \label{fig:recon_comparison}
\end{figure*}

\begin{table*}
    \centering
    \caption{\textbf{Quantitative Evaluation.} Following standard NSD metrics, \method performs on a par or better than the SOTA methods (we highlight \textbf{best} and \underline{second}). 
    We also report ablation of the two strategies fighting fMRI data scarcity: R-VAC without Data Augmentation (DA) and R-PKM without the third-stage decoder training (S3) that allows additional RGBD data without fMRI.
    }
    \label{tab:quantitative_nsd}
    \vspace{-2mm}
    \resizebox{0.85\linewidth}{!}{
        \begin{tabular}{l|cccc|cccc}
            \toprule
            \multirow{2}{*}{Method}  & \multicolumn{4}{c|}{Low-Level} & \multicolumn{4}{c}{High-Level}\\
            ~ & PixCorr $\uparrow$ & SSIM $\uparrow$ & AlexNet(2) $\uparrow$ & AlexNet(5) $\uparrow$ & Inception $\uparrow$ & CLIP $\uparrow$  & EffNet-B $\downarrow$ & SwAV $\downarrow$\\
            \midrule
            Mind-Reader~\cite{lin2022mind}  & $-$ & $-$ & $-$ & $-$  & $78.2\%$ & $-$ & $-$ & $-$\\
            Takagi~\etal~\cite{takagi2023high} & $-$ & $-$ &  $83.0\%$ & $83.0\%$  & $76.0\%$ & $77.0\%$ & $-$ & $-$\\
            Gu~\etal~~\cite{gu2022decoding}  & $.150$& $.325$ & $-$ & $-$  & $-$ & $-$ & $.862$ & $.465$\\
            Brain-Diffuser~\cite{ozcelik2023brain}  & $.254$ & $\textbf{.356}$ & $\underline{94.2}\%$ & $96.2\%$  & ${87.2\%}$ & $\underline{91.5\%}$ & $\underline{.775}$ & ${.423}$ \\
            MindEye~\cite{scotti2023reconstructing} & $\textbf{.309}$ & $.323$ & $\textbf{94.7}\%$ & $\textbf{97.8}\%$ & $\textbf{93.8}\%$ & $\textbf{94.1}\%$ & $\textbf{.645}$ & $\textbf{.367}$ \\
            \method (Ours) & $\underline{.274}$    & $\underline{.328}$ & 93.9\%      & $\underline{96.7}\%$      & $\underline{93.4}\%$     & $\textbf{94.1}\%$ & $\textbf{.645}$     & $\underline{.418}$ \\
            \midrule
            \method (sub01) & {.288} & {.338} & {95.0}\%  & 97.5\%  & {94.8}\% & {95.2}\% & {.638} & {.413} \\
             w/o DA (R-VAC) & .279 & .340 & 86.8$\%$  & 88.1$\%$  & 87.2$\%$ & 89.9$\%$ & .662 & .517\\ 
             w/o S3 (R-PKM) & .203 & .295 & 92.7$\%$  & 96.2$\%$  & 92.1$\%$ & 94.6$\%$ & .642 & .463 \\ 
             \bottomrule
        \end{tabular}
    }
\end{table*}

\section{Experiments}
\label{sec:experiments}

Following common practice in the field, we evaluate \method with the largest public neuroimaging dataset, the Natural Scene Dataset~\cite{allen2022massive}. We report our performance against five leading methods~\cite{lin2022mind,takagi2023high,gu2022decoding,ozcelik2023brain,scotti2023reconstructing}. We detail our experimental methodology in~\cref{sec:exp_setting} and report quantitative and qualitative evaluations in~\cref{sec:exp_results}. Ablation studies are presented in~\cref{sec:ablation}.

\subsection{Experimental Setting}
\label{sec:exp_setting}

\paragraph{Dataset.} 

We use the Natural Scenes Dataset (NSD)~\cite{allen2022massive} in all experiments, which follows the standard practices in the field~\cite{ozcelik2023brain,scotti2023reconstructing,gu2022decoding,takagi2023high,lin2022mind,lu2023minddiffuser}. 
NSD, as the largest fMRI dataset, records brain responses from eight human subjects successively isolated in an MRI machine and passively observed a wide range of visual stimuli, namely, natural images sourced from MS-COCO~\cite{lin2014microsoft}, which allows retrieving the associated captions. 
In practice, because brain activity patterns highly vary across subjects~\cite{huang2021fmri}, a separate model is trained per subject. 
The standardized splits contain 982 fMRI test samples and 24,980 fMRI training samples. 
Please refer to the supplementary material for more details.

\paragraph{Metrics for Visual Decoding.} 
 
The same set of eight metrics is utilized for our evaluation in accordance with prior research~\cite{ozcelik2023brain,scotti2023reconstructing}.  
To be specific, PixCorr is the pixel-level correlation between the reconstructed and ground-truth images.
PixCorr is the pixel-level correlation between the reconstructed and ground-truth images.
Structural Similarity Index (SSIM)~\cite{wang2004image} quantifies similarity between two images. It measures the structural and textural similarity rather than just pixel-wise differences.
AlexNet(2) and AlexNet(5) are two-way comparisons of the second and fifth layers of AlexNet~\cite{krizhevsky2017imagenet}, respectively. 
Inception is the two-way comparison of the last pooling layer of InceptionV3~\cite{szegedy2016rethinking}. CLIP is the two-way comparison of the last layer of the CLIP-Vision~\cite{radford2021learning} model. EffNet-B and SwAV are distances gathered from EfficientNet-B1~\cite{tan2019efficientnet} and SwAV-ResNet50~\cite{caron2020unsupervised}, respectively. The first four metrics can be categorized as low-level measurements, whereas the remaining four capture higher-level characteristics.

\condensedparagraph{Metrics for Depth and Color.} We use metrics from depth estimation and color correction to assess depth and color consistencies in the final reconstructed images. The depth metrics, as elaborated in~\cite{ming2021deep}, include Abs~Rel (absolute error), Sq~Rel (squared error), RMSE (root mean squared error), and RMSE~log (root mean squared logarithmic error). For color metrics, we use CD (Color Discrepancy)~\cite{xia2017color} and STRESS (Standardized Residual Sum of Squares)~\cite{garcia2007measurement}. 
Please consult the supplementary material for details.

\begin{table}
\centering
\caption{\textbf{Consistency of the Decoded Images.} 
We evaluate the color and depth consistencies in decoded images by comparing the distances between test images and visual decoding results. \method significantly outperforms the other two methods~\cite{ozcelik2023brain,scotti2023reconstructing}.
}
\label{tab:depth_color_estimation}
\small
\resizebox{\linewidth}{!}
{
    \setlength{\tabcolsep}{0.01\linewidth}
    \begin{tabular}{l|cccc|cc}
    \toprule
    \multirow{2}{*}{Method}   & \multicolumn{4}{c|}{Depth} & \multicolumn{2}{c}{Color}\\
    ~ & Abs Rel $\downarrow$ & Sq Rel $\downarrow$ & RMSE $\downarrow$ & RMSE log $\downarrow$ & CD $\downarrow$  & STRESS $\downarrow$ \\
    \midrule
    Brain-Diffuser~\cite{ozcelik2023brain}  & 10.162 & 4.819  & \underline{9.871} & 1.157 & 4.231 & 47.025  \\ 
    MindEye~\cite{scotti2023reconstructing} & \underline{8.391} & \underline{4.176} & 9.873 & \underline{1.075}  & \underline{4.172} & \underline{45.380} \\ 
    \method (Ours)     & \textbf{7.695} & \textbf{4.031} & \textbf{9.862} & \textbf{1.039} & \textbf{2.957} & \textbf{37.285} \\ 
    \bottomrule
\end{tabular}}
\end{table}

\begin{figure*}
\centering
\renewcommand{\arraystretch}{0.5}
\setkeys{Gin}{width=0.08\linewidth}
\setlength{\tabcolsep}{1.2pt}
\footnotesize
{
\begin{tabular}[t]{ccccccccccc}
\multicolumn{1}{c}{{Test}} & \multicolumn{2}{c}{{Ground-truth (${\texttt{D}}, {\texttt{C}}$)}} & \multicolumn{2}{c}{{Predictions ($\hat{\texttt{D}}, \hat{\texttt{C}}$)}} & \multicolumn{3}{c}{{\colorname{}}} & \multicolumn{3}{c}{{\colorname{} w/o Color Guidance}}\\
\cmidrule(lr){1-1}\cmidrule(lr){2-3}\cmidrule(lr){4-5}\cmidrule(lr){6-8}\cmidrule(lr){9-11}
\includegraphics{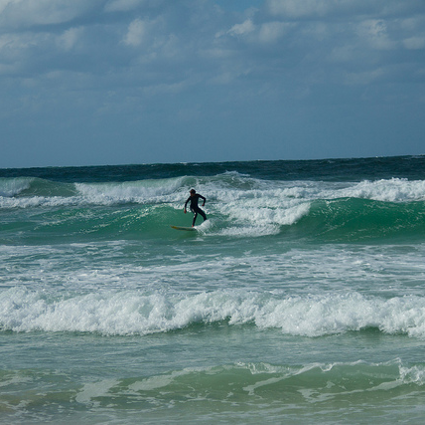} & \includegraphics{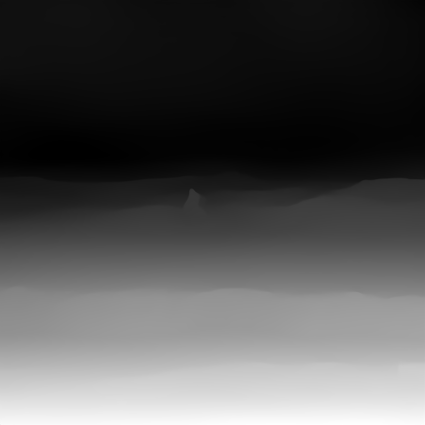} & 
\includegraphics{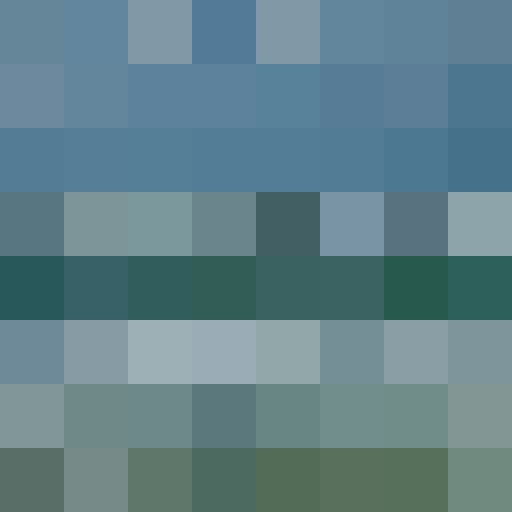} & 
\includegraphics{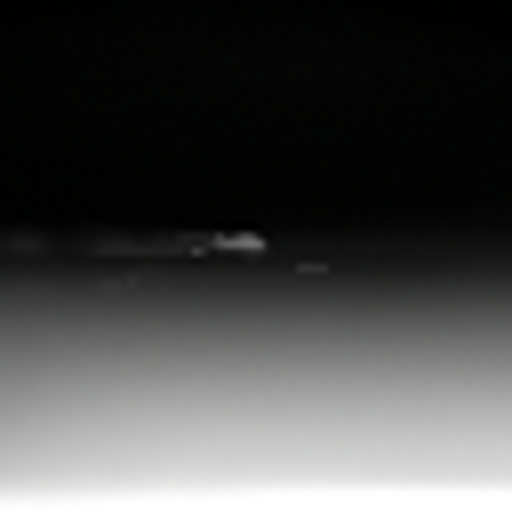} & 
\includegraphics{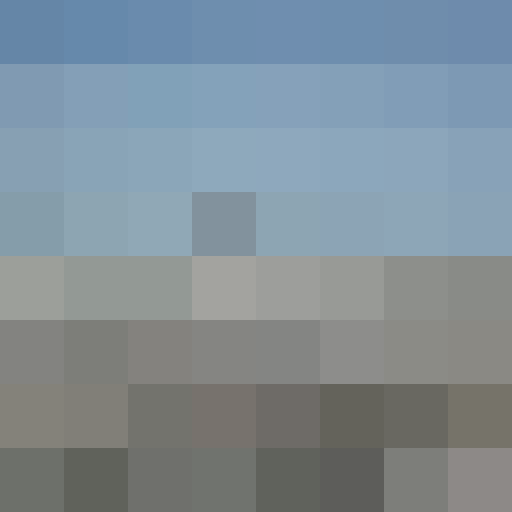} & \includegraphics{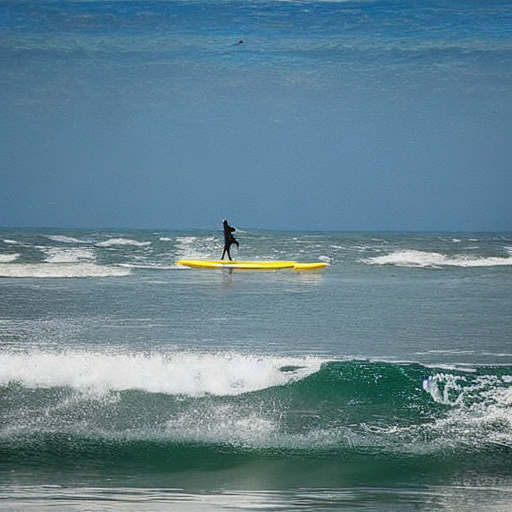} & \includegraphics{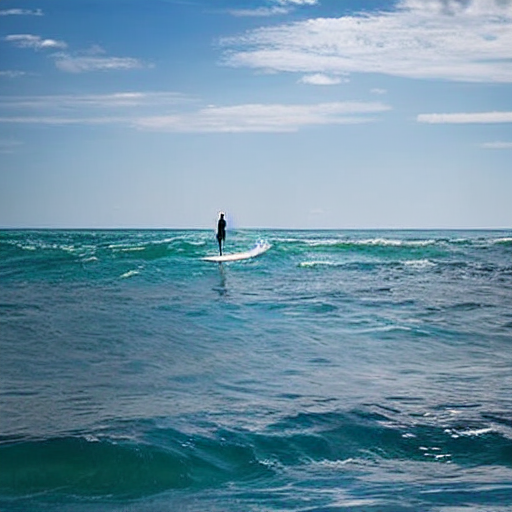} & \includegraphics{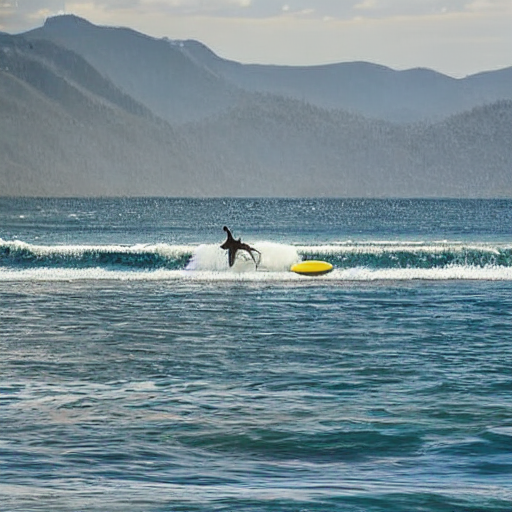} & 
\includegraphics{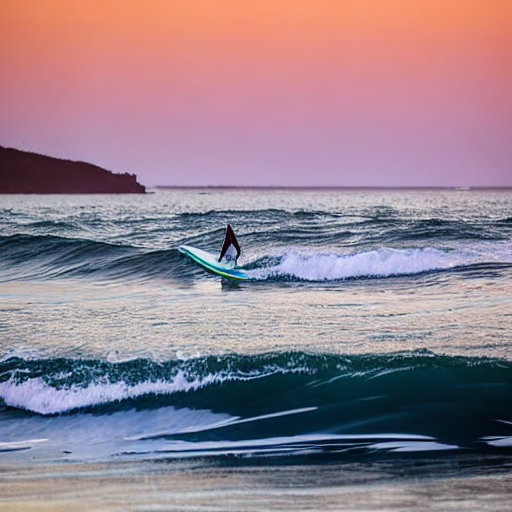} & \includegraphics{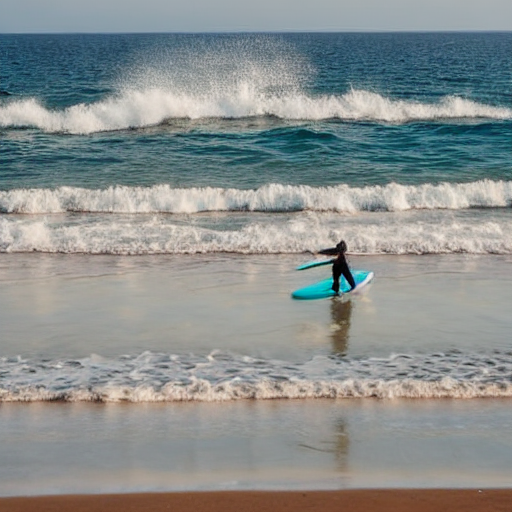} & \includegraphics{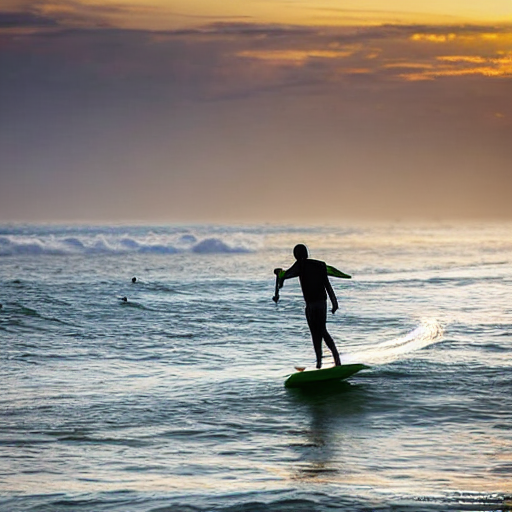} \\
\includegraphics{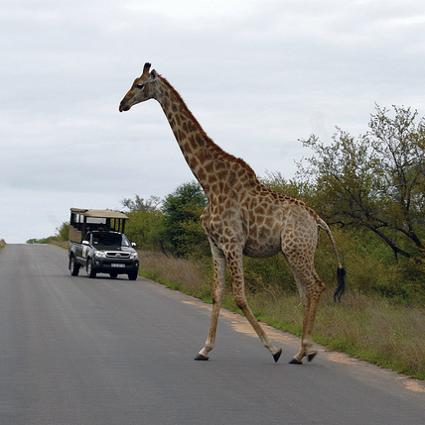} & 
\includegraphics{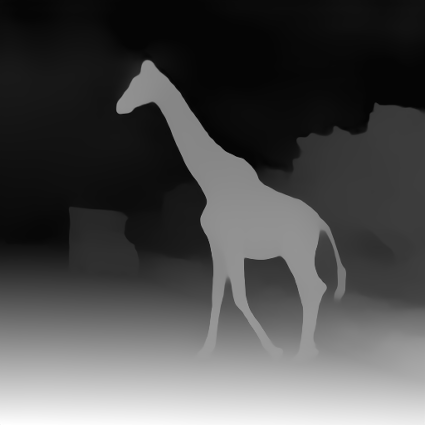} & 
\includegraphics{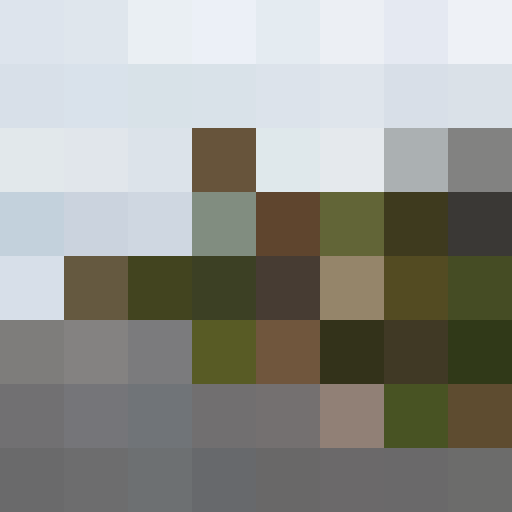} &
\includegraphics{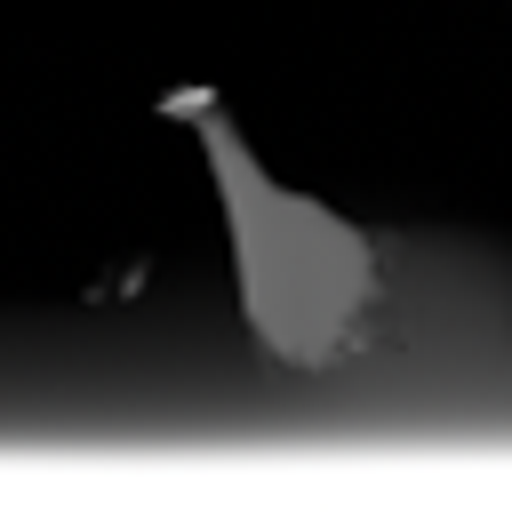} &  
\includegraphics{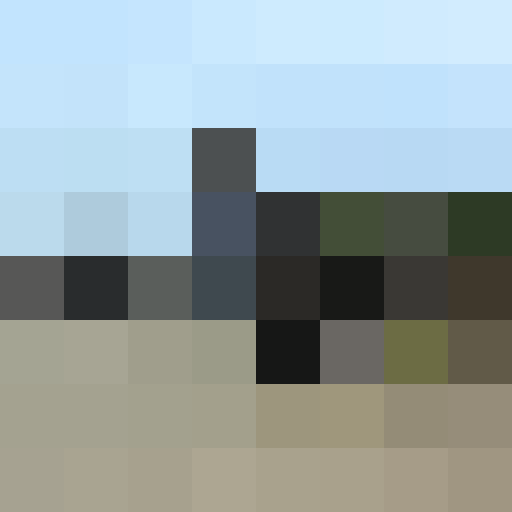} & \includegraphics{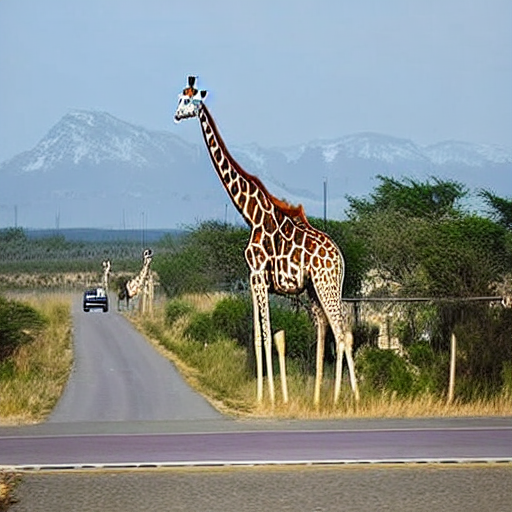} & \includegraphics{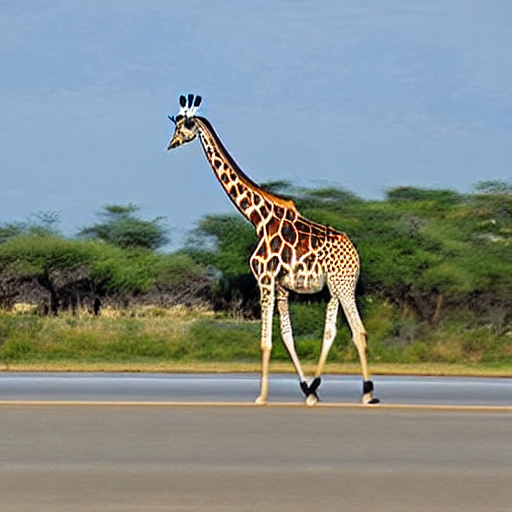} & \includegraphics{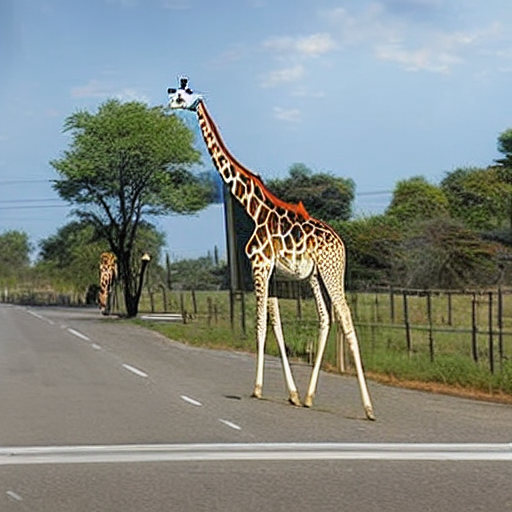} & 
\includegraphics{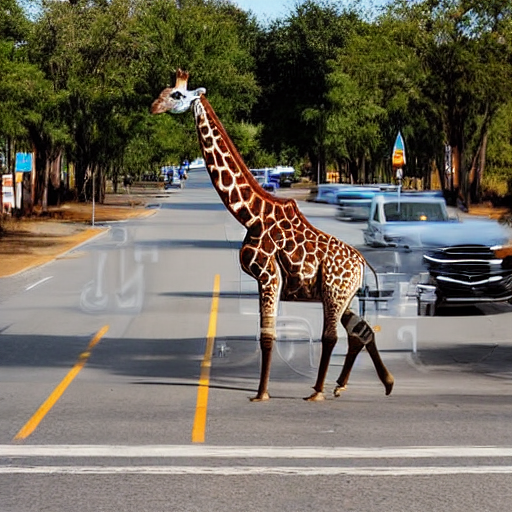} & \includegraphics{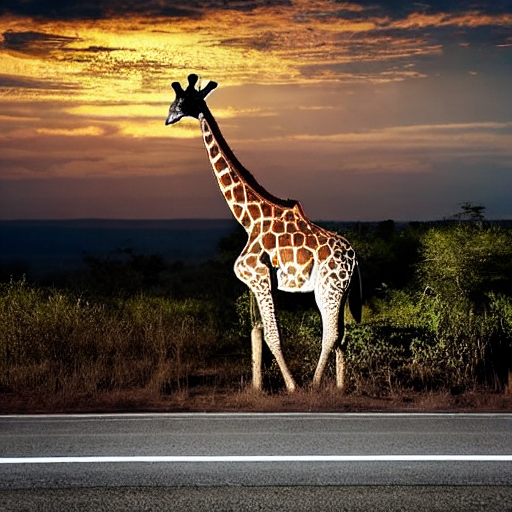} & \includegraphics{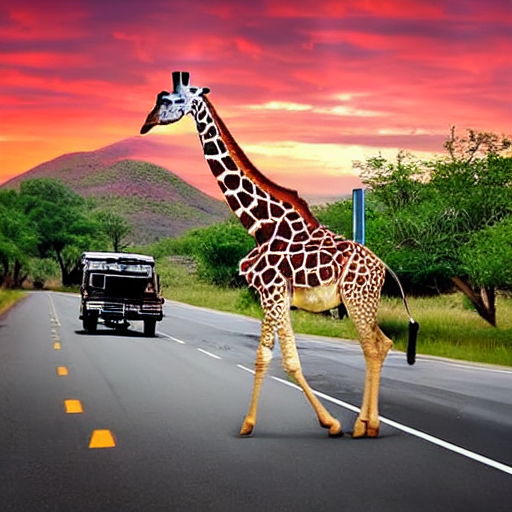} \\
\end{tabular}
\vspace{-2mm}
\caption{\textbf{Visual Decoding with \method.} Sample outputs demonstrate \method's ability to accurately decode the visual stimuli from fMRI. Our depth and color predictions from the R-PKM (\cref{subsec:method_pkm}) are in line with the pseudo ground-truth, despite the extreme complexity of the task.  \method reconstructions closely match the test images, and the rightmost samples demonstrate the benefit of color guidance.
}
\label{fig:ablation_color_significance}
}
\end{figure*}

\begin{table*}
    \centering
    \caption{\textbf{Effectiveness of R-VAC (Semantics Decipher) and R-PKM (Depth \& Color Decipher).} We conducted two sets of experiments using ground-truth (GT) or predicted (Pred) cues to reconstruct the visual stimuli, respectively.
    }
    \label{tab:ablation_pathways}
    \vspace{-2mm}
    \small
    \resizebox{0.88\linewidth}{!}{
    \begin{tabular}{c|l|cccc|cccc}
        \toprule
        ~ & \multirow{1}{*}{Reconstruction}   & \multicolumn{4}{c|}{Low-Level} & \multicolumn{4}{c}{High-Level}\\
        ~ & (\cref{subsec:meth_generation}) & PixCorr $\uparrow$ & SSIM $\uparrow$ & AlexNet(2) $\uparrow$ & AlexNet(5) $\uparrow$ & Inception $\uparrow$ & CLIP $\uparrow$  & EffNet-B $\downarrow$ & SwAV $\downarrow$\\
        \midrule
        \parbox[t]{2mm}{\multirow{3}{*}{\rotatebox[origin=c]{90}{GT}}} & semantics  $\{\texttt{S}\}$            & \underline{.244} & .272 & 96.68$\%$  & 97.39$\%$ & 87.82$\%$ & 92.45$\%$ & 1.00 & .415 \\ 
        ~ & +depth $\{\texttt{S}, \texttt{D}\}$             & .186 & \underline{.286} & \underline{99.58}$\%$  & \underline{99.78}$\%$ & \underline{98.78}$\%$ & \underline{98.09}$\%$ & \underline{.723} & \underline{.322} \\
        ~ & +color  $\{\texttt{S}, \texttt{D}, \texttt{C}\}$    & \textbf{.413} & \textbf{.366} & \textbf{99.99}$\%$ & \textbf{99.98}$\%$ & \textbf{99.19}$\%$ & \textbf{98.66}$\%$ & \textbf{.702} & \textbf{.278} \\
        \midrule
        \parbox[t]{2mm}{\multirow{3}{*}{\rotatebox[origin=c]{90}{Pred}}} &  semantics $\{\hat{\texttt{S}}\}$        & \underline{.194} & .278 & \underline{91.82}$\%$  & 92.57$\%$  & 93.11$\%$ & 91.24$\%$ & \underline{.645} & \textbf{.369} \\ 
        ~ & +depth $\{\hat{\texttt{S}}, \hat{\texttt{D}}\}$ & .083 & \underline{.282} & 88.07$\%$  & \underline{94.69}$\%$  & \underline{94.13}$\%$ & \textbf{96.05}$\%$ & .802 & .429 \\
        ~ & +color $\{\hat{\texttt{S}}, \hat{\texttt{D}}, \hat{\texttt{C}}\}$ & $\textbf{.288}$ & \textbf{.338} & \textbf{94.99}$\%$  & \textbf{97.50}$\%$  & \textbf{94.80}$\%$ & \underline{95.24}$\%$ & \textbf{.638} & \underline{.413} \\
         \bottomrule
    \end{tabular}
    }
\end{table*}

\condensedparagraph{Implementation Details.} One NVIDIA A100-SXM-80GB GPU is used in all experiments, including the training of fMRI $\mapsto$ Semantics encoder $\mathcal{E}_\texttt{fmri}$ and fMRI $\mapsto$ Depth \& Color encoder $\mathcal{E}$ and decoder $\mathcal{D}$. 
We use pretrained color and depth adapters from T2I-adapter~\cite{mou2023t2i} to extract guidance features from predicted spatial palettes and depth maps. These guidance features, along with predicted CLIP representations, are then input into the pretrained SD model for the purpose of image reconstruction.
The hyperparameters $\alpha$= 0.3, $\beta$ = 0.9, $\omega_c$ and $\omega_d$ are set as 1.0 unless otherwise mentioned.
For further details regarding the network architecture, please refer to the supplementary material.

\begin{figure*}[htbp]
\centering
\renewcommand{\arraystretch}{0.5}
\setkeys{Gin}{width=0.09\linewidth}
\setlength{\tabcolsep}{1.2pt}
\footnotesize
{
\begin{tabular}[t]{cccccccccc}

&Ground-Truth&\multicolumn{2}{c}{Predictions}&\multicolumn{6}{c}{$\omega_c$ / $\omega_d$}\\
\cmidrule(lr){2-2}\cmidrule(lr){3-4}\cmidrule(lr){5-10}
Test & ${\texttt{D}}$ & $\hat{\texttt{D}}$ & $\hat{\texttt{C}}$ & 1 / 1 (ours) & 1 / 0.6 & 1 / 0  & 0.6 / 0 & 0 / 1 & 0 / 0  \\
\includegraphics{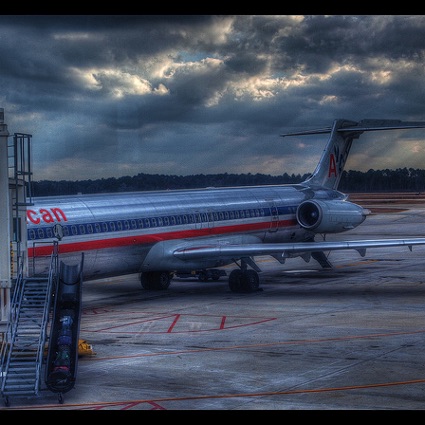} & \includegraphics{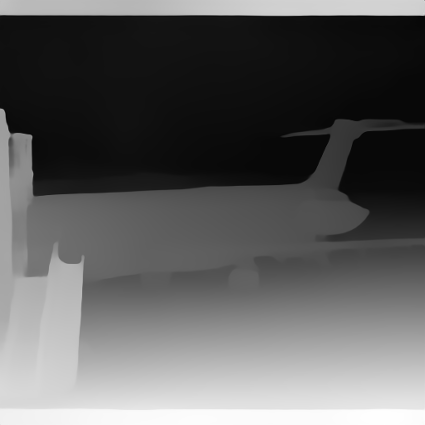} & 
\includegraphics{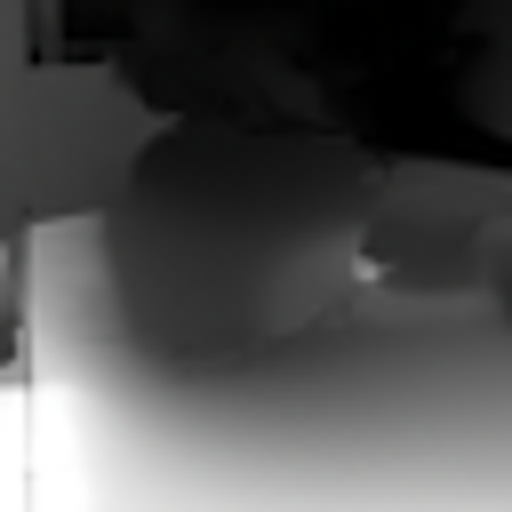} & 
\includegraphics{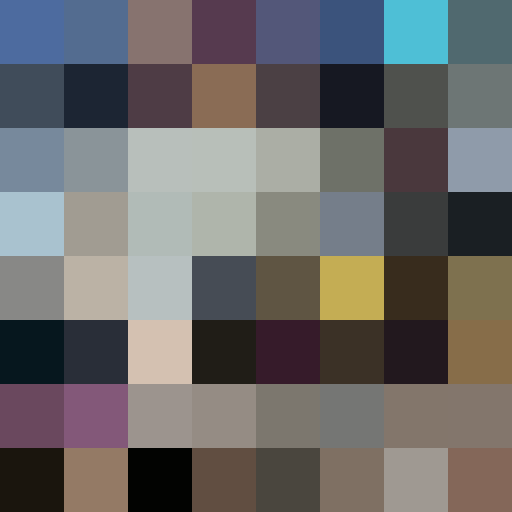} & 
\includegraphics{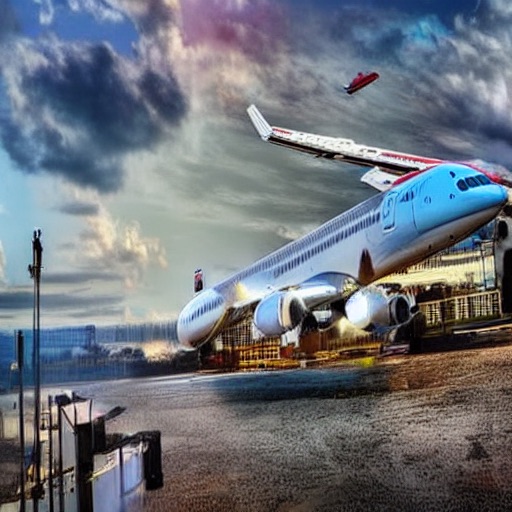} & \includegraphics{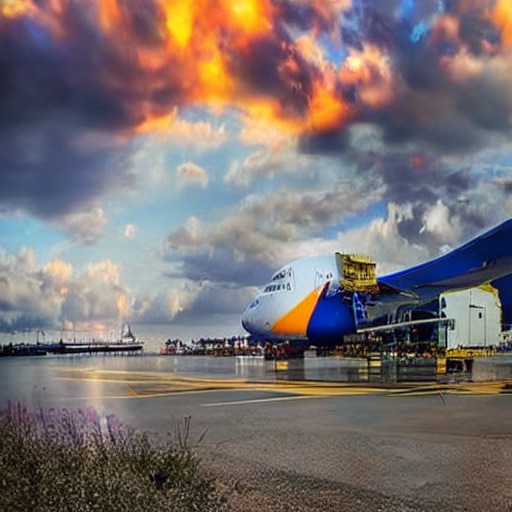} & 
\includegraphics{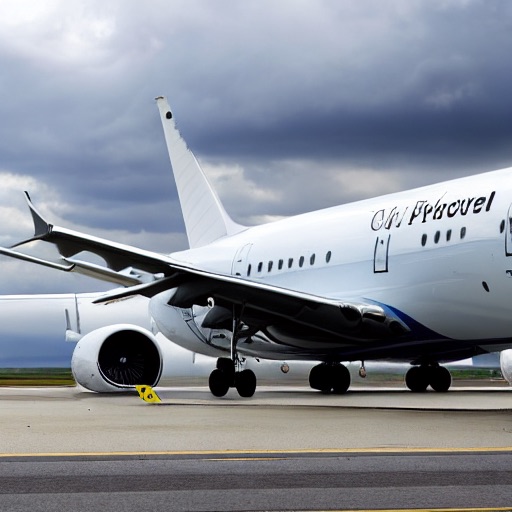} & 
\includegraphics{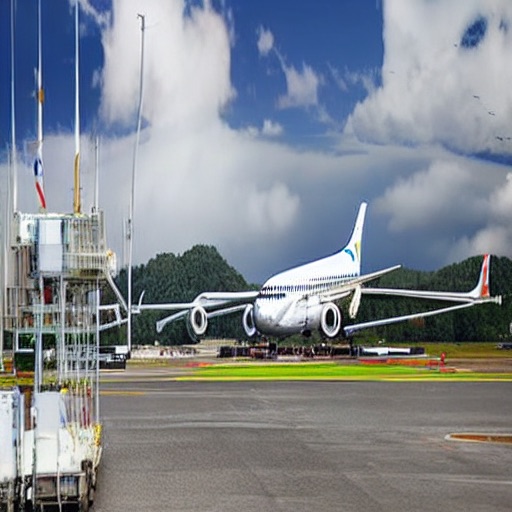} & 
\includegraphics{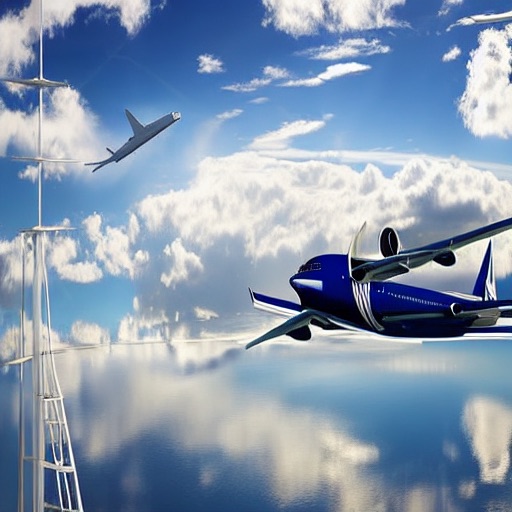} & 
\includegraphics{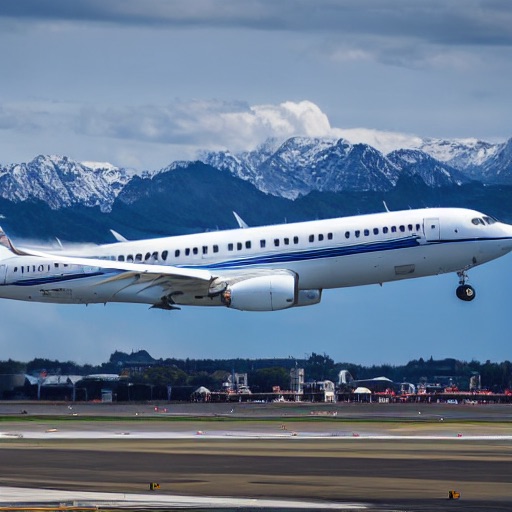} \\
\includegraphics{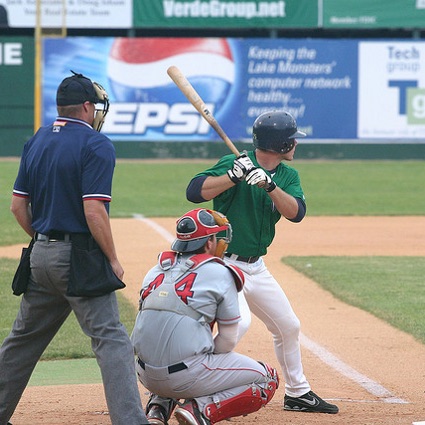} & \includegraphics{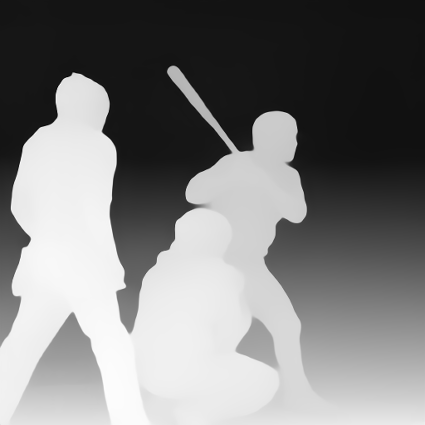} & 
\includegraphics{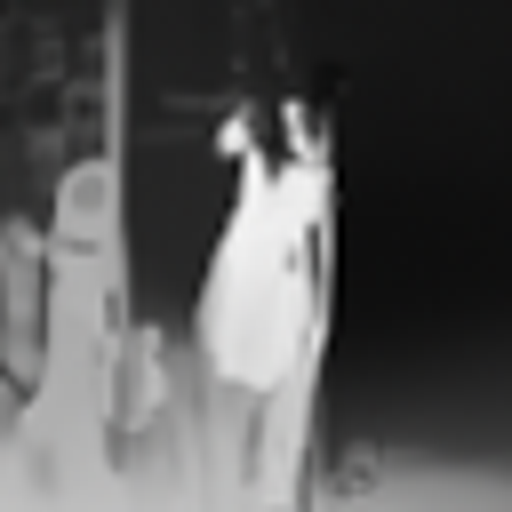} & 
\includegraphics{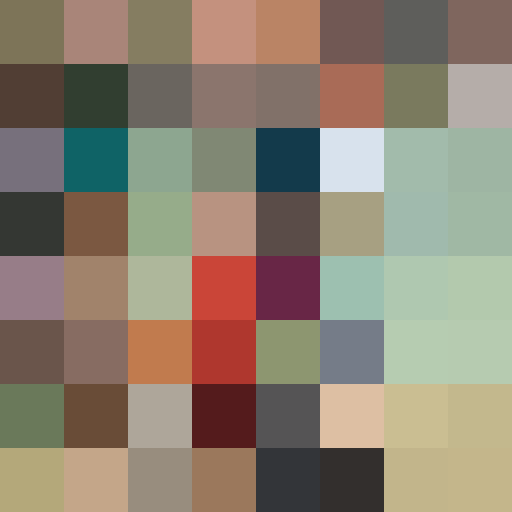} & 
\includegraphics{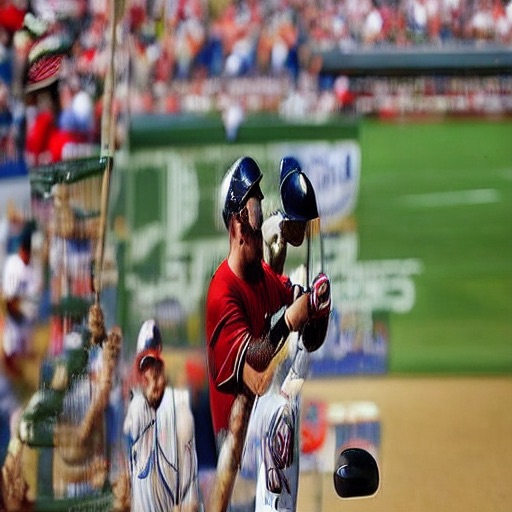} & \includegraphics{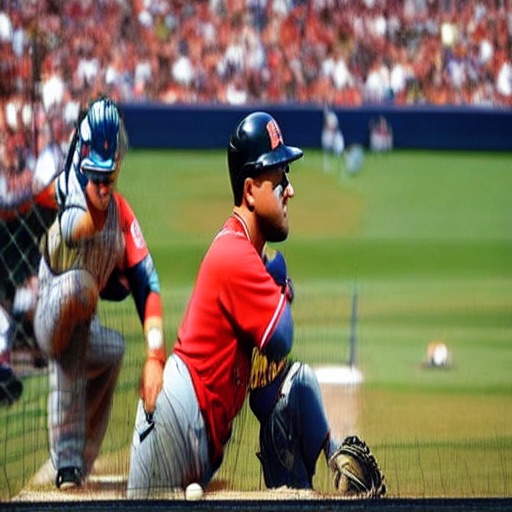} & 
\includegraphics{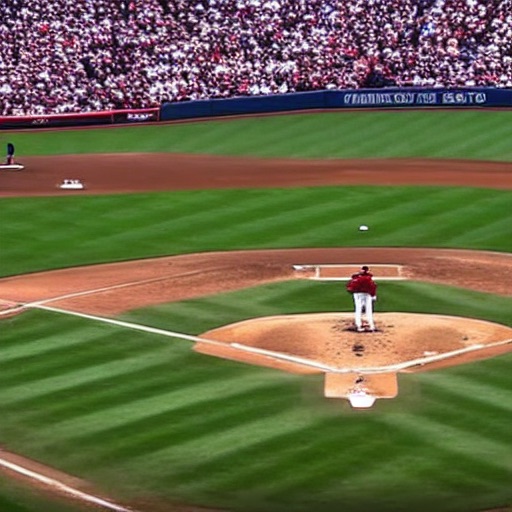} & 
\includegraphics{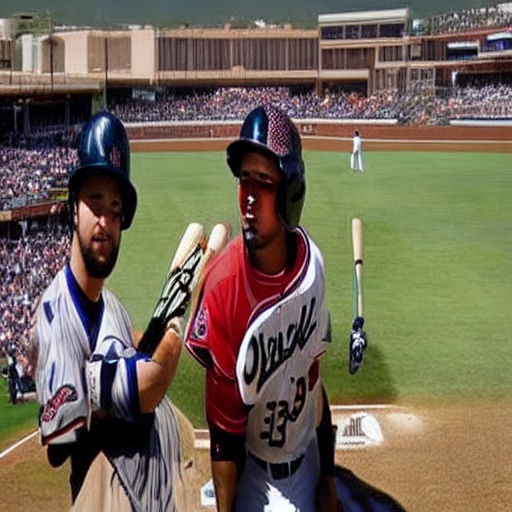} & 
\includegraphics{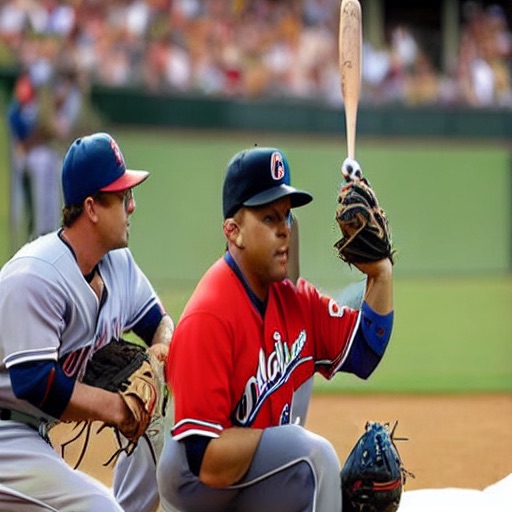} & 
\includegraphics{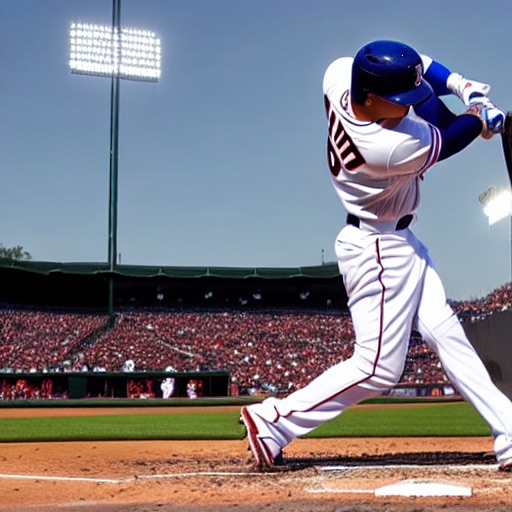} \\
\includegraphics{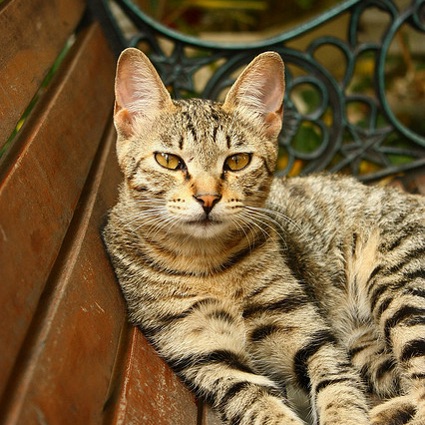} & \includegraphics{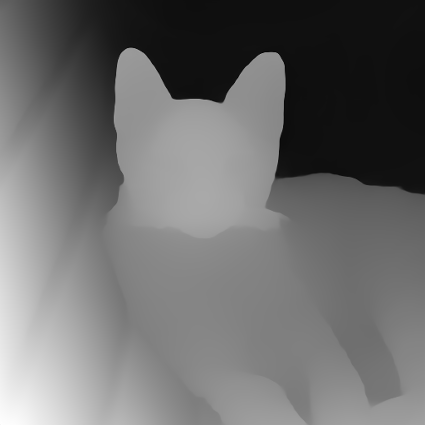} & 
\includegraphics{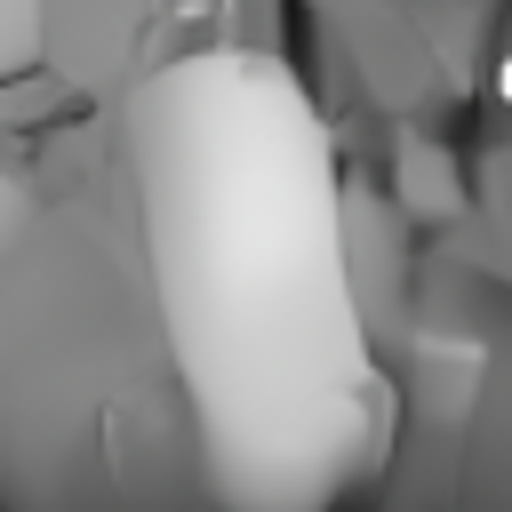} & 
\includegraphics{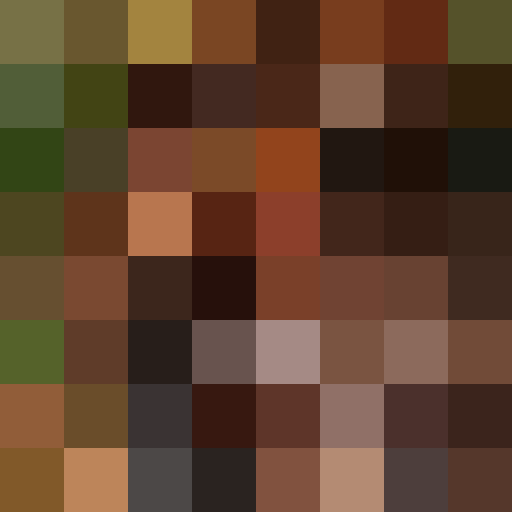} & 
\includegraphics{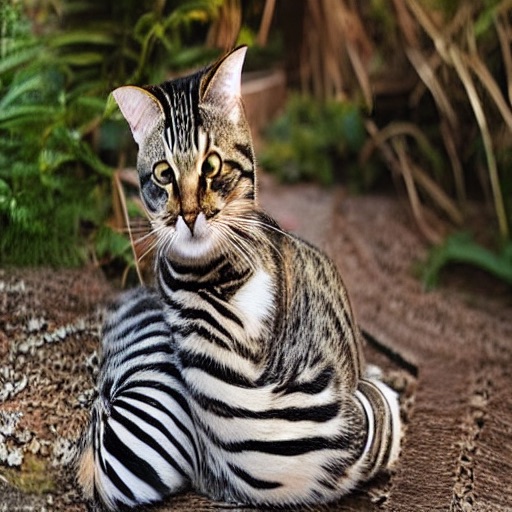} & \includegraphics{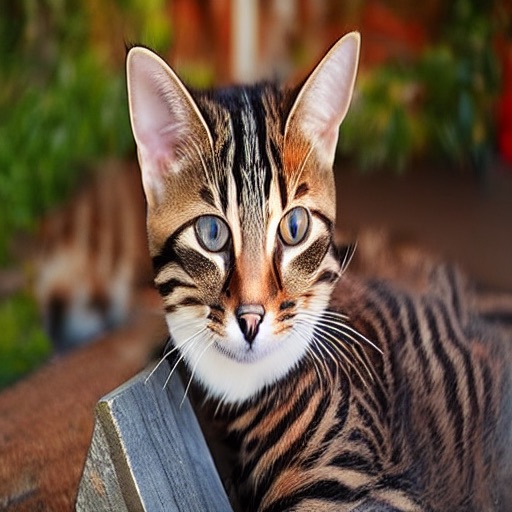} & 
\includegraphics{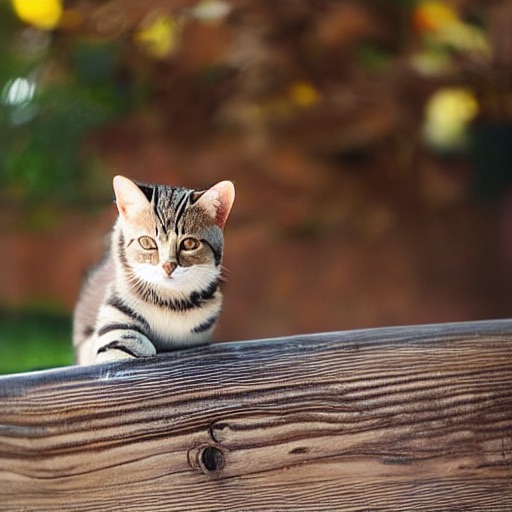} & 
\includegraphics{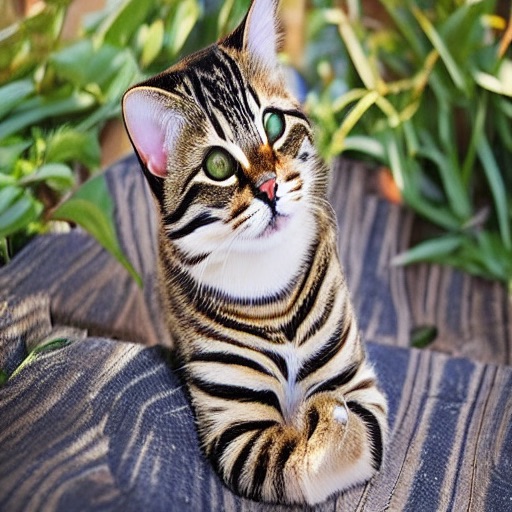} & 
\includegraphics{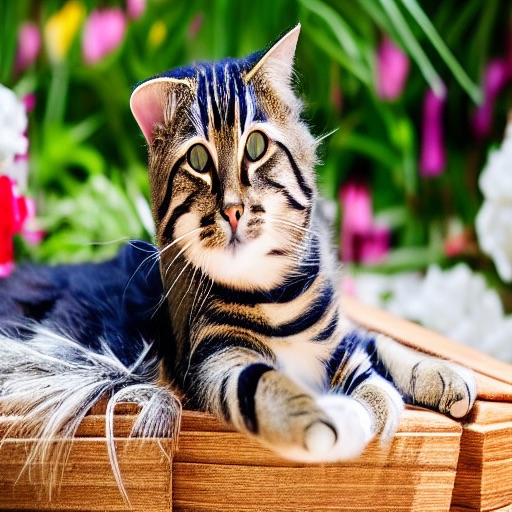} & 
\includegraphics{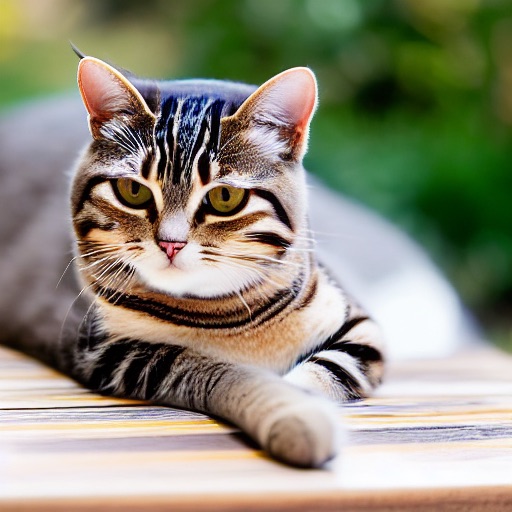} \\
\end{tabular}
\vspace{-2mm}
\caption{\textbf{Effect of the Composition Weight in GIR (\cref{subsec:meth_generation}).} The two weights $\omega_c$ and $\omega_d$ control the relative importance of corresponding features from three deciphered cues: semantics, color, and depth. 
When predicted depth or color fail to provide reliable guidance, we can manually tweak the weights to achieve satisfactory reconstructed results.
}
\label{fig:ablation_composition_weight}
}
\end{figure*}

\subsection{Experimental Results and Analysis
}
\label{sec:exp_results}

\vspace{-2.5mm}
\condensedparagraph{Visual Decoding.} Our method is compared with five state-of-the-art methods: Mind-Reader~\cite{lin2022mind}, Takagi~\etal~\cite{takagi2023high}, Gu~\etal~~\cite{gu2022decoding}, Brain-Diffuser~\cite{ozcelik2023brain}, and MindEye~\cite{scotti2023reconstructing}.
The quantitative visual decoding results are presented in \cref{tab:quantitative_nsd}, indicating a competitive performance.  
Our method, with explicit deciphering mechanism, appears to be more proficient at discerning scene structure and semantics, as evidenced by the favorable high-level metrics.
The qualitative results, depicted in~\cref{fig:recon_comparison}, align with the numerical findings, indicating that \method produces more realistic outcomes that maintains consistency with the viewed images in terms of semantics, appearance, and structure, compared to the other methods.
Striking DREAM outputs are the food plate (left, middle row) which accurately decodes the presence of vegetables and a tablespoon, and the baseball scene (right, middle row) which showcases the correct number of players (3) with poses similar to the test image.

Besides visual appearance, we wish to measure the consistency of depth and color in the decoded images with respect to the test images viewed by the subject.
We achieve this by measuring the variance in the estimated depth (and color palettes) of the test image and the reconstructed results from Brain-Diffuser~\cite{ozcelik2023brain}, MindEye~\cite{scotti2023reconstructing}, or \method.
Results presented in~\cref{tab:depth_color_estimation} indicate that our method yields images that align more consistently in color and depth with the visual stimuli than the other two methods.

\condensedparagraph{Cues Deciphering.} 
Our method decodes three cues from fMRI data: semantics, depth, and color. %
To assess semantics deciphered from R-VAC (\cref{subsec:method_vac}), we simply refer to the CLIP metric of~\cref{tab:quantitative_nsd} which quantifies CLIP embeddings distances with the test image. From the aforementioned table, \method is at least 1.1\% better than others. 
\cref{fig:ablation_color_significance} shows examples of the depth ($\hat{\texttt{D}}$) and color ($\hat{\texttt{C}}$) deciphered by R-PKM (\cref{subsec:method_pkm}). 
While accurate depth is beneficial for image reconstruction, faithfully recovering the original depth from fMRI is nearly impossible due to the information loss in capturing the brain activities~\cite{allen2022massive}.
Still, coarse depth is sufficient in most cases to guide the scene structure and object position such as determining the location of an airplane or the orientation of a bird standing on a branch.
This is intuitively understood from the bottom row of \cref{fig:ablation_color_significance}, where our coarse depth ($\hat{\texttt{D}}$) leaves no doubt on the giraffe's location and orientation. 
Interestingly, despite not precisely preserving the local color, the estimated color palettes ($\hat{\texttt{C}}$) provide a reliable constraint and guidance on the overall scene appearance.
This is further demonstrated in last three columns of \cref{fig:ablation_color_significance} by removing color guidance which, despite appealing visuals, proves to produce images drastically differing from the test image.

\section{Ablation Study}
\label{sec:ablation}

Here we present ablation studies while discussing first the effect of using color, depth and semantics as guidance for image reconstruction thanks to our reversed pathways (\ie, R-VAC and R-PKM).

\condensedparagraph{Effect of Color Palettes.} As highlighted earlier, the deciphered color guidance noticeably enhances the visual quality of the reconstructed images in \cref{fig:ablation_color_significance}.
We further quantify the color's significance through two additional sets of experiments detailed in \cref{tab:ablation_pathways}, where the reconstruction uses: 1) ground-truth (GT) depth and caption with or without color, and 2) fMRI-predicted (Pred) depth and semantic embedding with or without the predicted color. The results using ground-truth cues serve as a proxy.
The generated results exhibit improved color consistency and enhanced quantitative performance across the board, underscoring the importance of using color for visual decoding.

\condensedparagraph{Effect of Depth and Semantics.}
\cref{tab:ablation_pathways} presents results where we ablate the use of depth and semantics. 
Comparison of GT and predicted semantics (\ie,~$\{\texttt{S}\}$ and $\{\hat{\texttt{S}}\}$) suggests that the fMRI embedding effectively incorporates high-level semantic cues into the final images. 
The overall image quality can be further improved by integrating either ground truth depth (represented as $\{\texttt{S}, \texttt{D}\}$) or predicted depth ($\{\hat{\texttt{S}}, \hat{\texttt{D}}\}$) combined with color (denoted as $\{\texttt{S}, \texttt{D}, \texttt{C}\}$ and $\{\hat{\texttt{S}}, \hat{\texttt{D}}, \hat{\texttt{C}}\}$).
Introducing color cues not only bolsters the structural information but also strengthens the semantics, possibly because it compensates for the color information absent in the predicted fMRI embedding.
Of note, all metrics (except for PixCorr) improve smoothly with more GT guidance. Yet, the impact of predicted cues varies across metrics, highlighting intriguing research avenues and emphasizing the need for more reliable measures.

\condensedparagraph{Effect of Data Scarcity Strategies.} 
We ablate the two strategies introduced to fight fMRI data scarcity: data augmentation (DA) in R-VAC and the third-stage decoder training (S3) which allows R-PKM to use additional RGBD data without fMRI. The results shown in the two bottom rows of~\cref{tab:quantitative_nsd} demonstrate that the two data augmentation strategies address the limited availability of fMRI data and subsequently bolster the model's generalization capability.

\condensedparagraph{Effect of Weighted Guidance.} The features inputted into SD are formulated as $\texttt{S} + \omega_c \mathcal{R}_c(\texttt{C}) + \omega_d \mathcal{R}_d(\texttt{D})$, where two weights $\omega_c$ and $\omega_d$ from \cref{eq:im_guided} control the relative importance of the deciphered cues and play a crucial role in the final image quality and alignment with the cues. In \method, $\omega_c$ and $\omega_d$ are set to 1.0 showing no preference of color guidance over depth guidance.  Still,~\cref{fig:ablation_composition_weight} shows that in some instances the predicted components fail to provide dependable guidance on structure and appearance, thus compromising results.  There is also empirical evidence that indicates the T2I-adapter slightly underperforms when compared to ControlNet. The performance of the T2I-adapter further diminishes when multiple conditions are used, as opposed to just one. Taking both factors into account, there are instances where manual adjustments to the weighting parameters become necessary to achieve images of the desired quality, semantics, structure, and appearance.

\section{Conclusion}
\label{sec:conclusion}

This paper presents \method, a visual decoding method founded on principles of human perception. We design reverse pathways that mirror the forward pathways from visual stimuli to fMRI recordings. These pathways specialize in deciphering semantics, color, and depth cues from fMRI data and then use these predicted cues as guidance to reconstruct visual stimuli. Experiments demonstrate that our method surpasses current state-of-the-art models in terms of consistency in appearance, structure, and semantics.

\condensedparagraph{Acknowledgements} This work was supported by the Engineering and Physical Sciences Research Council [grant number EP/W523835/1] and a UKRI Future Leaders Fellowship [grant number G104084].

\clearpage

\setcounter{figure}{0}
\setcounter{table}{0}
\setcounter{equation}{0}

\begin{appendices}
\label{appendices}

In the following, we provide more details and discussion on the background knowledge, experiments, and more results of our method.
We first provide more details on the NSD neuroimaging dataset in~\cref{sec:supmat_datasets} and extend background knowledge of the Human Visual System in~\cref{sec:supmat_hvs}, which together shed light on our design choices. We then detail T2I-Adapter in~\cref{sec:supmat_base_model}.
\cref{sec:supmat_implementation} provides thorough implementation of \method, including architectures, representations and metrics. Finally, in~\cref{sec:supmat_more_results} we further demonstrate the ability of our method with new results of cues deciphering, reconstruction, and reconstruction across subjects.

\section{NSD Dataset}
\label{sec:supmat_datasets}

The Natural Scenes Dataset (NSD)~\cite{allen2022massive} is currently the largest publicly available fMRI dataset. It features in-depth recordings of brain activities from 8 participants (subjects) who passively viewed images for up to 40 hours in an MRI machine.
Each image was shown for three seconds and repeated three times over 30-40 scanning sessions, amounting to 22,000-30,000 fMRI response trials per participant. These viewed  natural scene images are sourced from the Common Objects in Context (COCO) dataset~\cite{lin2014microsoft}, enabling the utilization of the original COCO captions for training.

The fMRI-to-image reconstruction studies that used NSD~\cite{ozcelik2022reconstruction,gu2022decoding,takagi2023high} typically follow the same procedure: training individual-subject models for the four participants who finished all scanning sessions (participants 1, 2, 5, and 7), and employing a test set that corresponds to the common 1,000 images shown to each participant.
For each participant, the training set has 8,859 images and 24,980 fMRI tests (as each image being tested up to 3 times). Another 982 images and 2,770 fMRI trials are common across the four individuals.
We use the preprocessed fMRI voxels in a 1.8-mm native volume space that corresponds to the ``nsdgeneral'' brain region. This region is described by the NSD authors as the subset of voxels in the posterior cortex that are most responsive to the presented visual stimuli.
For fMRI data spanning multiple trials, we calculate the average response as in prior research~\cite{lu2023minddiffuser}.
\cref{tab:supmat_nsd_dataset} details the characteristics of the NSD dataset and the region of interests (ROIs) included in the fMRI data.

\begin{table}[thbp]
\centering
\caption{\textbf{Details of the NSD dataset.} %
}
\label{tab:supmat_nsd_dataset}
 \resizebox{0.9\linewidth}{!}
 {
    \begin{tabular}{ccccc}
        \toprule
        Training & Test & ROIs & Subject ID & Dimensions \\ 
        \midrule
        \multirow{4}{*}{8859} & \multirow{4}{*}{982} & \multirow{4}{*}{\begin{tabular}[c]{@{}c@{}}V1, V2, V3, hV4,\\ VO, PHC, MT,\\ MST, LO, IPS\end{tabular}} & sub01 & 15,724  \\ 
                  &              &      & sub02      & 14,278   \\ 
                  &              &      & sub05      & 13,039   \\ 
                  &              &      & sub07      & 12,682   \\ 
        \bottomrule
    \end{tabular}
 }
\end{table}

\section{Detailed Human Visual System}
\label{sec:supmat_hvs}

Our approach aims to decode semantics, color, and depth from fMRI data, thus inherently bounded by the ability of fMRI data to capture the ad hoc brain activities. It is crucial to ascertain whether fMRI captures the alterations in the respective human brain regions responsible for processing the visual information. 
Here, we provide a comprehensive examination of the specific brain regions in the human visual system recorded by the fMRI data.

The flow of visual information~\cite{brodal2004central} in neuroscience is presented as follows. \cref{fig:supmat_hvs} presents a comprehensive depiction of the functional anatomy of the visual perception.
Sensory input originating from the Retina travels through the LGN in the thalamus and then reaches the Visual Cortex.
\textbf{Retina} is a layer within the eye comprised of photoreceptor and glial cells. These cells capture incoming photons and convert them into electrical and chemical signals, which are then relayed to the brain, resulting in visual perception.
Different types of information are processed through the parvocellular and magnocellular pathways, details of which are elaborated in the main paper.
\textbf{LGN} then channels the conveyed visual information into the~\textbf{Visual Cortex}, where it diverges into two streams in~\textbf{Visual Association Cortex} (VAC) for undertaking intricate processing of high-level semantic contents from the visual image.

\begin{figure}
\centering
\includegraphics[width=0.95\linewidth]{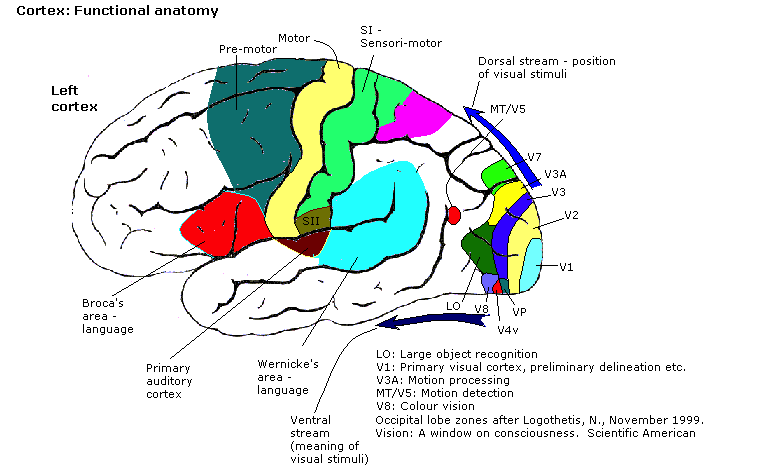}
\caption{\textbf{Functional Anatomy of Cortex.} The functional localization in the human brain is based on findings from functional brain imaging, which link various anatomical regions of the brain to their associated functions.\\ %
\textit{Source}: \href{https://upload.wikimedia.org/wikipedia/commons/d/db/Constudproc.png}{\color{black}{Wikimedia Commons}}. This image is licensed under the Creative Commons Attribution-Share Alike 3.0 Unported license.
}
\label{fig:supmat_hvs}
\end{figure}

The Visual Cortex, also known as visual area 1 (V1), serves as the initial entry point for visual perception within the cortex. Visual information flows here first before being relayed to other regions.
VAC comprises multiple regions surrounding the visual cortex, including V2, V3, V4, and V5 (also known as the middle temporal area, MT).
V1 transmits information into two primary streams: the ventral stream and the dorsal stream. 
\begin{itemize}
    \item The ventral stream (black arrow) begins with V1, goes through V2 and V4, and to the inferior temporal cortex (IT cortex). The ventral stream
    is responsible for the ``meaning'' of the visual stimuli, such as object recognition and identification. 
    \item The dorsal stream  (blue arrow) begins with V1, goes through visual area V2, then to the dorsomedial area (DM/V6) and medial temporal area (MT/V5) and to the posterior parietal cortex. The dorsal stream 
    is engaged in analyzing information associated with ``position'', particularly the spatial properties of objects. 
\end{itemize}

After juxtaposing the explanations illustrated in~\cref{fig:supmat_hvs} with the collected information demonstrated in~\cref{tab:supmat_nsd_dataset}, it becomes apparent that the changes occurring in brain regions linked to the processing of semantics, color, and depth are indeed present within the fMRI data. This observation emphasizes the capability to extract the intended information from the provided fMRI recordings.

\section{T2I-Adapter}
\label{sec:supmat_base_model}

T2I-Adapter~\cite{mou2023t2i} and ControlNet~\cite{zhang2023adding} learn versatile modality-specific encoders to improve the control ability of text-to-image SD model~\cite{rombach2022high}. 
These encoders extract guidance features from various conditions $\mathbf{y}$ (\eg sketch, semantic label, and depth). They aim to align external control with internal knowledge in SD, thereby enhancing the precision of control over the generated output.
Each encoder $\mathcal{R}$ produces $n$ hierarchical feature maps $\text{F}^i_\mathcal{R}$ from the primitive condition $\mathbf{y}$. Then each $\text{F}^i_\mathcal{R}$ is added with the corresponding intermediate feature $\text{F}^i_{\text{SD}}$ in the denoising U-Net encoder:
\begin{equation}
\begin{aligned} 
\text{F}_{\mathcal{R}} &=\mathcal{R}\left(\mathbf{y}\right), \\
\hat{\text{F}}_{\text{SD}}^i & =\text{F}_{\text{SD}}^i+\text{F}_{\mathcal{R}}^i, \quad i \in\{1,2,\cdots,n\}.
\end{aligned}
\end{equation}

T2I-Adapter consists of a pretrained SD model and several adapters. These adapters are used to extract guidance features from various conditions. The pretrained SD model is then utilized to generate images based on both the input text features and the additional guidance features. 
The CoAdapter mode becomes available when multiple adapters are involved, and a composer processes features from these adapters before they are further fed into the SD.
Given the deciphered semantics, color, and depth information from fMRI, we can reconstruct the final images using the color and depth adapters in conjunction with SD.

\section{Implementation Details}
\label{sec:supmat_implementation}

\subsection{Network Architectures} 
\label{subsec:supmat_network_architectures}

The fMRI $\mapsto$ Semantics encoder $\mathcal{E}_\texttt{fmri}$ maps fMRI voxels to the shared CLIP latent space~\cite{radford2021learning} to decipher semantics. The network architecture includes a linear layer followed by multiple residual blocks, a linear projector, and a final MLP projector, akin to previous research~\cite{chen2023seeing,scotti2023reconstructing}.
The learned embedding is with a feature dimension of 77 $\times$ 768, where 77 denotes the maximum token length and 768 represents the encoding dimension of each token.
It is then fed into the pretrained Stable Diffusion~\cite{rombach2022high} to inject semantic information into the final reconstructed images.

The fMRI $\mapsto$ Depth \& Color encoder $\mathcal{E}$ and decoder $\mathcal{D}$ decipher depth and color information from the fMRI data.
Given that spatial palettes are generated by first downsampling (with bicubic interpolation) an image and then upsampling (with nearest interpolation) it back to its original resolution, the primary objective of the encoder $\mathcal{E}$ and the decoder $\mathcal{D}$ shifts towards predicting RGBD images from fMRI data.
The architecture of $\mathcal{E}$ and $\mathcal{D}$ is built on top of~\cite{gaziv2022self}, with inspirations drawn from VDVAE~\cite{child2020very}.

\subsection{Representation of Semantics, Color and Depth}
\label{subsec:supmat_representations}

This section serves as an introduction to the possible choices of representations for semantics, color, and depth. 
We currently use CLIP embedding, depth map~\cite{ranftl2020towards}, and spatial color palette~\cite{mou2023t2i} to facilitate subsequent processing of T2I-Adapter~\cite{mou2023t2i} in conjunction with a pretrained Stable Diffusion~\cite{rombach2022high} for image reconstruction from deciphered cues. However, there are other possibilities that can be utilized within our framework.

\condensedparagraph{Semantics.} The Stable Diffusion utilizes a frozen CLIP ViT-L/14 text encoder to condition the model on text prompts.
It is with a feature space dimension of 77 $\times$ 768, where 77 denotes the maximum token length and 768 represents the encoding dimension of each token. 
The CLIP ViT-L/14 image encoder is with a feature space dimension of 257 $\times$ 768.
We maps flattened voxels to an intermediate space of size 77 $\times$ 768, corresponding to the last hidden layer of CLIP ViT/L-14. 
The learned embeddings inject semantic information into the reconstructed images.

\begin{figure}[t!]
\center
\setlength\tabcolsep{0pt}
{
\renewcommand{\arraystretch}{0.5}
\setkeys{Gin}{width=0.33\linewidth}
\small
\setlength{\tabcolsep}{3.5pt}{
\resizebox{0.85\linewidth}{!}
{
\begin{tabular}{ccc} 
    Test image ($\texttt{I}$) & GT Depth ($\texttt{D}$) & GT Color ($\texttt{C}$)  \\
    \includegraphics{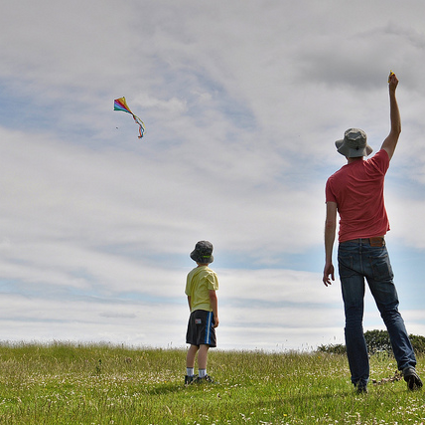} &
    \includegraphics{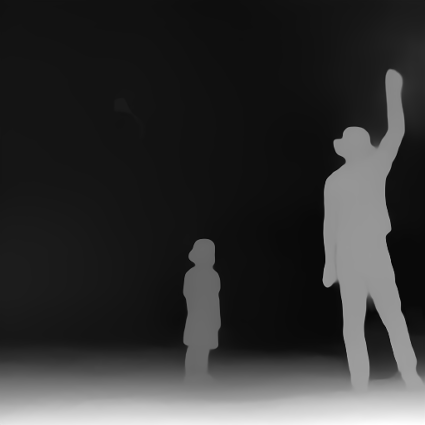} &
    \includegraphics{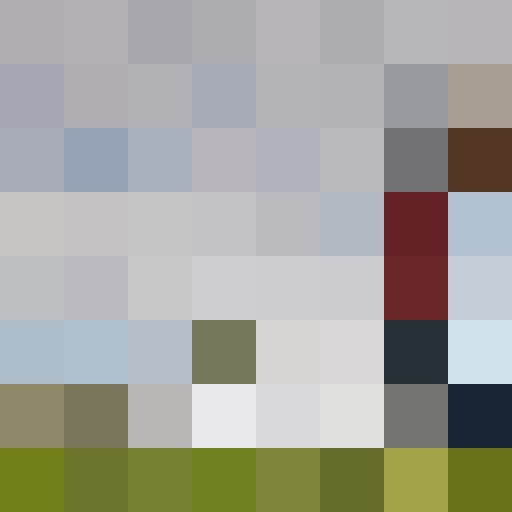} \\
    \includegraphics{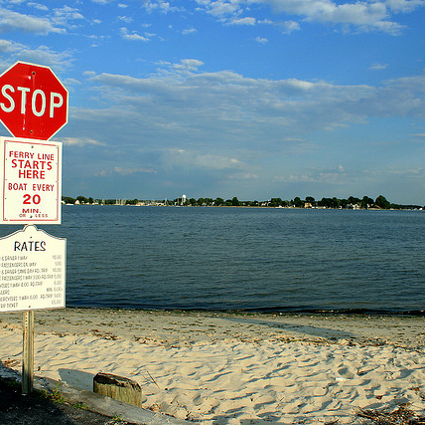} &
    \includegraphics{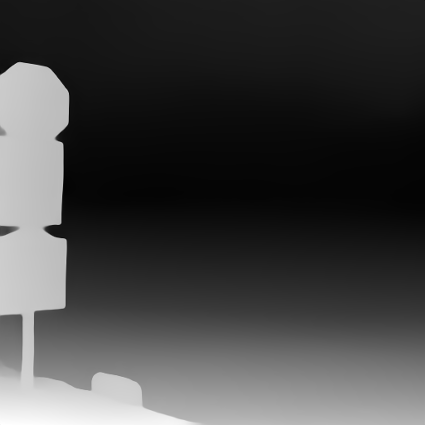} &
    \includegraphics{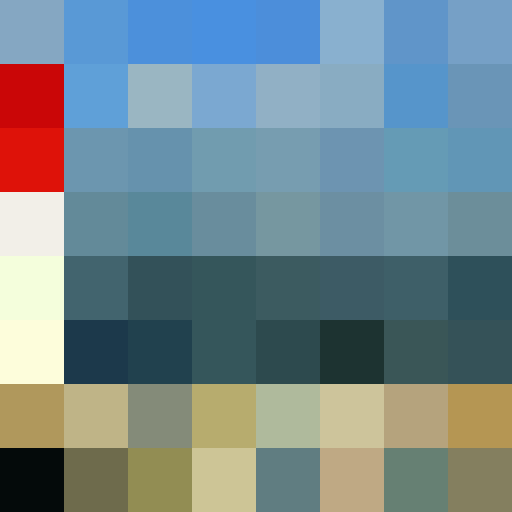} 
\end{tabular}
}
}
}
\vspace{-2mm}
\caption{\textbf{Depth and Color Representations.} We present pseudo ground truth samples of Depth (MiDaS prediction~\cite{ranftl2020towards}) and Color ($\times{}64$ downsampling of the test image) for a NSD input image.}
\label{fig:supmat_data_example}
\end{figure}

\condensedparagraph{Depth.} 
We select depth as the structural guidance for two main reasons: alignment with the human visual system, and better performance demonstrated in our preliminary experiments.
Following prior research~\cite{zhang2023adding,mou2023t2i}, we use the MiDaS predictions~\cite{ranftl2020towards} as the surrogate ground truth depth maps, which are visualized in~\cref{fig:supmat_data_example}. 

\condensedparagraph{Color.} There are many representations that can provide the color information, such as histogram and probabilistic palette~\cite{wang2022palgan,le2021semantic}
However, ControlNet~\cite{zhang2023adding} and T2I-Adapter~\cite{mou2023t2i} only accept spatial inputs, which leaves no alternative but to utilize the \textit{spatial color palettes} as the color representation. In practice, spatial color palettes resemble coarse resolution images, as seen in~\cref{fig:supmat_data_example}, and are generated by first $\times$64 downsampling (with bicubic interpolation) an image and then upsampling (with nearest interpolation) it back to its original resolution.

During the image reconstruction phase, the spatial palettes contribute the color and appearance information to the final images. 
These spatial palettes are derived from the image estimated by the RGBD decoder in R-PKM. We refer to the images produced at this stage as the ``initial guessed image'' to differentiate them from the final reconstruction.
The initial guessed image offers color cues but it also contains inaccuracies. By employing a $\times64$ downsampling, we can effectively extract necessary color details from this image while minimizing the side effects of inaccuracies.

\condensedparagraph{Other Guidance.} In the realm of visual decoding with pretrained diffusion models~\cite{zhang2023adding,mou2023t2i,xu2022versatile}, any guidance available in these models can be harnessed to fill in gaps of missing information, thereby enhancing performance.
This spatial guidance includes representations such as sketch~\cite{su2021pixel}, Canny edge detection, HED (Holistically-Nested Edge Detection)~\cite{xie2015holistically}, and semantic segmentation maps~\cite{caesar2018coco}.
These alternatives could potentially serve as the intermediate representations for the reverse pathways in our method.
HED and Canny are edge detectors, which provide object boundaries within images. 
However, during our preliminary experiments, both methods were shown to face challenges in providing reliable edges for all images.
Sketches encounter similar difficulties in providing reliable guidance.
The semantic segmentation map provides both structural and semantic cues. However, it overlaps in function with CLIP semantics and depth maps, and leads to diminished performance gain on top of the other two representations.

\begin{figure*}[t!]
\centering
\setlength\tabcolsep{0pt}
\renewcommand{\arraystretch}{0.8}
\setkeys{Gin}{width=0.95\linewidth}
\setlength{\tabcolsep}{0pt}
\footnotesize
\resizebox{0.95\linewidth}{!}
{
\begin{tabular}{>{\centering\arraybackslash}p{0.1\linewidth}>{\centering\arraybackslash}p{0.1\linewidth}>{\centering\arraybackslash}p{0.1\linewidth}>{\centering\arraybackslash}p{0.1\linewidth}>{\centering\arraybackslash}p{0.1\linewidth}p{0.05\linewidth}>{\centering\arraybackslash}p{0.1\linewidth}>{\centering\arraybackslash}p{0.1\linewidth}>{\centering\arraybackslash}p{0.1\linewidth}>{\centering\arraybackslash}p{0.1\linewidth}>{\centering\arraybackslash}p{0.1\linewidth}} %
Test image&GT Depth&\multicolumn{3}{c}{Predictions}&&Test image&GT Depth&\multicolumn{3}{c}{Predictions}\\
\cmidrule(lr){2-2}\cmidrule(lr){3-5}\cmidrule(lr){8-8}\cmidrule(lr){9-11}
$\texttt{I}$ & $\texttt{D}$ & $\hat{\texttt{D}}$ & $\hat{\texttt{I}}_0$ & $\hat{\texttt{C}}$ & & $\texttt{I}$ & $\texttt{D}$ & $\hat{\texttt{D}}$ & $\hat{\texttt{I}}_0$ & $\hat{\texttt{C}}$\\
\multicolumn{5}{c}{\includegraphics[width=0.5\linewidth]{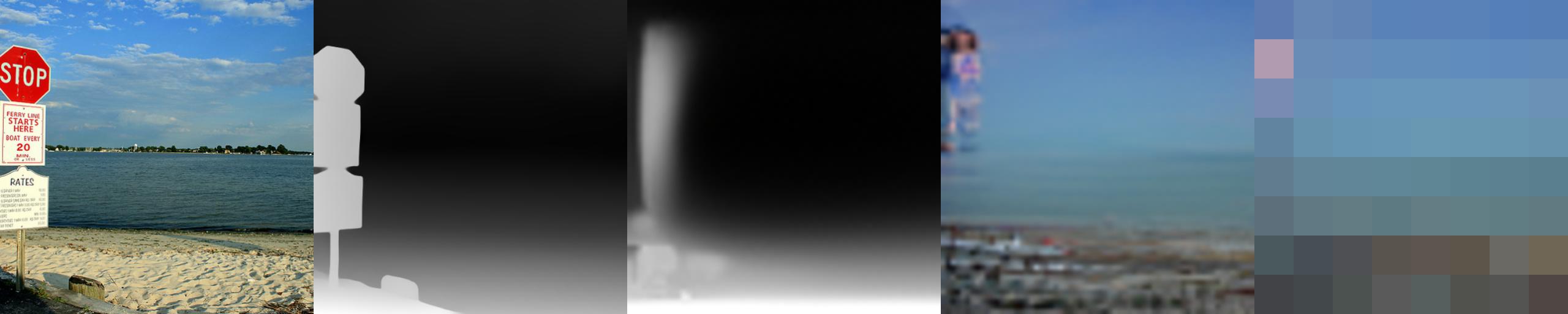}} & & \multicolumn{5}{c}{\includegraphics[width=0.5\linewidth]{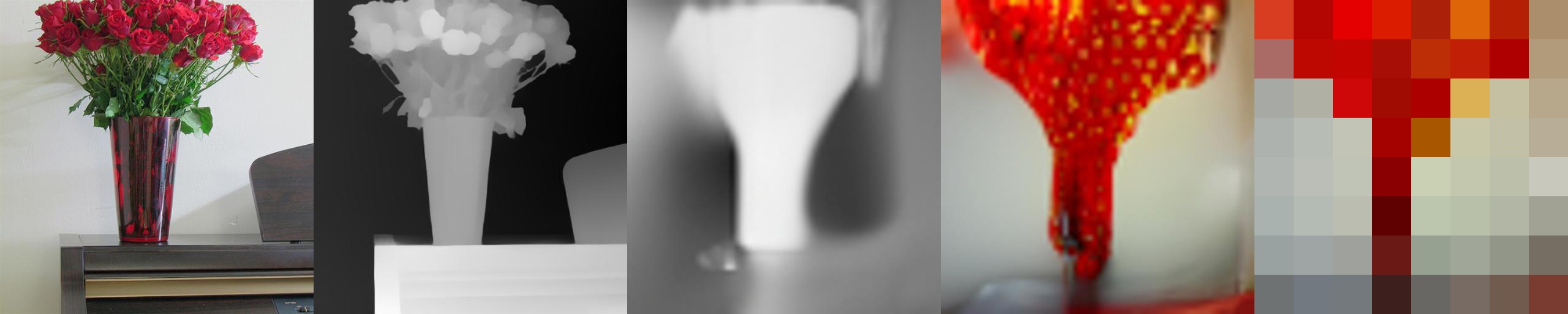}} \\
\multicolumn{5}{c}{\includegraphics[width=0.5\linewidth]{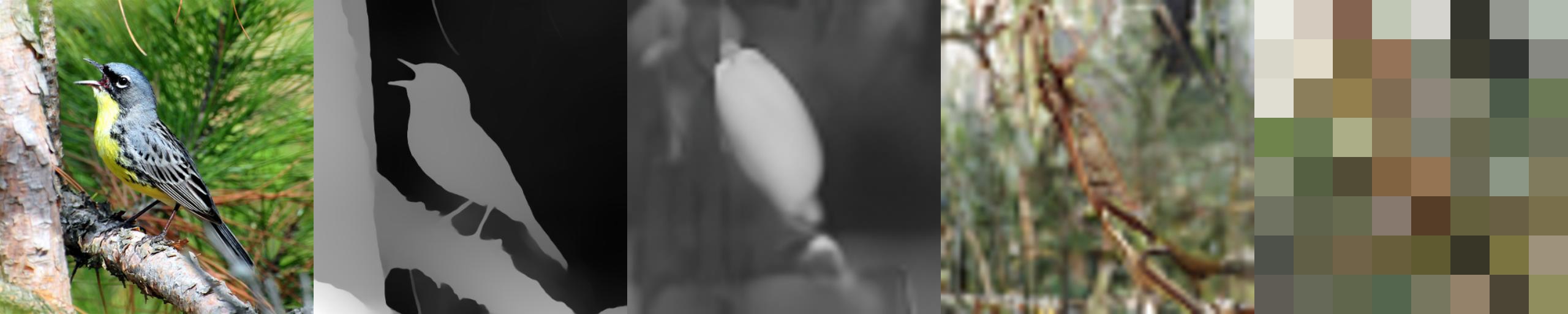}} & & \multicolumn{5}{c}{\includegraphics[width=0.5\linewidth]{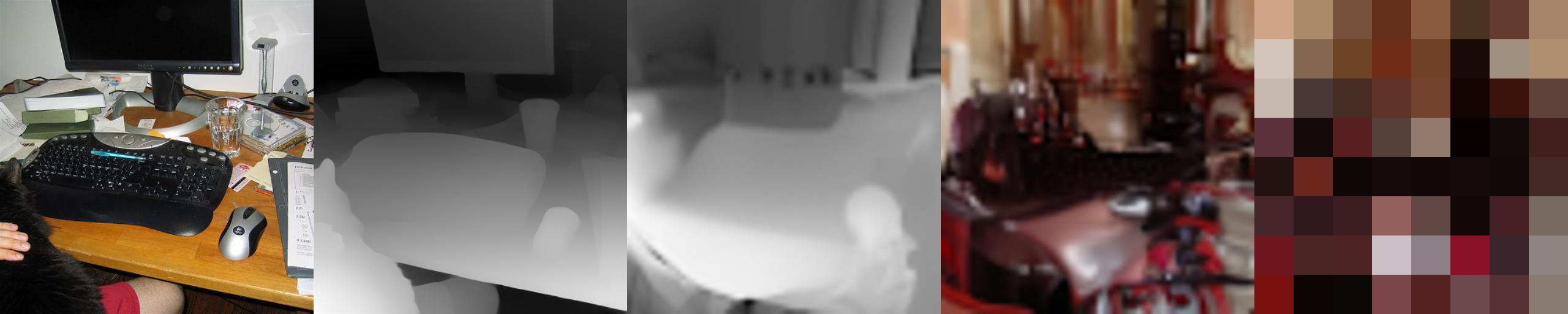}} \\
\multicolumn{5}{c}{\includegraphics[width=0.5\linewidth]{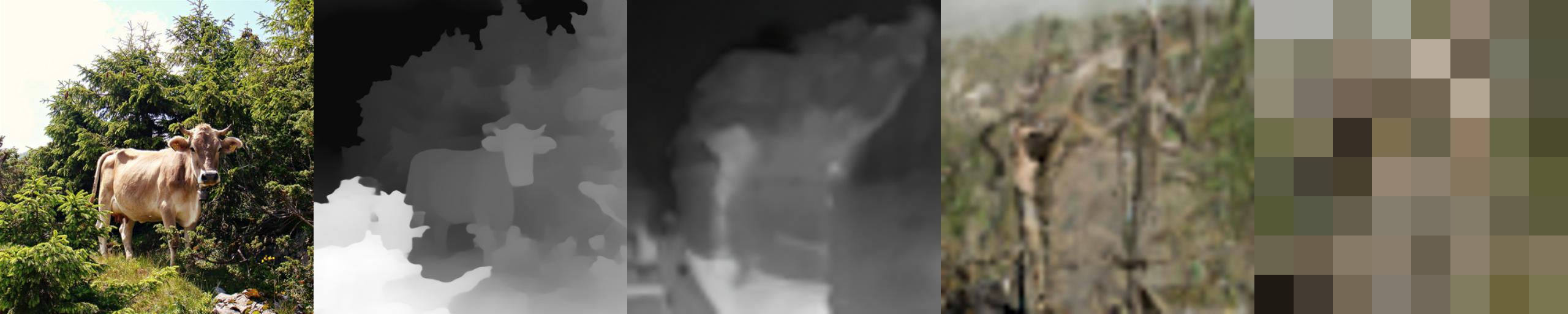}} & & \multicolumn{5}{c}{\includegraphics[width=0.5\linewidth]{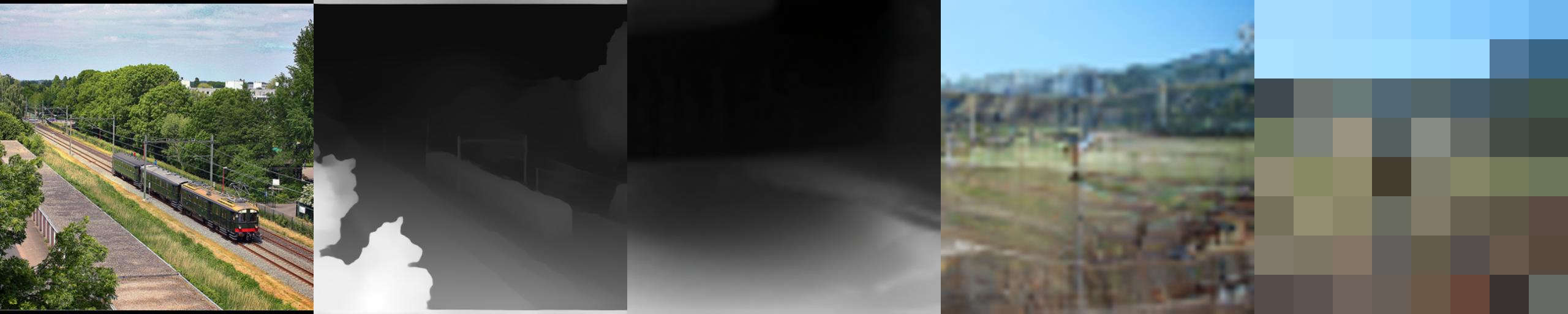}} \\
\multicolumn{5}{c}{\includegraphics[width=0.5\linewidth]{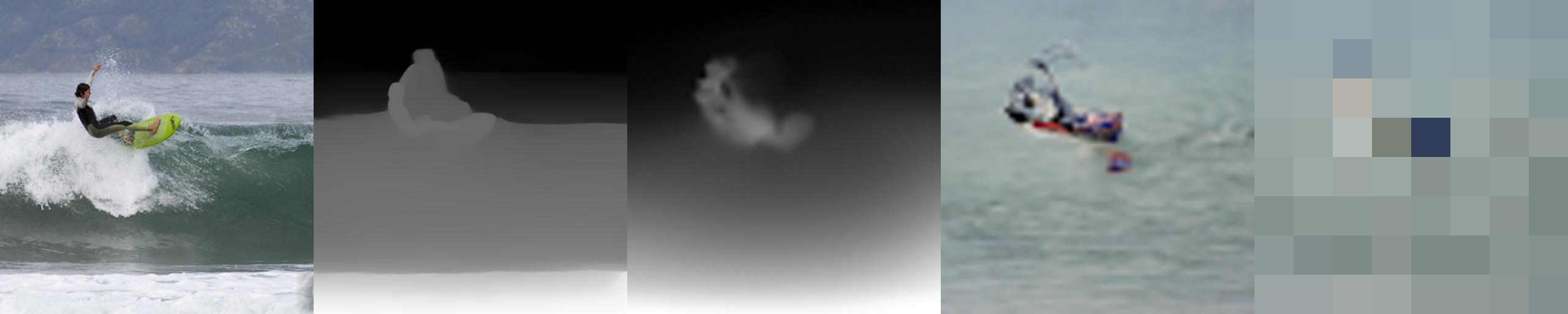}} & & \multicolumn{5}{c}{\includegraphics[width=0.5\linewidth]{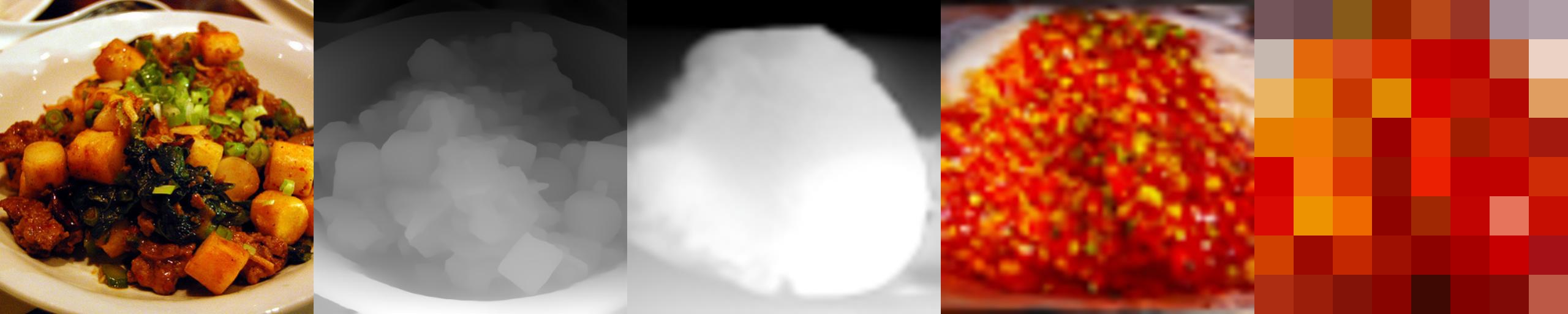}} \\
\multicolumn{5}{c}{\includegraphics[width=0.5\linewidth]{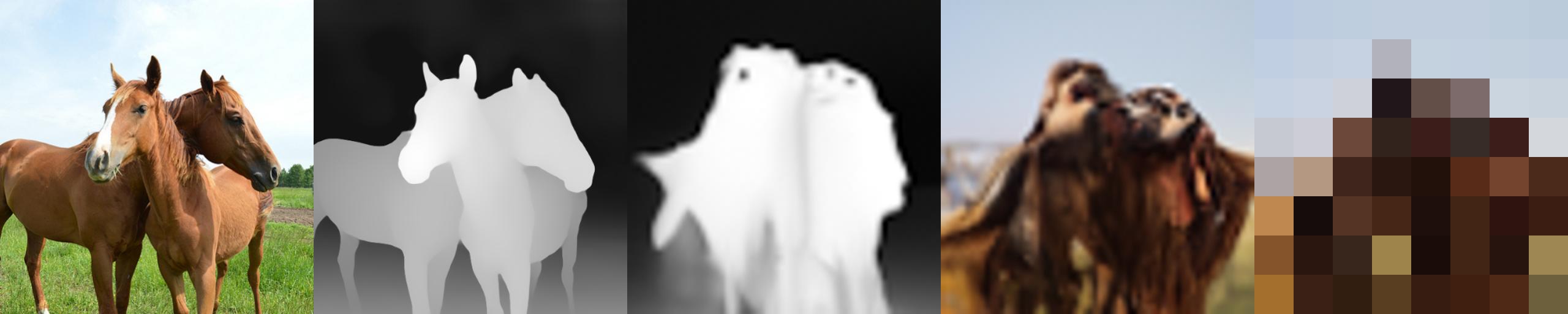}} & & \multicolumn{5}{c}{\includegraphics[width=0.5\linewidth]{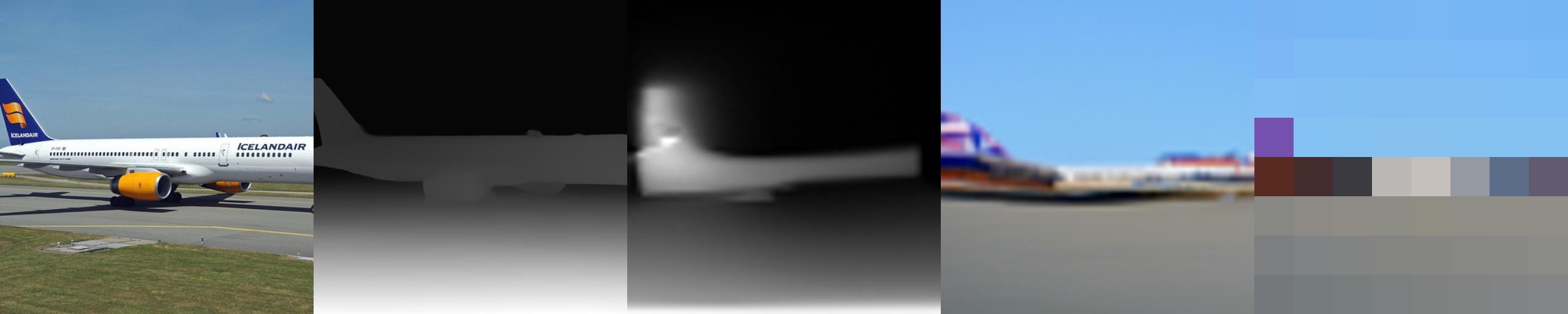}} \\
\end{tabular}
}
\caption{\textbf{\method Decoding of Depth and Color.} 
We display the test image corresponding to fMRI, alongside the depth ground truth ($\texttt{D}$) and the depth/color predictions ($\hat{\texttt{D}}, \hat{\texttt{C}}$). The R-PKM component predicts depth maps and the initial guessed RGB images ($\hat{\texttt{I}}_0$). The predicted spatial palettes are derived from these initial guessed images. The results highlight the proficiency of our R-PKM module in capturing and converting intricate aspects from fMRI recordings into essential cues for visual reconstructions.
}
\label{fig:supmat_fmri_to_depth_color_result}
\end{figure*}

\subsection{Evaluation Methodology}
\label{subsec:supmat_evaluation_metrics}

\paragraph{Metrics for Visual Decoding.} For visual decoding metrics, we employ the same suite of eight evaluation criteria as previously used in research~\cite{scotti2023reconstructing,ozcelik2022reconstruction,gu2022decoding,takagi2023high}. PixCorr, SSIM, AlexNet(2), and AlexNet(5) are categorized as low-level, while Inception, CLIP, EffNet-B, and SwAV are considered high-level.
Following~\cite{ozcelik2022reconstruction}, we downsampled the generated images from a $512\times512$ resolution to a $425\times425$ resolution (corresponding to the resolution of ground truth images in the NSD dataset) for PixCorr and SSIM metrics. For the other metrics, the generated images were adjusted based on the input specifications of each respective network. 
It should be noted that not all evaluation outcomes are available for earlier models, depending on the metrics they chose to experiment with.
Our quantitative comparisons with MindEye~\cite{scotti2023reconstructing}, Takagi~\etal~\cite{takagi2023high}, and Gu~\etal~\cite{gu2022decoding} are made according to the exact same test set, \ie, the 982 images that are shared for all 4 subjects. Lin~\etal~\cite{lin2022mind} disclosed their findings exclusively for Subject 1, with a custom training-test dataset split.

\paragraph{Metrics for Depth and Color.} 
We additionally measure consistency of our extracted depth and color. We borrow some common metrics from depth estimation~\cite{ming2021deep} and color correction~\cite{xia2017color} to assess depth and color consistencies in the final reconstructed images.
For depth metrics, we report Abs~Rel (absolute error), Sq~Rel (squared error), RMSE (root mean squared error), and RMSE log (root mean squared logarithmic error) --- detailed in~\cite{ming2021deep}.

For color metrics, we use CD (Color Discrepancy)~\cite{xia2017color} and STRESS (Standardized Residual Sum of Squares)~\cite{garcia2007measurement}. 
CD calculates the absolute differences between the ground truth ${I}$ and the reconstructed image $\hat{I}$ by utilizing the normalized histograms of images segmented into bins:
\begin{equation}
\mathrm{CD}({I}, \hat{I})=\sum\left|\mathcal{H}\left({I}\right)-\mathcal{H}(\hat{I})\right|,
\end{equation}
where $\mathcal{H}(\cdot)$ represents the histogram function over the given range (\eg~[0,~255]) and number of bins. In simpler terms, this equation computes the absolute difference between the histograms of the two images for all bins and then sums them up. The number of bins for histogram is set as 64.
STRESS calculates a scaled difference between the ground-truth $\texttt{C}$ and the estimated color palette $\hat{\texttt{C}}$:
\begin{equation}
   \text{STRESS}=100 \sqrt{\frac{\sum_{i=1}^n\left(F \hat{\texttt{C}}_i-\texttt{C}_i\right)^2}{\sum_{i=1}^n \texttt{C}_i^2}},
\end{equation}
where $n$ is the number of samples and $F$ is calculated as
\begin{equation}
F=\frac{\sum_{i=1}^n \hat{\texttt{C}}_i \texttt{C}_i}{\sum_{i=1}^n \hat{\texttt{C}}_i^2}.
\end{equation}

\section{Additional \method Results}
\label{sec:supmat_more_results}

This section presents additional results of our method, to showcase the effectiveness of \method. \cref{subsec:supmat_depth_color} presents the \makebox{fMRI $\mapsto$ depth} \& color results, which demonstrates how the deciphered and represented color and depth information helps to boost the performance of visual decoding. 
\cref{subsec:supmat_more_reconstruction_result} provides more examples of fMRI test reconstructions from subject 1. The results shows that the extracted essential cues from fMRI recordings lead to enhanced consistency in appearance, structure, and semantics when compared to the viewed visual stimuli.
\cref{subsec:supmat_other_subject} provides results of all four subjects.

\begin{figure}
\centering
\renewcommand{\arraystretch}{0.5}
\setkeys{Gin}{width=0.15\linewidth}
\setlength{\tabcolsep}{1.2pt}
\scriptsize
\resizebox{\linewidth}{!}
{
\begin{tabular}[t]{cccccccc}
\parbox[t]{2mm}{\multirow{1}{*}{\rotatebox[origin=c]{90}{Test image\hspace{-5em}}}}&\multirow{1}{*}{\rotatebox[origin=c]{90}{$\texttt{I}$\hspace{-5.5em}}}&\includegraphics{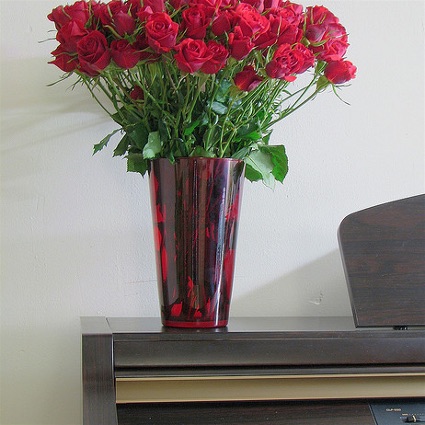} & \includegraphics{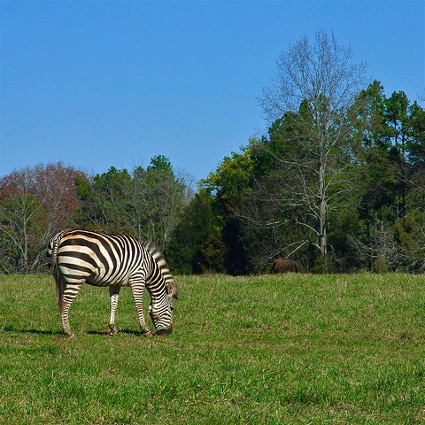} & \includegraphics{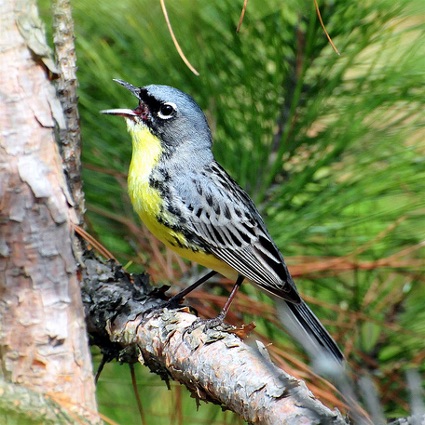} & 
\includegraphics{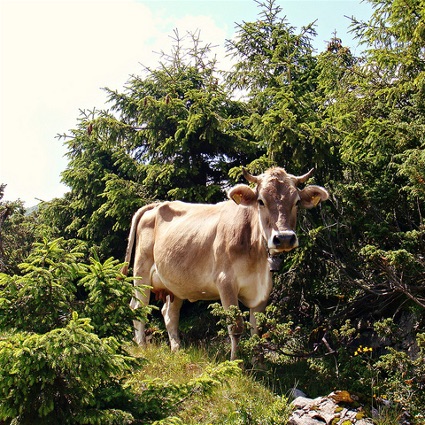} & \includegraphics{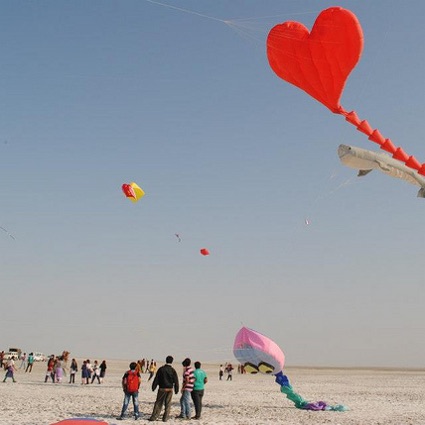} & \includegraphics{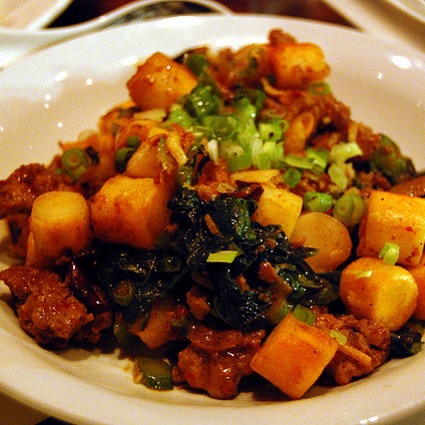} \\
\parbox[t]{2mm}{\multirow{1}{*}{\rotatebox[origin=c]{90}{GT Depth\hspace{-5.3em}}}}&\multirow{1}{*}{\rotatebox[origin=c]{90}{${\texttt{D}}$\hspace{-5em}}}&\includegraphics{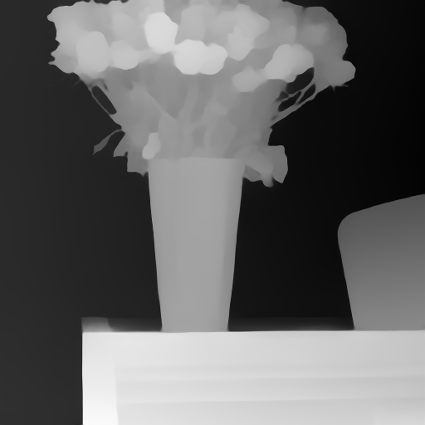} & \includegraphics{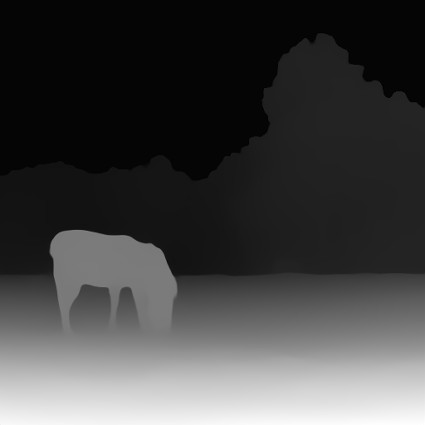} & \includegraphics{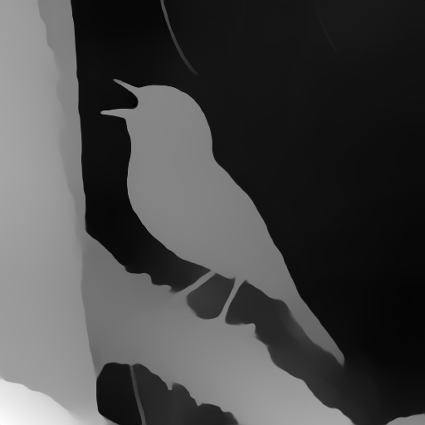} & 
\includegraphics{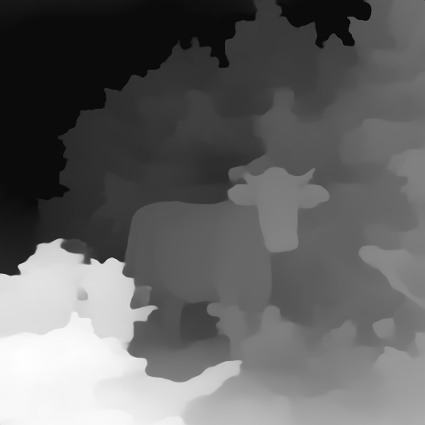} & \includegraphics{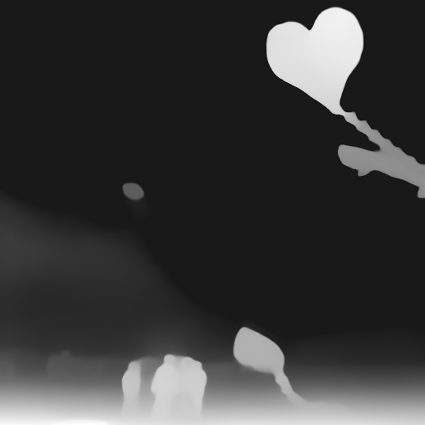} & \includegraphics{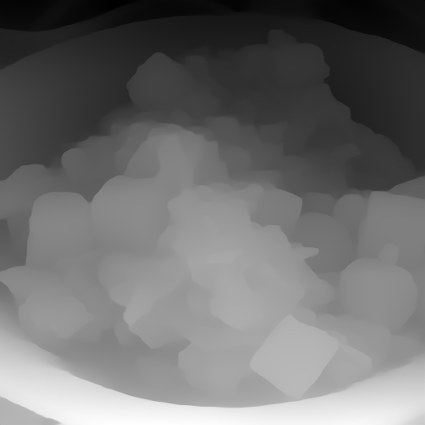} \\
\parbox[t]{2mm}{\multirow{1}{*}{\rotatebox[origin=c]{90}{Pred. Depth\hspace{-5em}}}}&\multirow{1}{*}{\rotatebox[origin=c]{90}{$\hat{\texttt{D}}$\hspace{-5em}}}&\includegraphics{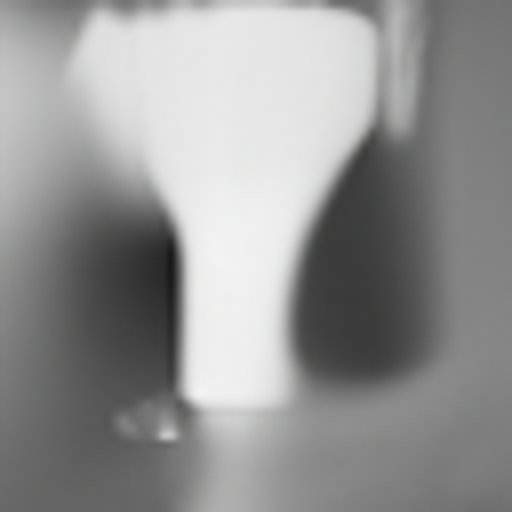} & \includegraphics{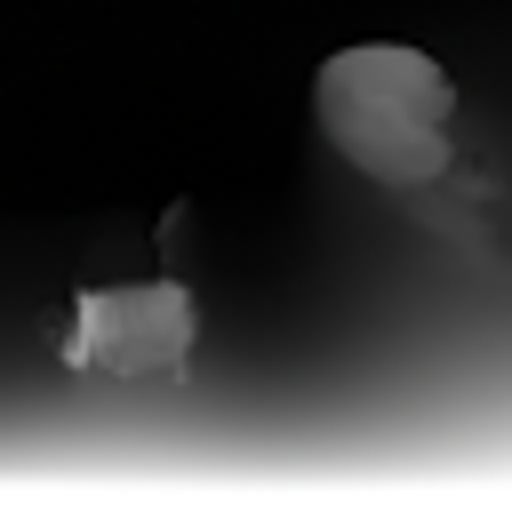} & \includegraphics{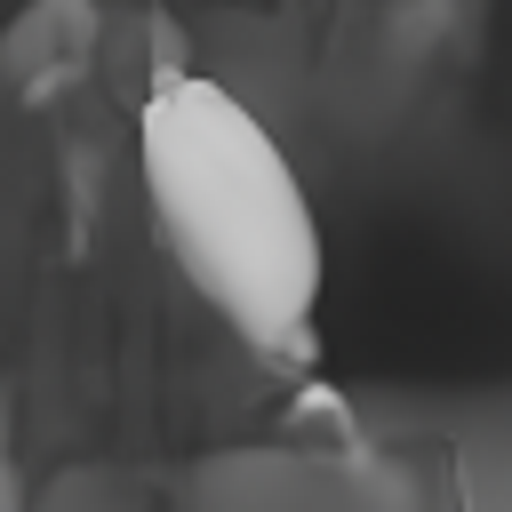} & 
\includegraphics{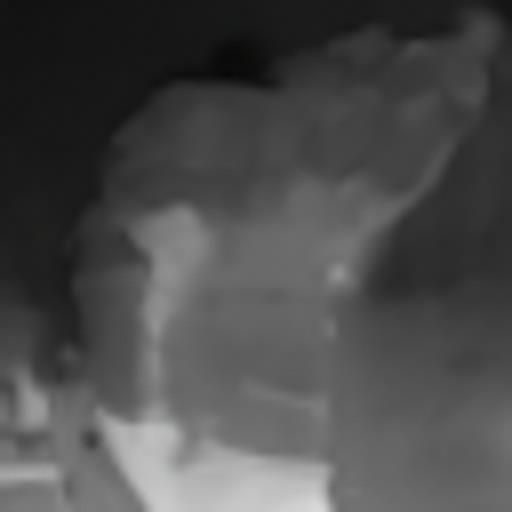} & \includegraphics{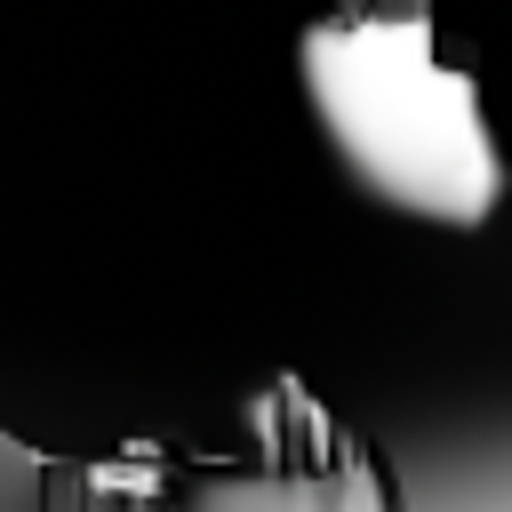} & \includegraphics{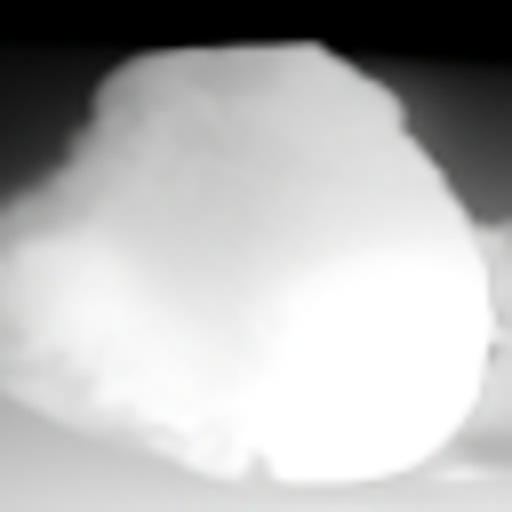} \\
\end{tabular}
}
\caption{\textbf{Sample Depth}. We show sample depth maps ($\hat{\texttt{D}}$) deciphered from fMRI using R-PKM, alongside the ground-truth depth ($\texttt{D}$) estimated from MiDaS~\cite{ranftl2020towards} on the original test image ($I$).}
\label{fig:depth_results}
\end{figure}

\begin{figure*}
\centering
\setlength\tabcolsep{0pt}
\renewcommand{\arraystretch}{0.8}
\setkeys{Gin}{width=0.95\linewidth}
\setlength{\tabcolsep}{0pt}
\footnotesize
\resizebox{0.90\linewidth}{!}{
\begin{tabular}{>{\centering\arraybackslash}p{0.1\linewidth}>{\centering\arraybackslash}p{0.1\linewidth}>{\centering\arraybackslash}p{0.1\linewidth}>{\centering\arraybackslash}p{0.1\linewidth}>{\centering\arraybackslash}p{0.1\linewidth}>{\centering\arraybackslash}p{0.1\linewidth}>{\centering\arraybackslash}p{0.1\linewidth}>{\centering\arraybackslash}p{0.1\linewidth}} %
&&&&&&&\\ %
Test image&GT (${\texttt{D}}$)&\multicolumn{2}{c}{Predictions ($\hat{\texttt{D}}$, $\hat{\texttt{C}}$)}&\multicolumn{4}{c}{\colorname}\\
\cmidrule(lr){2-2}\cmidrule(lr){3-4}\cmidrule(lr){5-8}
\multicolumn{8}{c}{\includegraphics[width=0.8\linewidth]{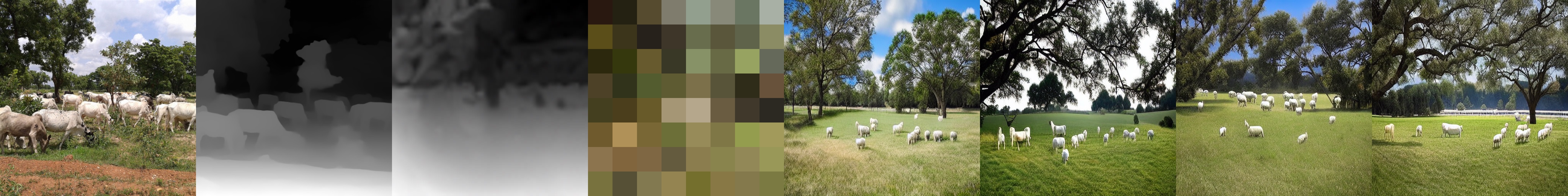}} \\
\multicolumn{8}{c}{\includegraphics[width=0.8\linewidth]{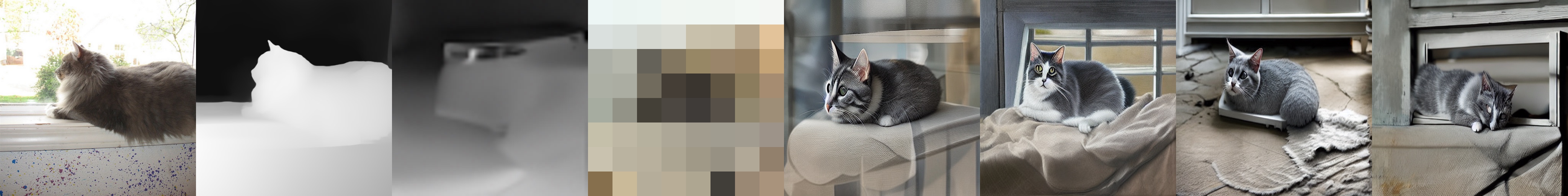}} \\
\multicolumn{8}{c}{\includegraphics[width=0.8\linewidth]{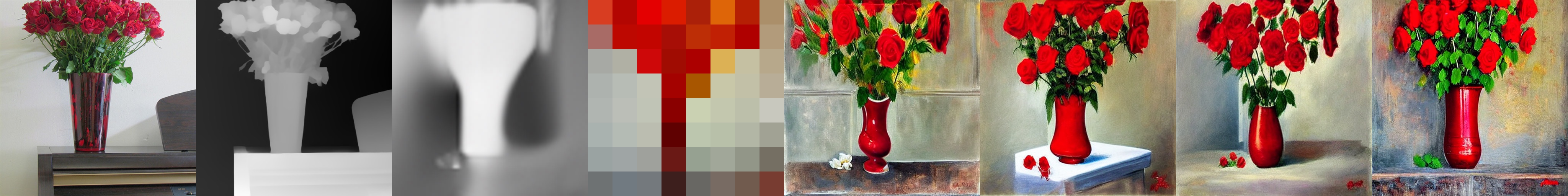}} \\
\multicolumn{8}{c}{\includegraphics[width=0.8\linewidth]{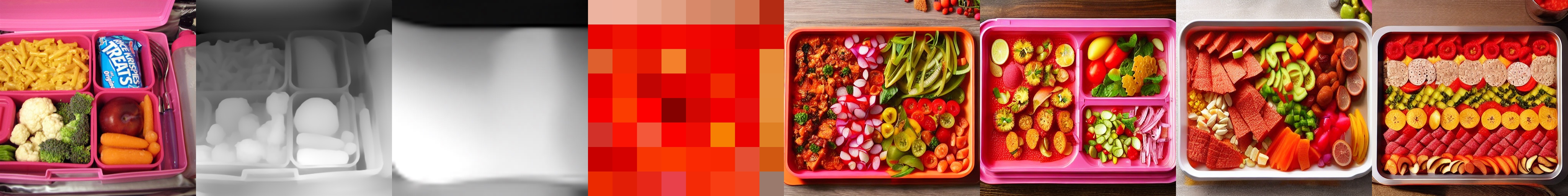}} \\
\multicolumn{8}{c}{\includegraphics[width=0.8\linewidth]{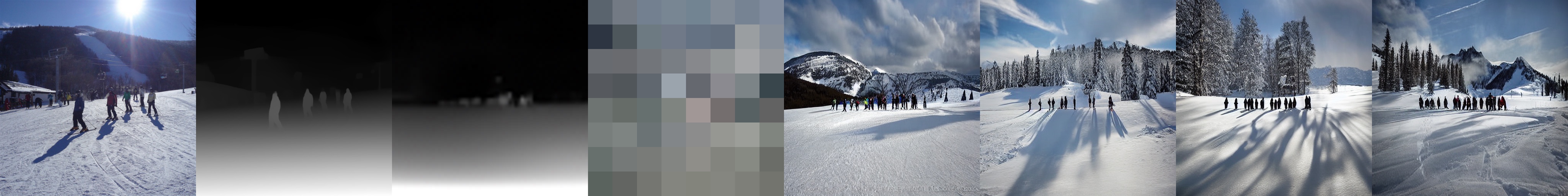}} \\
\multicolumn{8}{c}{\includegraphics[width=0.8\linewidth]{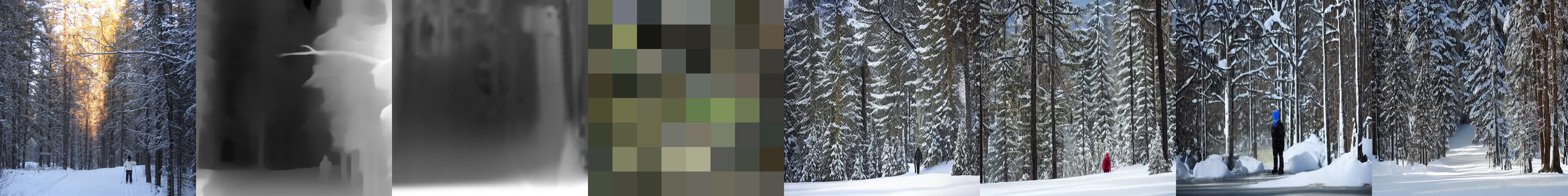}} \\
\multicolumn{8}{c}{\includegraphics[width=0.8\linewidth]{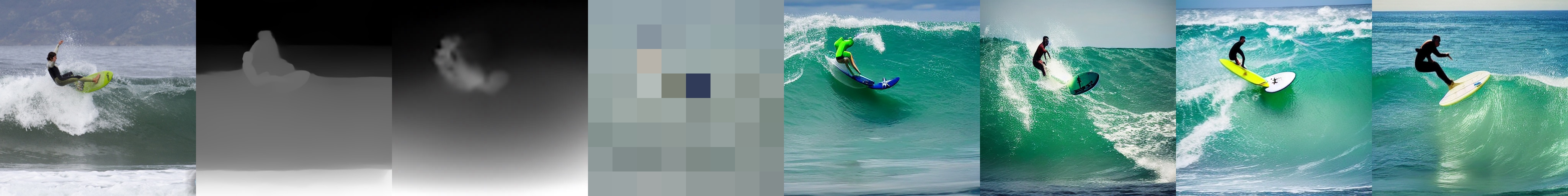}} \\
\multicolumn{8}{c}{\includegraphics[width=0.8\linewidth]{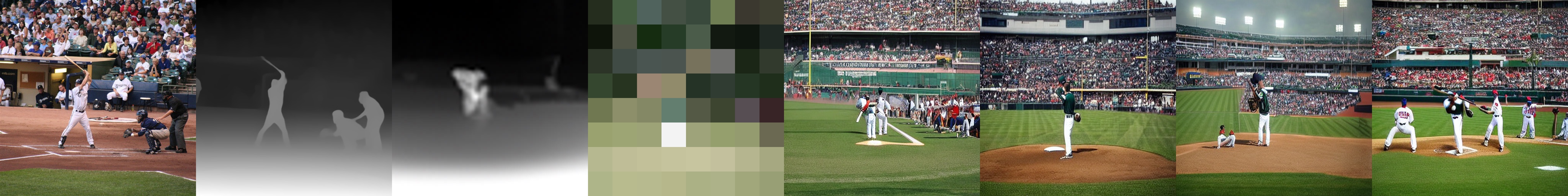}} \\
\multicolumn{8}{c}{\includegraphics[width=0.8\linewidth]{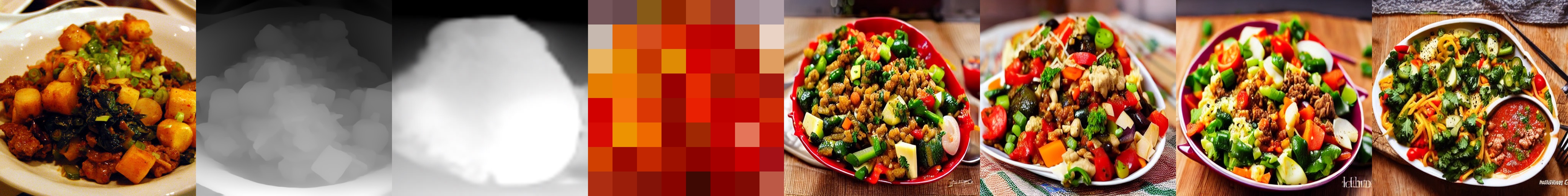}} \\
\multicolumn{8}{c}{\includegraphics[width=0.8\linewidth]{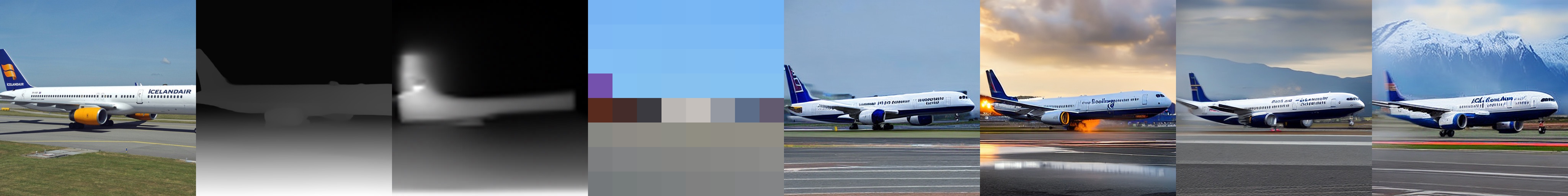}} \\
\end{tabular}
}
\vspace{-2mm}
\caption{\textbf{\method Reconstructions.} We show reconstruction for subject 1 (sub01) from the NSD dataset. 
Our approach extracts essential cues from fMRI recordings, leading to enhanced consistency in appearance, structure, and semantics when compared to the viewed visual stimuli. The results are randomly selected. The illustrated depth, color, and final images demonstrate that the deciphered and represented color and depth cues help to boost the performance of visual decoding.}
\label{fig:supmat_more_reconstruction_result}
\end{figure*}

\begin{figure*}[htbp]
\center
\setlength\tabcolsep{0pt}
{
\renewcommand{\arraystretch}{0.5}
\setkeys{Gin}{width=0.09\linewidth}
\small
\setlength{\tabcolsep}{1.2pt}{
{
\begin{tabular}{c|ccccccccccc} 
    \multicolumn{1}{c}{}&\parbox[t]{2mm}{\multirow{1}{*}{\rotatebox[origin=c]{90}{Test image\hspace{-5em}}}}&\includegraphics{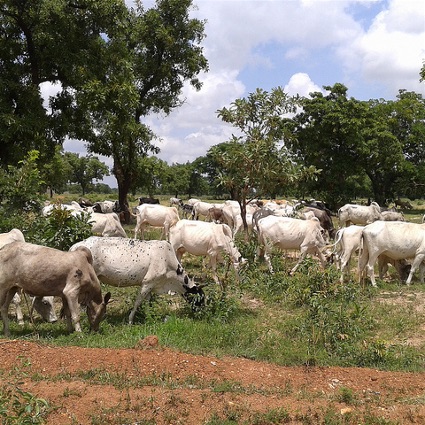} &
    \includegraphics{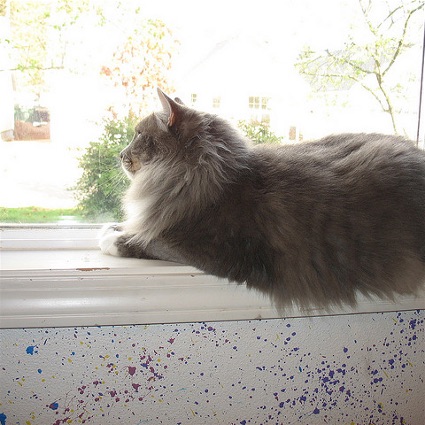} &
    \includegraphics{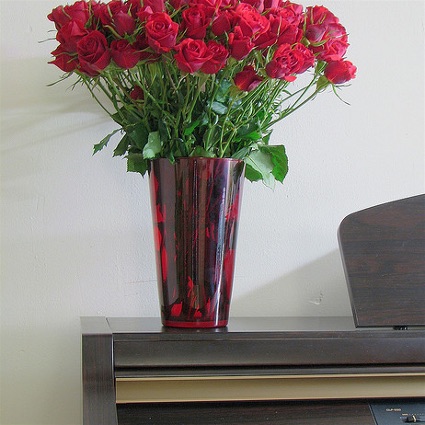} &
    \includegraphics{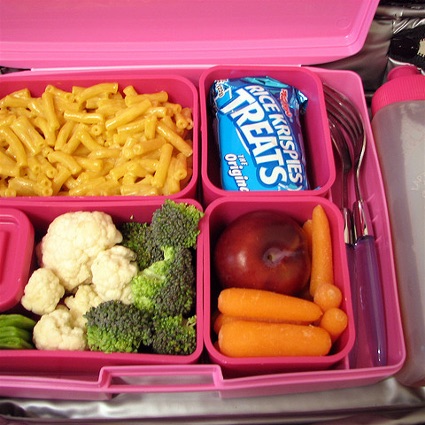} &
    \includegraphics{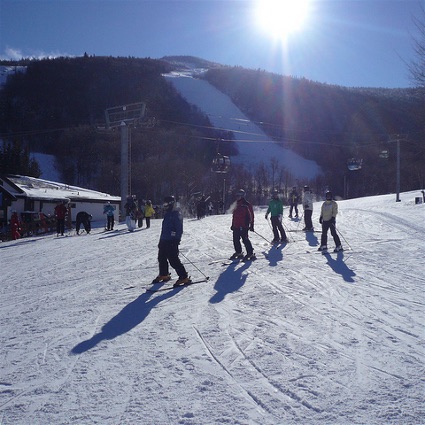} &
    \includegraphics{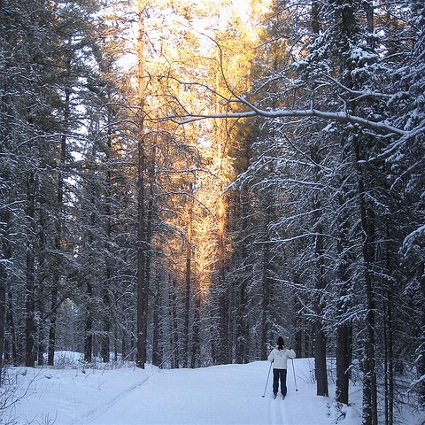} &
    \includegraphics{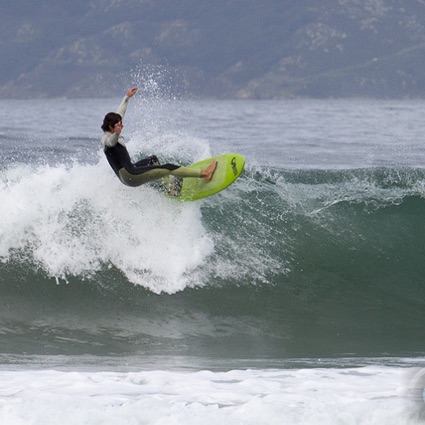} &
    \includegraphics{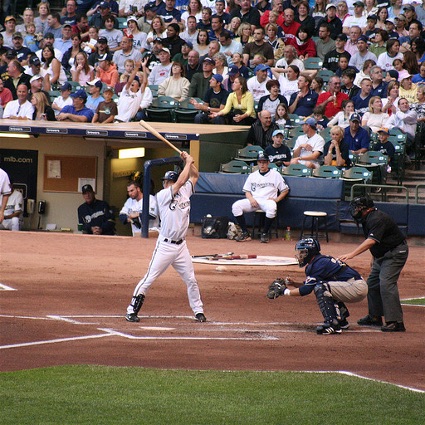} &
    \includegraphics{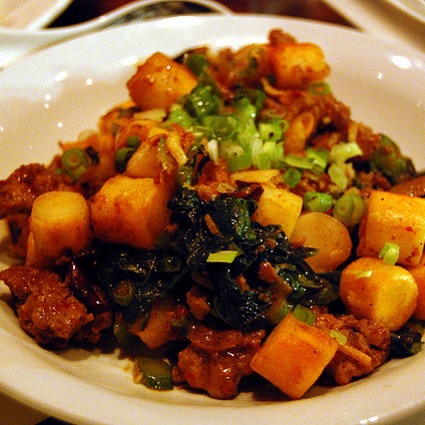} &
    \includegraphics{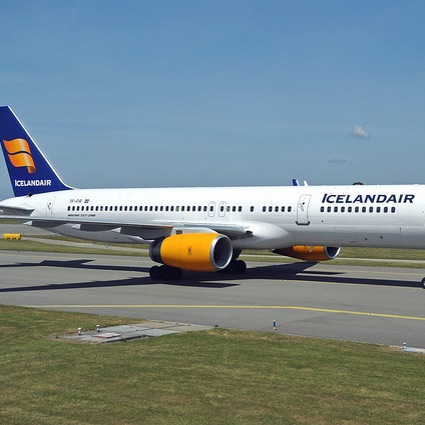} \\
    \multirow{4}{*}{\rotatebox[origin=c]{90}{\colorname\hspace{3.5em}}~}&\parbox[t]{2mm}{\multirow{1}{*}{\rotatebox[origin=c]{90}{sub01\hspace{-5em}}}}&
    \adjustbox{trim={.5\width} {.0\height} {0.380\width} {.0\height},clip}{\includegraphics{supmat/supmat_final_image/0.jpeg}} & \adjustbox{trim={.5\width} {.0\height} {0.380\width} {.0\height},clip}{\includegraphics{supmat/supmat_final_image/17.jpeg}} & \adjustbox{trim={.5\width} {.0\height} {0.380\width} {.0\height},clip}{\includegraphics{supmat/supmat_final_image/18.jpeg}} & \adjustbox{trim={.5\width} {.0\height} {0.380\width} {.0\height},clip}{\includegraphics{supmat/supmat_final_image/33.jpeg}} & \adjustbox{trim={.5\width} {.0\height} {0.380\width} {.0\height},clip}{\includegraphics{supmat/supmat_final_image/37.jpeg}} & \adjustbox{trim={.5\width} {.0\height} {0.380\width} {.0\height},clip}{\includegraphics{supmat/supmat_final_image/44.jpeg}} & \adjustbox{trim={.5\width} {.0\height} {0.380\width} {.0\height},clip}{\includegraphics{supmat/supmat_final_image/53.jpeg}} & \adjustbox{trim={.5\width} {.0\height} {0.380\width} {.0\height},clip}{\includegraphics{supmat/supmat_final_image/55.jpeg}} & \adjustbox{trim={.5\width} {.0\height} {0.380\width} {.0\height},clip}{\includegraphics{supmat/supmat_final_image/74.jpeg}} & \adjustbox{trim={.5\width} {.0\height} {0.380\width} {.0\height},clip}{\includegraphics{supmat/supmat_final_image/115.jpeg}} \\
    &\parbox[t]{2mm}{\multirow{1}{*}{\rotatebox[origin=c]{90}{sub02\hspace{-5em}}}}&\includegraphics{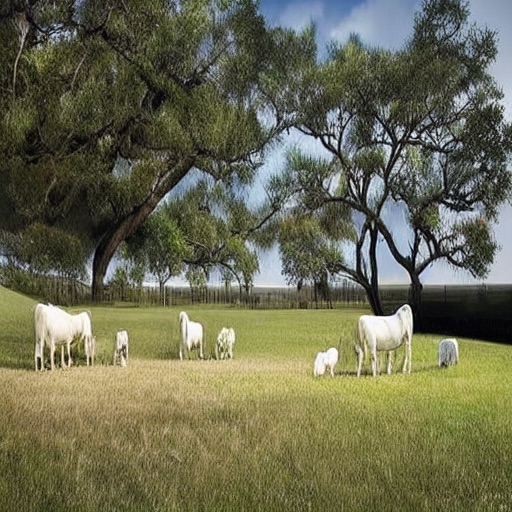} &
    \includegraphics{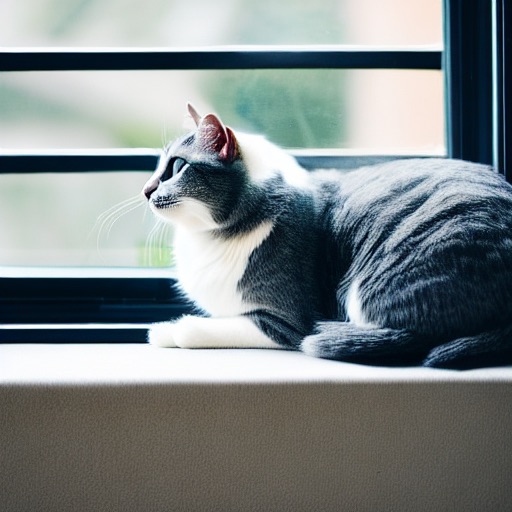} &
    \includegraphics{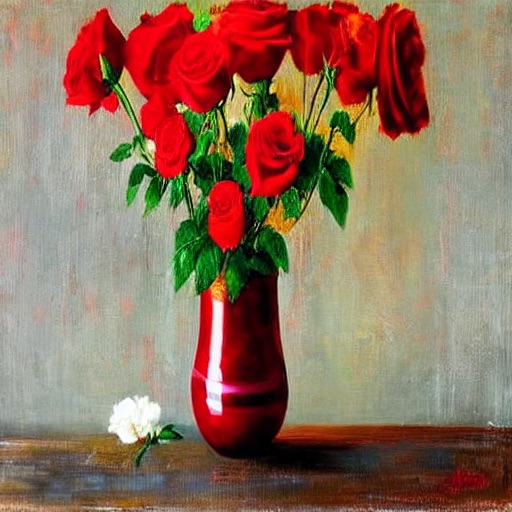} &
    \includegraphics{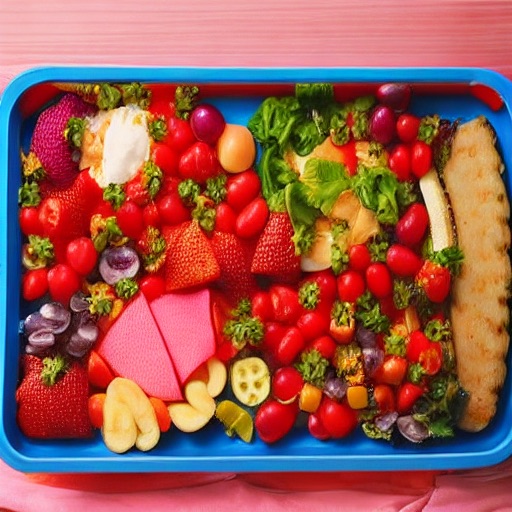} &
    \includegraphics{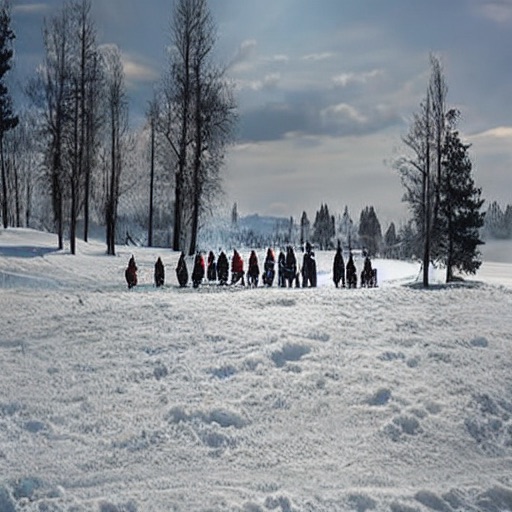} &
    \includegraphics{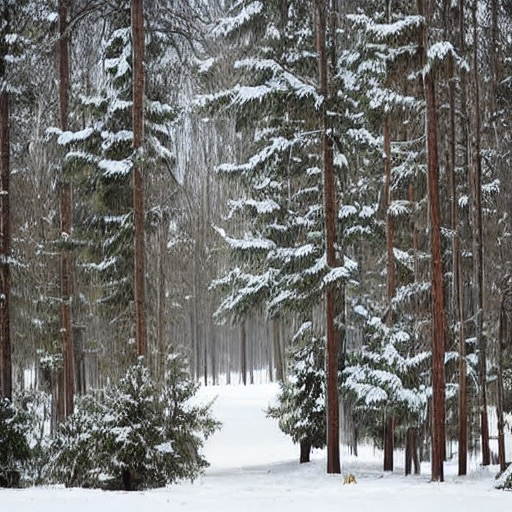} &
    \includegraphics{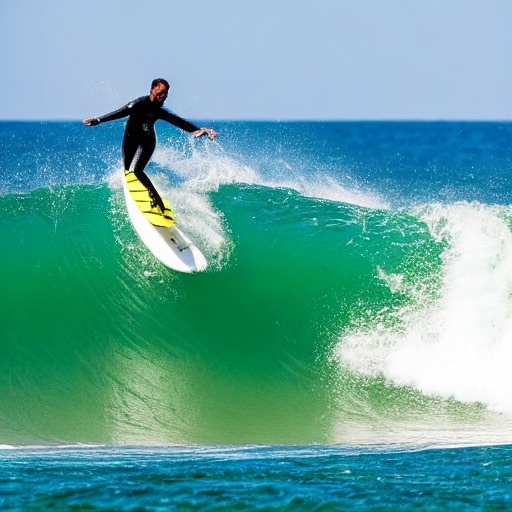} &
    \includegraphics{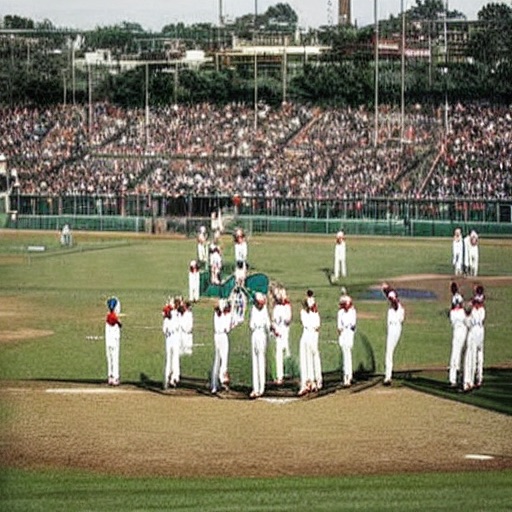} &
    \includegraphics{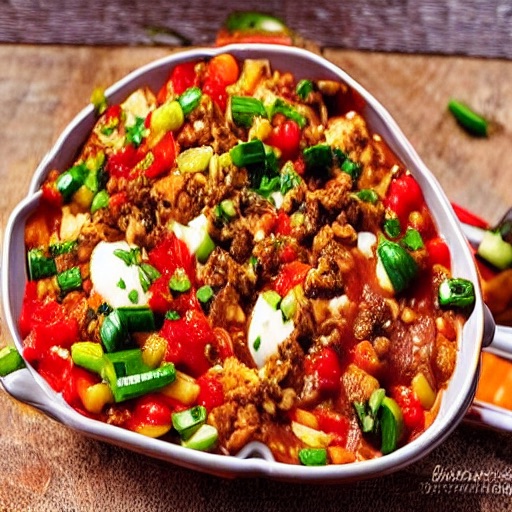} &
    \includegraphics{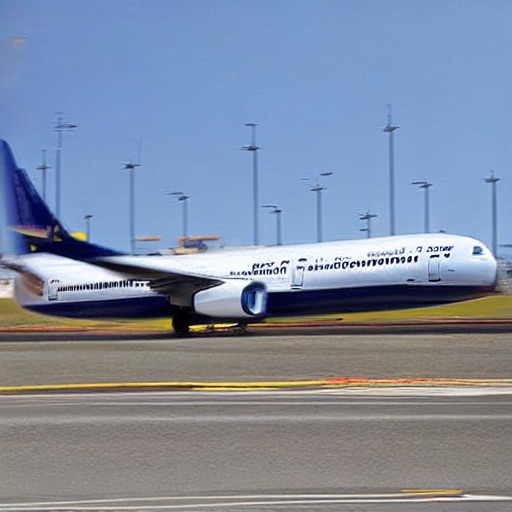} \\
    &\parbox[t]{2mm}{\multirow{1}{*}{\rotatebox[origin=c]{90}{sub05\hspace{-5em}}}}&\includegraphics{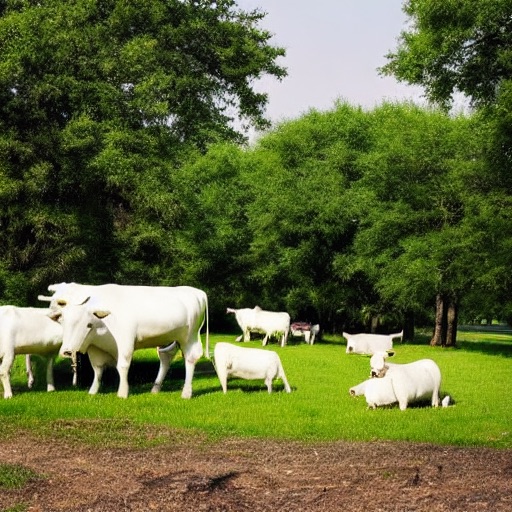} &
    \includegraphics{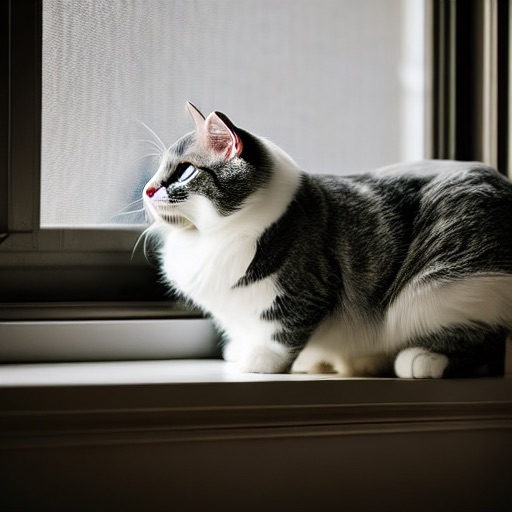} &
    \includegraphics{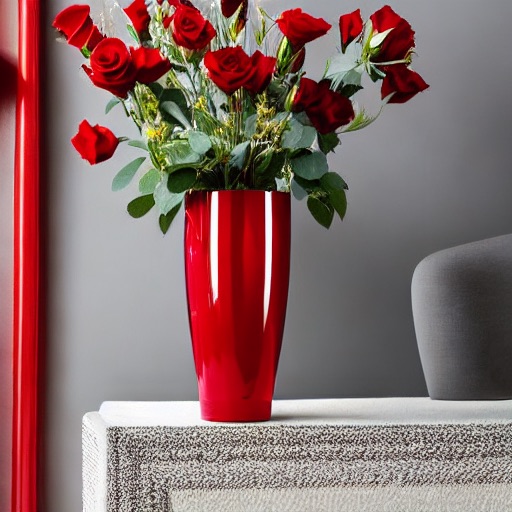} &
    \includegraphics{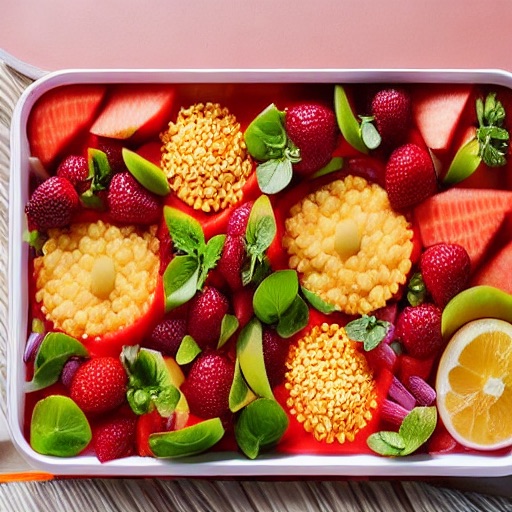} &
    \includegraphics{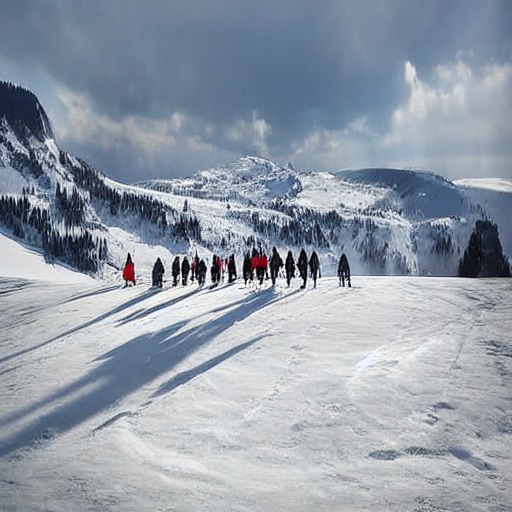} &
    \includegraphics{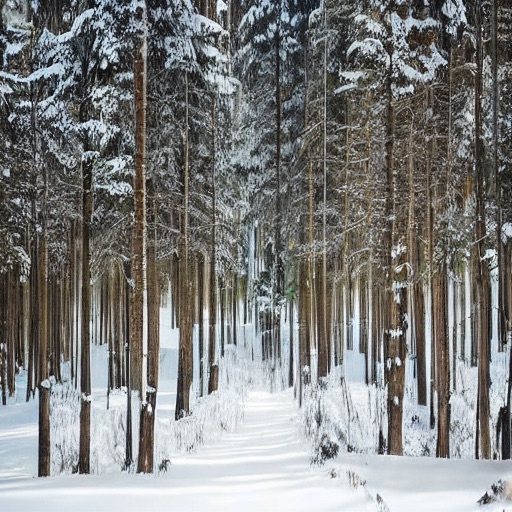} &
    \includegraphics{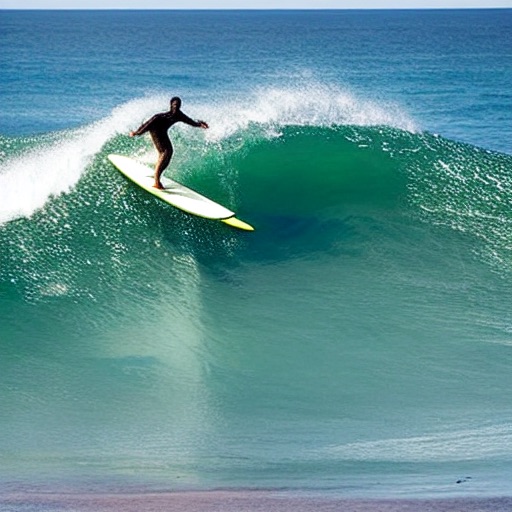} &
    \includegraphics{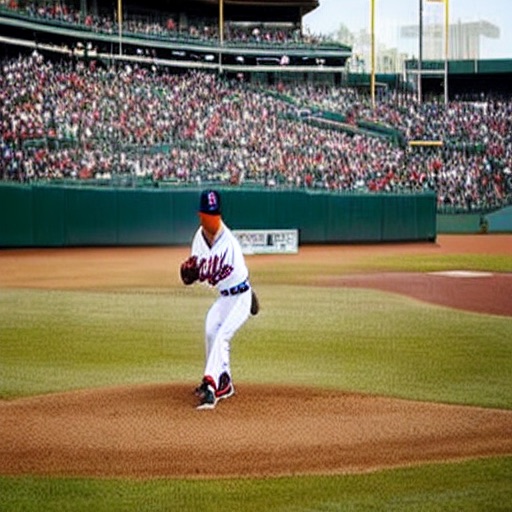} &
    \includegraphics{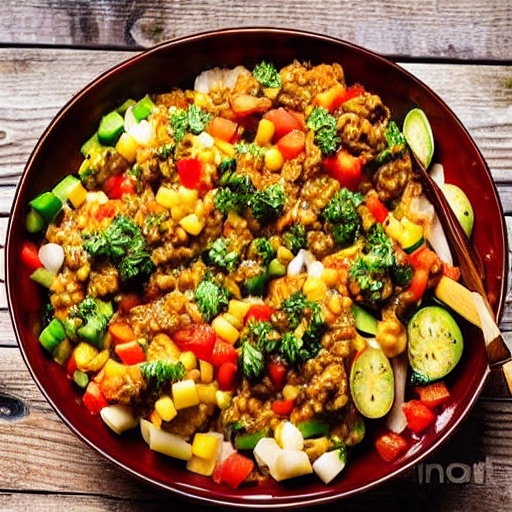} &
    \includegraphics{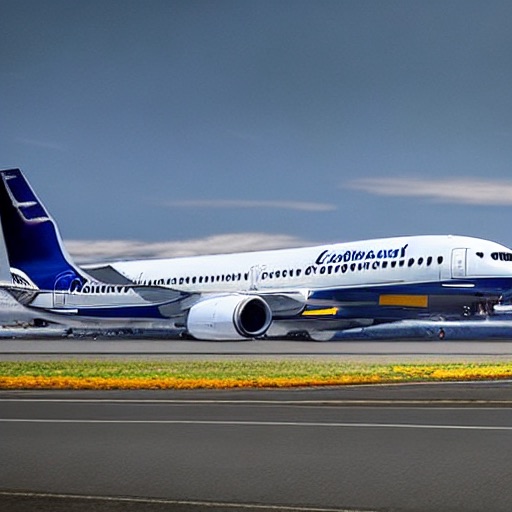} \\
    &\parbox[t]{2mm}{\multirow{1}{*}{\rotatebox[origin=c]{90}{sub07\hspace{-5em}}}}&\includegraphics{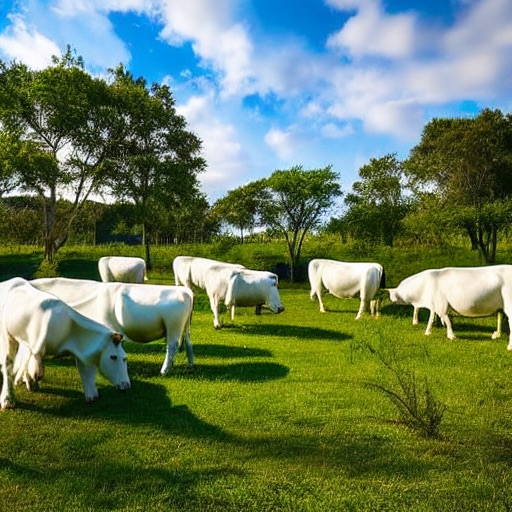} &
    \includegraphics{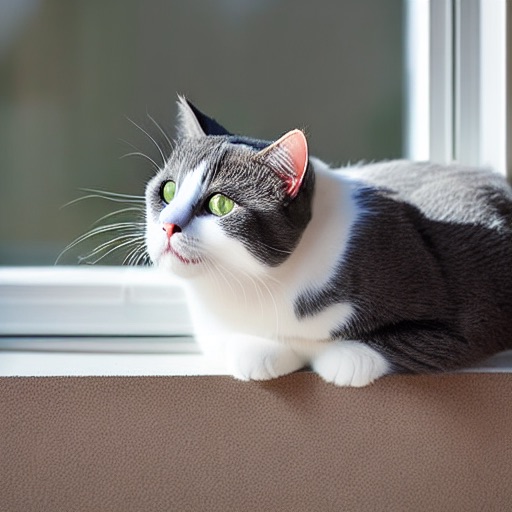} &
    \includegraphics{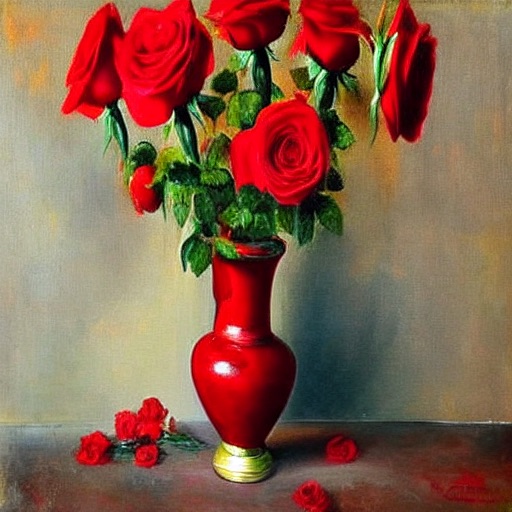} &
    \includegraphics{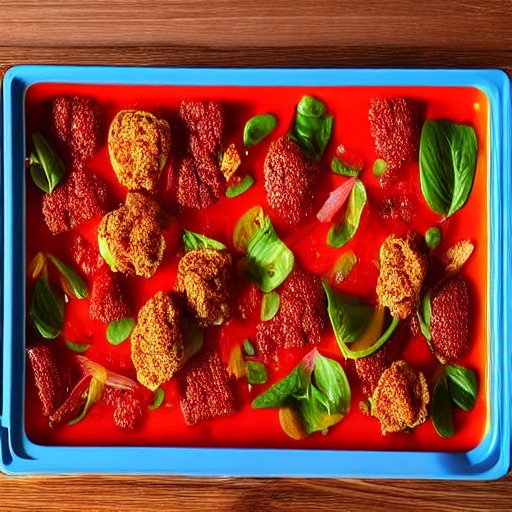} &
    \includegraphics{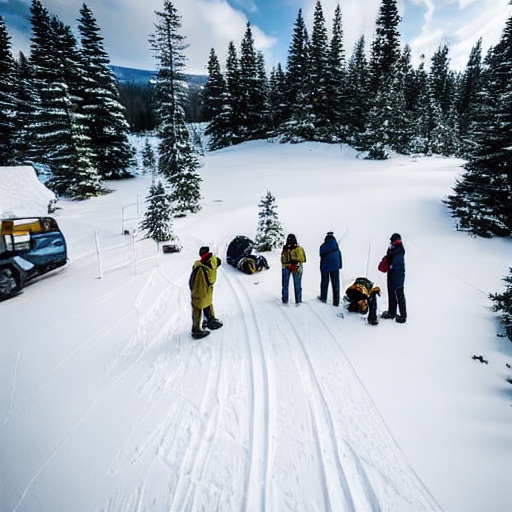} &
    \includegraphics{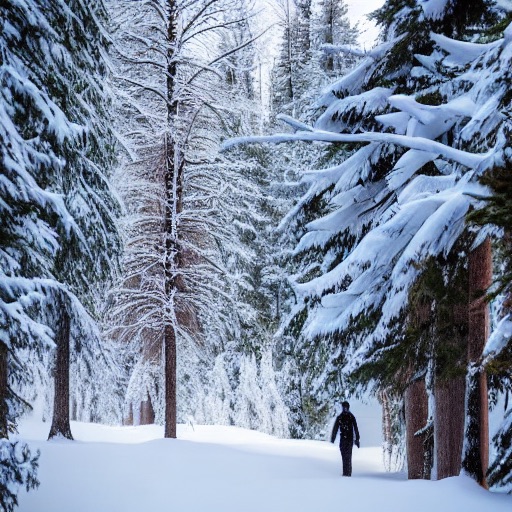} &
    \includegraphics{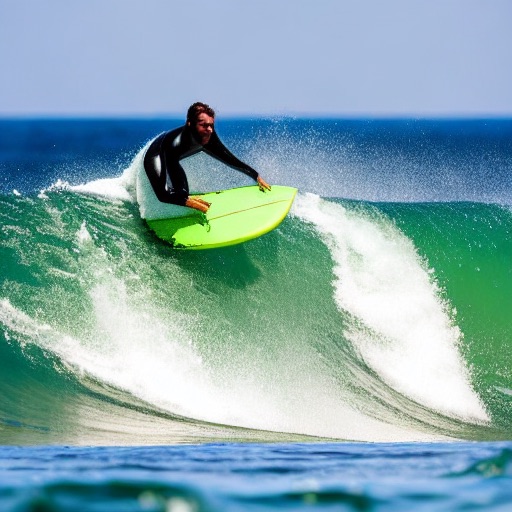} &
    \includegraphics{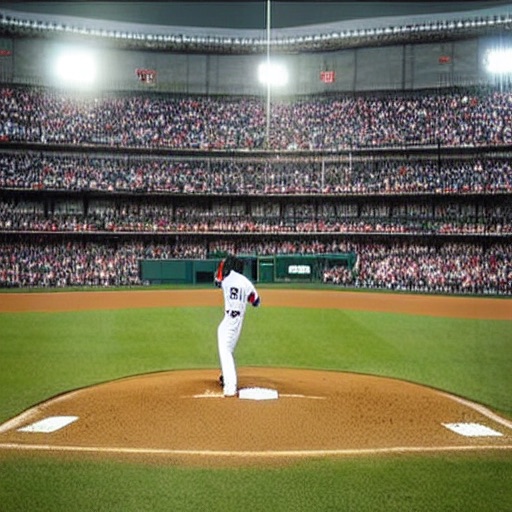} &
    \includegraphics{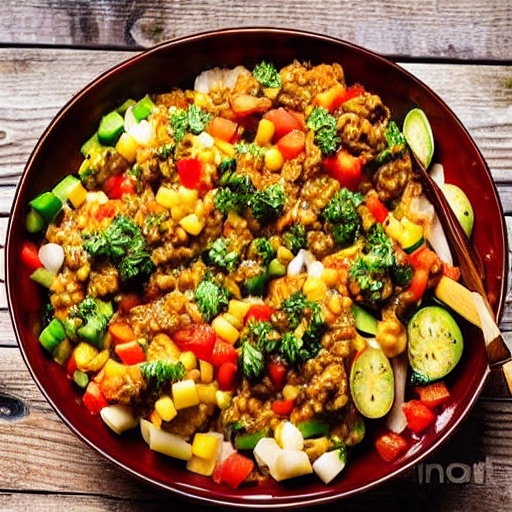} &
    \includegraphics{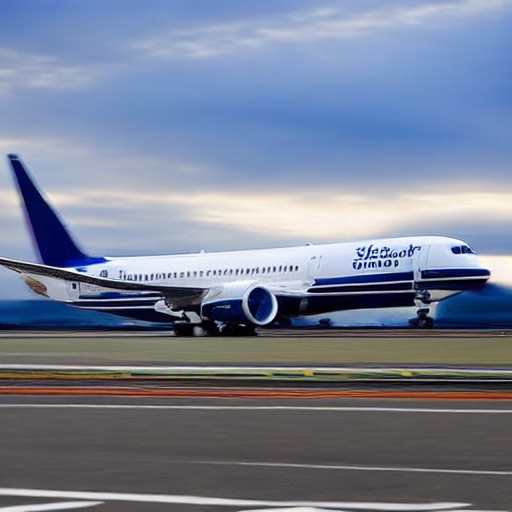} \\
\end{tabular}
}
}
}
\caption{\textbf{Subject-Specific Results.} We visualize subject-specific outputs of \method on the NSD dataset. For each subject, the model is retrained because the brain activity varies across subjects. Overall, it consistently reconstructs the test image for all subjects while we note that some reconstruction inaccuracies are shared across subjects (cf.~\cref{subsec:supmat_other_subject}). Quantitative metrics are in~\cref{tab:supmat_other_subject}.}
\label{fig:supmat_other_subject}
\end{figure*}

\begin{table*}
    \centering
    \caption{\textbf{Subject-Specific Evaluation.} Quantitative evaluation of the \method reconstructions for the participants (sub01, sub02, sub05, and sub07) of the NSD dataset. Performance is stable accross all participants, and consistent with the results reported in the main paper. Some example visual results can be found in~\cref{fig:supmat_other_subject}. }
    \label{tab:supmat_other_subject}
    \resizebox{0.85\linewidth}{!}{
        \begin{tabular}{l|cccc|cccc}
            \toprule
            \multirow{2}{*}{Subject}  & \multicolumn{4}{c|}{Low-Level} & \multicolumn{4}{c}{High-Level}\\
            ~ & PixCorr $\uparrow$ & SSIM $\uparrow$ & AlexNet(2) $\uparrow$ & AlexNet(5) $\uparrow$ & Inception $\uparrow$ & CLIP $\uparrow$  & EffNet-B $\downarrow$ & SwAV $\downarrow$\\
            \midrule
            sub01 & .288 & .338 & 95.0$\%$ & 97.5$\%$ & 94.8$\%$ & 95.2$\%$ & .638 & .413 \\
            sub02 & .273 & .331 & 94.2$\%$ & 97.1$\%$ & 93.4$\%$ & 93.5$\%$ & .652 & .422 \\
            sub05 & .269 & .325 & 93.5$\%$ & 96.6$\%$ & 93.8$\%$ & 94.1$\%$ & .633 & .397 \\
            sub07 & .265 & .319 & 92.7$\%$ & 95.4$\%$ & 92.6$\%$ & 93.7$\%$ & .656 & .438 \\
            \bottomrule
        \end{tabular}
    }
\end{table*}

\subsection{Depth \& Color Deciphering}
\label{subsec:supmat_depth_color}

\cref{fig:supmat_fmri_to_depth_color_result} showcases additional depth and color results deciphered from the R-PKM component. Overall, it is able to capture and translate these intricate aspects from fMRI recordings to spatial guidance crucial for more accurate image reconstructions.

For depth, the second and third columns show example depth reconstruction alongside their corresponding estimated ground truth obtained from the original RGB images. Results show that the depth estimated, while far from perfect, is sufficient to provide coarse guidance on the scene structure and object position/orientation for our reconstruction guidance purpose. 

The last two columns show the color results. 
The predicted spatial palettes are generated by downscaling the ``initial guessed images'' denoted $\hat{\texttt{I}}_0$ (not to be confused with $\hat{\texttt{I}}$) which corresponds to the RGB channels of the R-PKM RGBD output.
As discussed in~\cref{subsec:supmat_representations}, employing a $\times64$ downsampling on the ``initial guessed images'' achieves a trade-off between efficiently extracting essential color cues and effectively mitigating the inaccuracies in these images.
Despite not accurately preserving the color of local regions due to the resolution, the produced color palettes provide a relevant constraint and guidance on the overall color tone. Additional depth outputs are in~\cref{fig:depth_results}.

Although depth and color guidance are sufficient to reconstruct images reasonably resembling the test one, it is yet unclear if better depth and color cues can be extracted from the fMRI data or if depth and color are doomed to be coarse estimation due to loss of data in the fMRI recording.

\subsection{More Image Reconstruction Results}
\label{subsec:supmat_more_reconstruction_result}

More examples of image reconstruction for subject 1 are shown in~\cref{fig:supmat_more_reconstruction_result}. From left to right: the first two columns display the test images and their corresponding ground truth depth maps. The third and fourth columns depict the predicted depth and color, respectively, in the form of depth maps and spatial palettes. The remaining columns represent the final reconstructed images. 
The results are randomly selected. 
The illustrated final images demonstrate that the deciphered and represented color and depth cues help to boost the performance of visual decoding.
Overall, \method evidently extracts good-enough cues from the fMRI recordings, leading to consistent reconstruction of the appearance, structure, and semantics of the viewed visual stimuli.

\subsection{Subject-Specific Results}
\label{subsec:supmat_other_subject}

We used the same standardized training-test data splits as other NSD reconstruction papers~\cite{scotti2023reconstructing,ozcelik2022reconstruction,mozafari2020reconstructing}, training subject-specific models for each of 4 participants (sub01, sub02, sub05, and sub07). More details on the different participants can be found in~\cref{sec:supmat_datasets} and~\cref{tab:supmat_nsd_dataset}. %
\cref{fig:supmat_other_subject} shows \method outputs for all four participants, with individual subject evaluation metrics reported in~\cref{tab:supmat_other_subject}. More sub01 results can be found in~\cref{fig:supmat_more_reconstruction_result}.
Overall, \method proves to work well regardless of the subject. However, it is interesting to note that some reconstructions mistakes are shared across subjects. For example, fMRIs of the \textit{vase flowers} picture (3rd column) are often reconstructed as paintings, except for sub05, and the \textit{food plate} (2nd rightmost column) which is taken at an angle is almost always reconstructed as a more top-view photography. These consistent mistakes across subjects may suggest dataset biases.

\end{appendices}

{\small
\bibliographystyle{ieee_fullname}
\bibliography{reference}
}

\end{document}